%% file: main.tex
\documentclass[10pt,twocolumn,letterpaper]{article}

\usepackage[pagenumbers]{cvpr} 

\input{preamble}

\definecolor{cvprblue}{rgb}{0.21,0.49,0.74}
\usepackage[pagebackref,breaklinks,colorlinks,citecolor=cvprblue]{hyperref}
\usepackage{booktabs}
\usepackage{bbm}
\usepackage{multirow}
\usepackage{float}
\usepackage{amsthm}
\usepackage{color,xcolor}
\usepackage{diagbox}
\usepackage{inconsolata}
\usepackage{colortbl}
\usepackage{url}
\usepackage{amsmath,amssymb,amsfonts}
\usepackage{mathrsfs}
\usepackage[ruled,linesnumbered]{algorithm2e}
\newtheorem{theorem}{Theorem}

\newtheorem{assumption}{Assumption}
\newtheorem{remark}{Remark}
\usepackage[figuresright]{rotating}
\usepackage{textcomp}
\usepackage{graphicx}
\usepackage{flushend} 
\usepackage{adjustbox}
\usepackage{makecell}
\makeatletter

\newcommand{\Rmnum}[1]{\expandafter\@slowromancap\romannumeral #1@}
\def\paperID{***} 
\def\confName{CVPR}
\def\confYear{2025}

\title{Neighborhood and Global Perturbations Supported SAM \\in Federated Learning:  From Local Tweaks To Global Awareness}

\begin{small}

\author{Boyuan Li$\textsuperscript{1}$, Zihao Peng$\textsuperscript{2}$, Yafei Li$\textsuperscript{1,*}$, Mingliang Xu$\textsuperscript{1,*}$,\\ Shengbo Chen$\textsuperscript{3}$, Baofeng Ji$\textsuperscript{4}$, Cong Shen$\textsuperscript{5}$\\
Zhengzhou Univerisity$\textsuperscript{1}$, Beijing Normal University$\textsuperscript{2}$, Henan Univerisity$\textsuperscript{3}$\\
 Henan University of Science and Technology$\textsuperscript{4}$, University of Virginia$\textsuperscript{5}$\\
}
\end{small}
\begin{document}
\maketitle
\begin{abstract}

Federated Learning (FL) can be coordinated under the orchestration of a central server to collaboratively build a privacy-preserving model without the need for data exchange.
However, participant data heterogeneity leads to local optima divergence, subsequently affecting convergence outcomes. Recent research has focused on global sharpness-aware minimization (SAM) and dynamic regularization techniques to enhance consistency between global and local generalization and optimization objectives. Nonetheless, the estimation of global SAM introduces additional computational and memory overhead, while dynamic regularization suffers from bias in the local and global dual variables due to training isolation.
In this paper, we propose a novel FL algorithm, FedTOGA, designed to consider optimization and generalization objectives while maintaining minimal uplink communication overhead. By linking local perturbations to global updates, global generalization consistency is improved.    Additionally, global updates are used to correct local dynamic regularizers, reducing dual variables bias and enhancing optimization consistency.    Global updates are passively received by clients, reducing overhead.
We also propose neighborhood perturbation to approximate local perturbation, analyzing its strengths and limitations. Theoretical analysis shows FedTOGA achieves faster convergence $O(1/T)$ under non-convex functions. Empirical studies demonstrate that FedTOGA outperforms state-of-the-art algorithms, with a 1\% accuracy increase and 30\% faster convergence, achieving state-of-the-art.

\end{abstract}

\section{Introduction}
\label{sec:intro}

\begin{figure}[]
\centering 
 \subcaptionbox{Dirichlet 0.1\label{1-a}}{\includegraphics[width=0.15\textwidth]{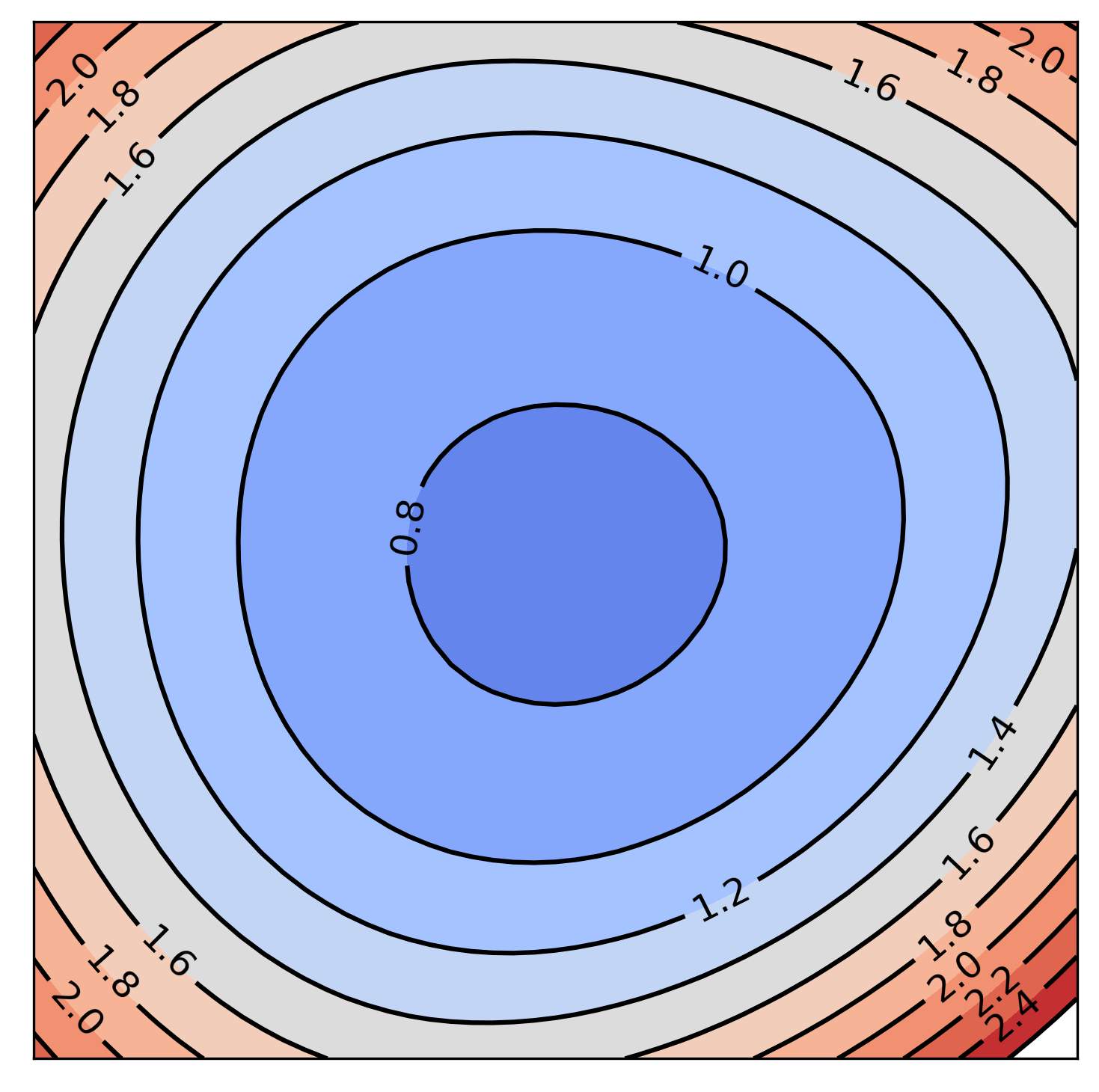}}
 \subcaptionbox{Dirichlet 0.6\label{1-b}}{\includegraphics[width=0.15\textwidth]{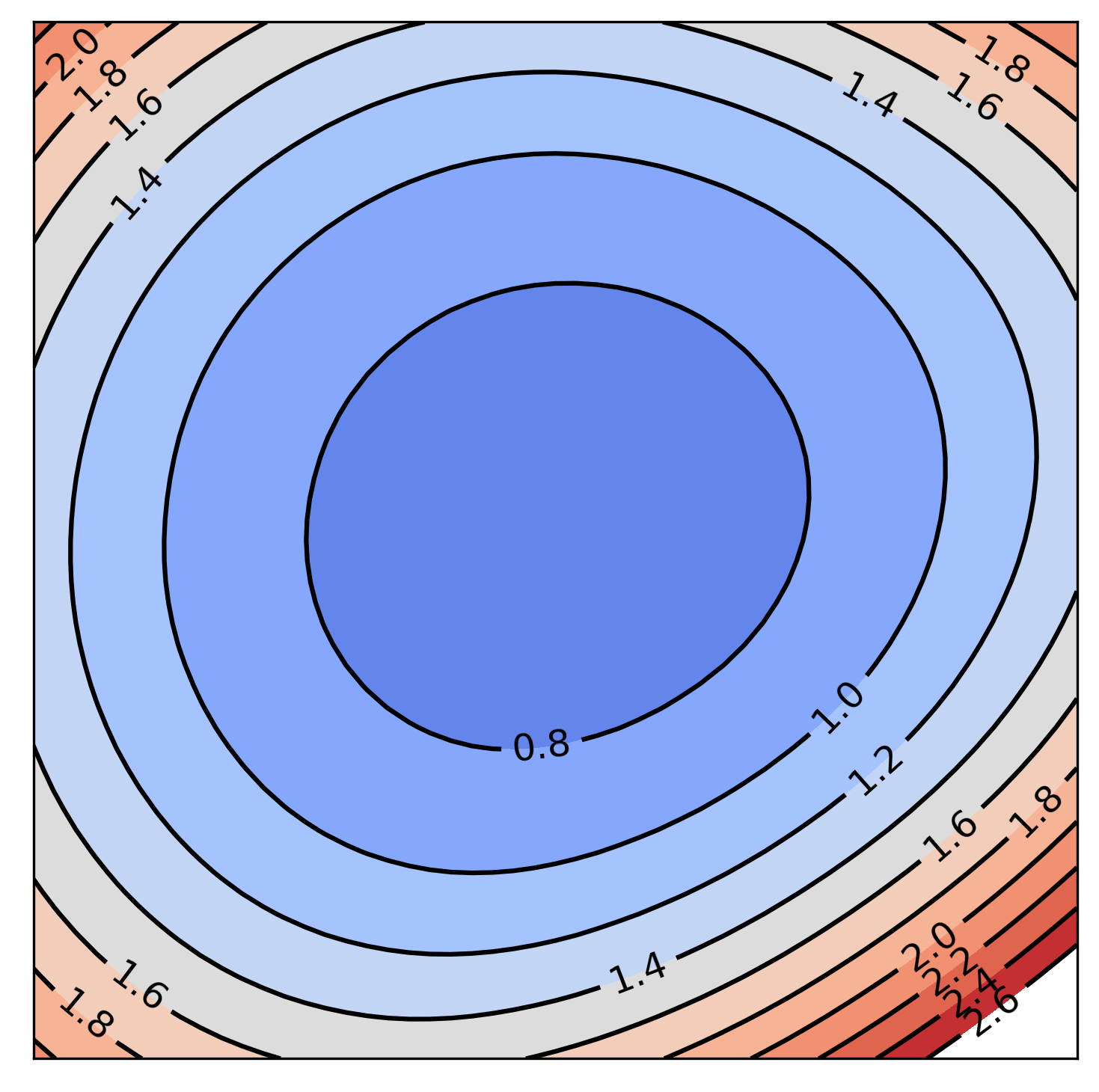}}
 \subcaptionbox{IID \label{1-c}}{\includegraphics[width=0.15\textwidth]{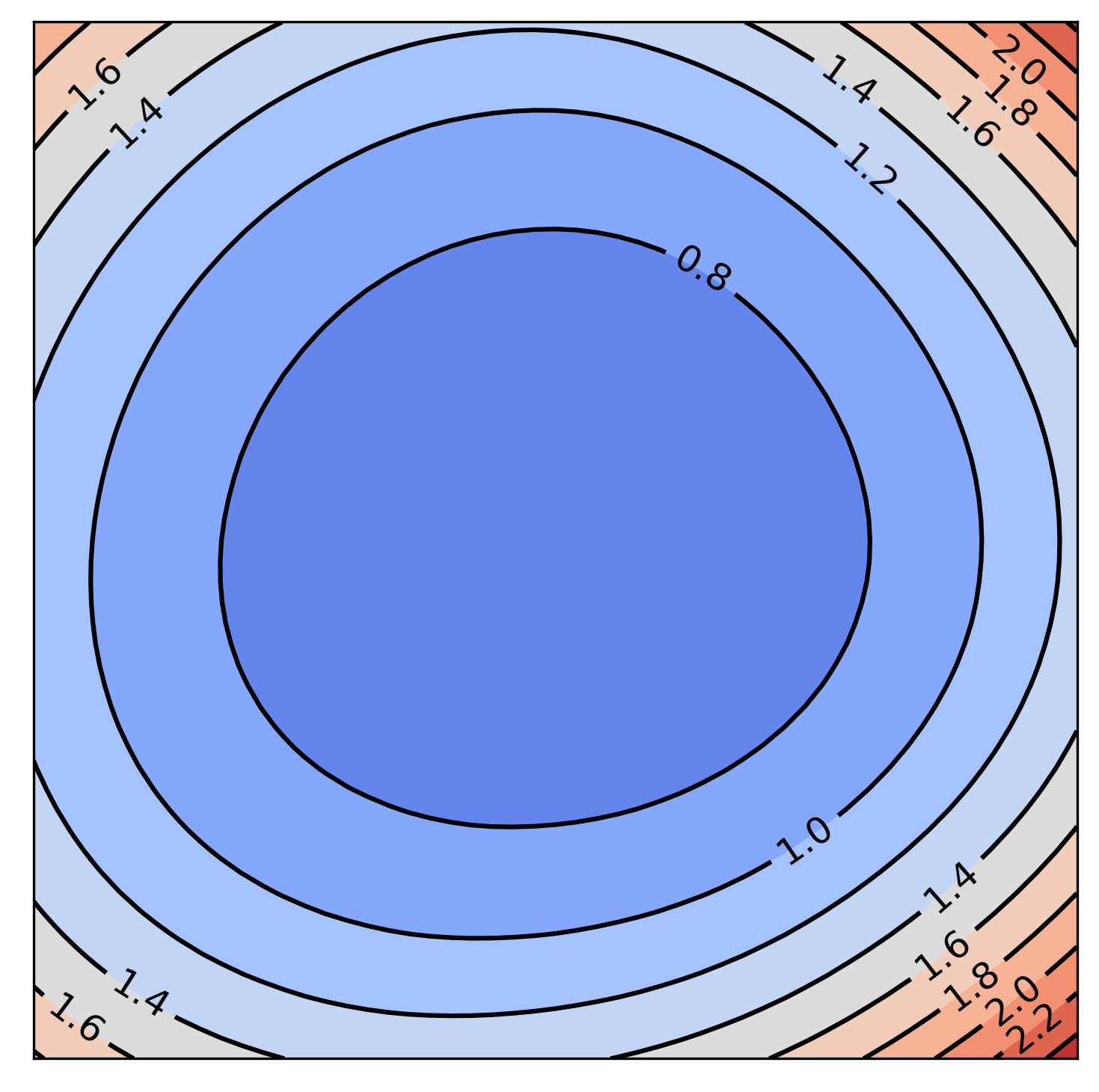}}

 \subcaptionbox{FedSAM\label{1-d}}{\includegraphics[width=0.15\textwidth]{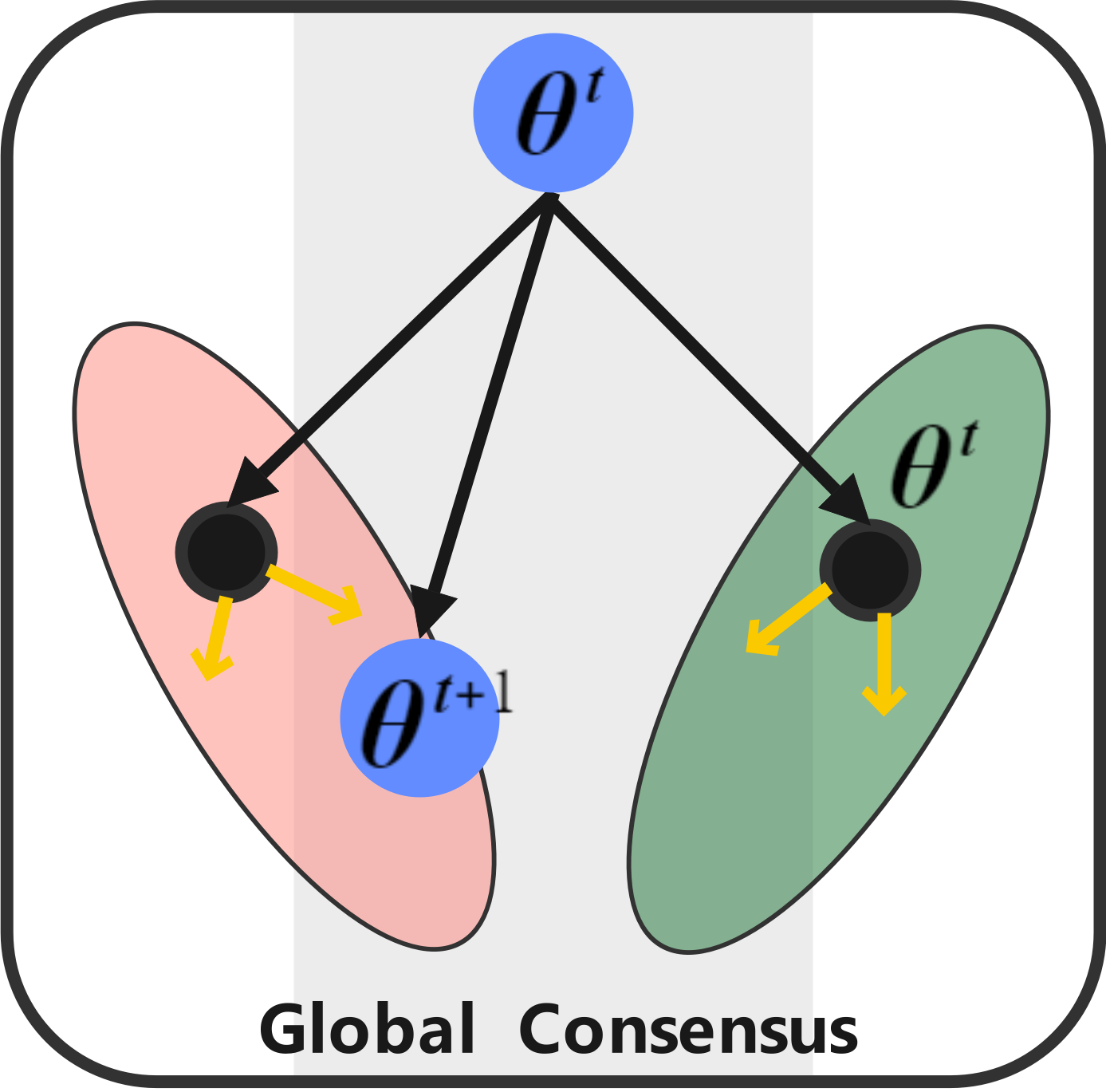}}
 \subcaptionbox{FedSMOO\label{1-e}}{\includegraphics[width=0.15\textwidth]{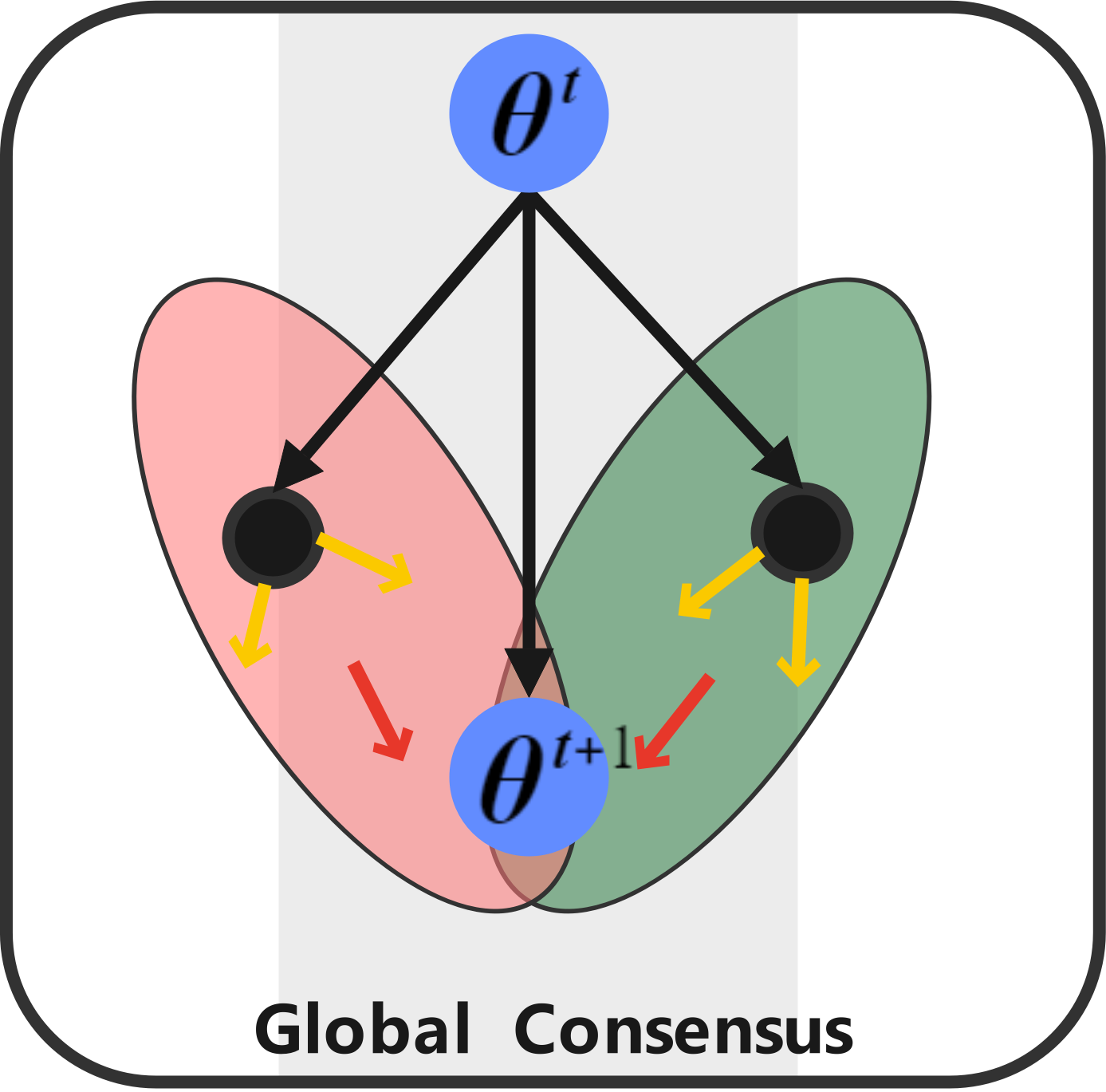}}
 \subcaptionbox{FedTOGA\label{1-f}}{\includegraphics[width=0.15\textwidth]{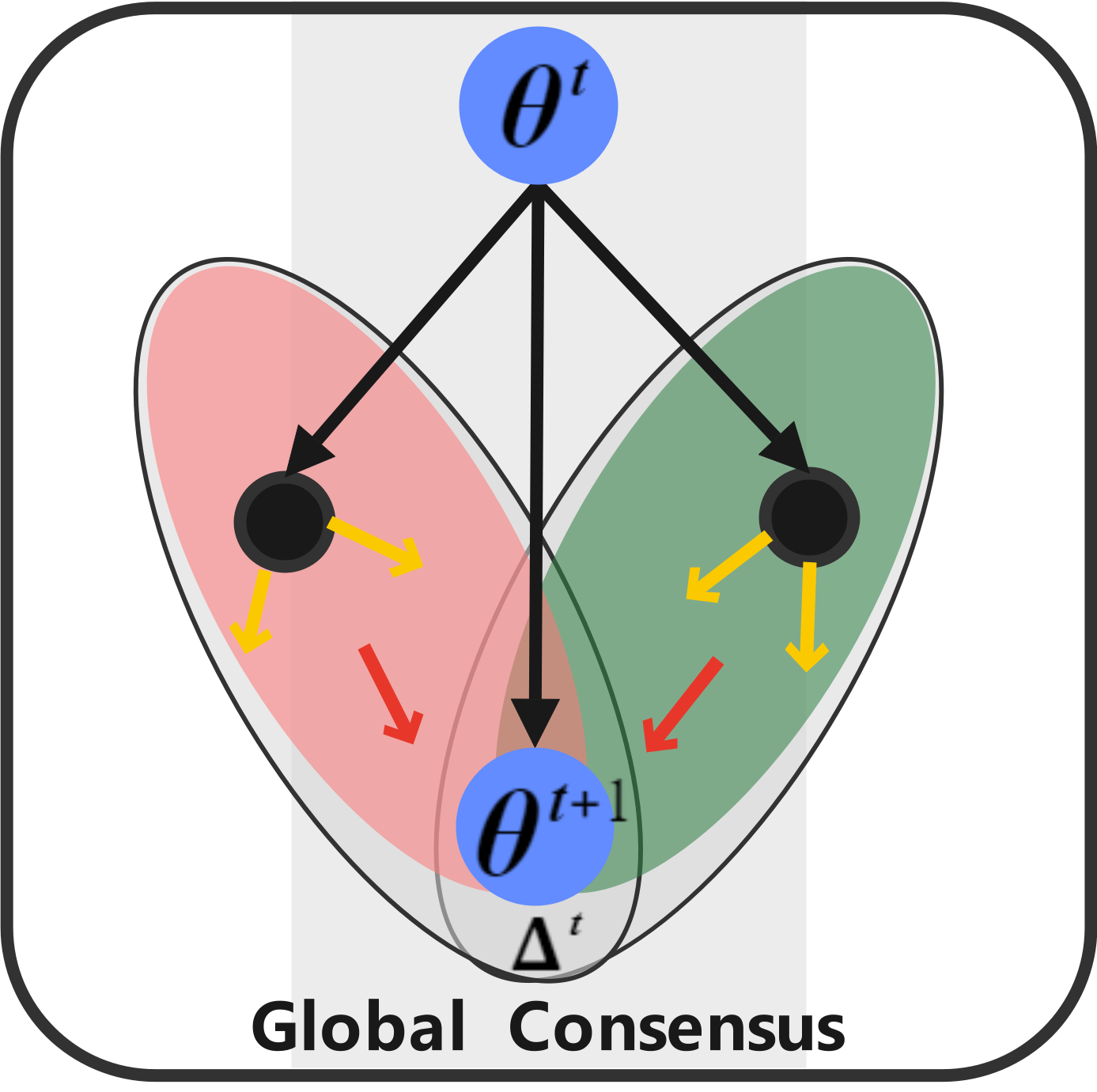}}
\caption{
Fig.(a)-(c) shows the loss surface under FL IID  as well as Non-IID setting, and Fig.(d)-(f) shows the FL system, where the gray color represents the global consensus while the \colorbox{red!60}{colored} \colorbox{green!60}{regions} represent the local knowledge. In Fig.(d), no further consensus can be increased in FL  only supported by the \colorbox{yellow!60}{SAM optimizer}. In Fig.(e), a \colorbox{red!60}{dynamic regularizer} is introduced in some work to increase global generalization. In Fig.(f), we further introduce \colorbox{gray!60}{Global Update} to extend the generalization.
}
\end{figure}

The widespread connectivity of mobile terminals has greatly propelled the development of industries related to big data. However, the massive data throughput has led to communication link congestion and increased privacy risks. Consequently, to safeguard data privatization and localization, FL \cite{DBLP:conf/aistats/McMahanMRHA17} has garnered significant attention as a distributed machine learning (ML) method that avoids the need for data exchange. FL employs the ``Computation-Then-Aggregation''(CTA) strategy \cite{Zhang2020FedPDAF}. Due to the uplink bandwidth constraints on global servers \cite{netSpeed}, FL adopts multiple local training steps and partial participation to alleviate communication bottlenecks. Simultaneously, the diverse data collection preferences of participants \cite{fan2022fedskipcombattingstatisticalheterogeneity,fan2024federatedlearningpartiallyclassdisjoint} result in conflicts between local optimization, which causes the global loss to converge to a steep local minimum \cite{woodworth2020localsgdbetterminibatch}. As illustrated in Figures \ref{1-a}-\ref{1-c}, the sharpness of the global model increases dramatically with the enhancement of local heterogeneity. Moreover, this disparity is significantly magnified by increasing synchronization intervals and reducing participation rates \cite{wang2020tacklingobjectiveinconsistencyproblem,li2020convergencefedavgnoniiddata}.

In response to these challenges, the majority of studies addressing global consistency issues via the Empirical Risk Minimization (ERM)\cite{malinovsky2020localsgdlocalfixedpoint}. However, when handling highly heterogeneous datasets, global solutions may become trapped in steep local minima, rendering it difficult to provide reliable estimates \cite{sun2023dynamic} and potentially causing the optimizer to stagnate. Consequently, recent innovations have leveraged Sharpness-Aware Minimization (SAM) \cite{foret2021sharpnessawareminimizationefficientlyimproving}, which seeks to identify a flatter minimum by minimizing the perturbed loss of the model, thereby enhancing generalization capabilities. FedSAM was introduced by incorporating SAM into FL \cite{pmlr-v162-qu22a}, and further, momentum-based algorithms were integrated, resulting in the proposal of MoFedSAM. FedGAMMA\cite{10269141} replaced ERM with SAM in Scaffold to enhance its performance and alleviate model bias. FedSpeed\cite{sun2023fedspeed} integrated FedDyn into FedSAM to bolster performance.
However, the approach based on minimizing local sharpness loss fails to capture the flatness of the global loss surface, as depicted in Fig. \ref{1-a}. Therefore, \cite{sun2023dynamic} proposed FedSMOO, which corrects local updates and local perturbations by introducing an additional control variable, as illustrated in Fig. \ref{1-e}. Furthermore, FedLESAM \cite{FedLESAM} estimates the global perturbation as the difference between the locally stored historical model from the activation round and the global model received in the current round, thereby avoiding extra computational costs. Nonetheless, FedSMOO introduces additional communication overhead and storage requirements, which can be unacceptable in real-world environments with limited bandwidth. In FedLESAM \cite{FedLESAM}, the global perturbation estimate does not encompass additional current local gradient ascent computations, while reducing computational overhead, may lead to insufficient local generalization. Additionally, in both methods, the difference in local perturbation scales may be significant when facing clients with prolonged disconnections, thereby disrupting generalization. Apart from addressing the consistency of perturbation generalization, the consistency of local objective optimization has also been deeply studied, as seen in ADMM-like algorithms \cite{Zhang2020FedPDAF,acar2021federated,sun2023dynamic}. However, performance significantly degrades, due to the CTA strategy causing local dual variables to be updated separately on the clients, preventing timely synchronization with global dual variables.


To achieve a reliable, stable, and consistent global model, we propose a novel algorithm named FedTOGA, as illustrated in Fig.\ref{1-f}. FedTOGA initially guides global update gradients to merge with local perturbations, thereby enhancing local generalization consistency. Simultaneously, it employs global updates to correct the local dynamic regularizer, reinforcing consistency with the global optimization objective. This approach significantly improves performance, even under extreme conditions characterized by highly heterogeneous data or limited client participation. 

Specifically, due to the communication interval, the universal SAM optimizer applied on the global server cannot precisely capture the perturbations occurring during local updates on client devices. To maintain local consistency, we introduce the global update gradient as an approximation, replacing the global perturbation estimation used in FedSMOO \cite{sun2023dynamic}. Furthermore, the universal dynamic regularizer on the global server is unable to accurately obtain the dual variable updates during client-side local updates. Hence, we propose leveraging the global update gradient to correct local dual variables, thereby mitigating discrepancies between local and global dual variables and further aligning the global and local objectives. This further aligns with the global and local objectives. To further reduce computational overhead, a method termed Neighborhood Gradient Perturbation has been proposed. When the interval between local training sessions on the client side exceeds one, the client simulates the current perturbation by utilizing cached gradients stored in a gradient register, thereby alleviating computational costs. Unlike FedCM\cite{xu2021fedcmfederatedlearningclientlevel} and MoFedSAM\cite{pmlr-v162-qu22a}, the global update is not treated as a trade-off term with the local perturbation, meaning that their coefficients do not sum to one. As the active local clients converge, they ultimately reach a globally stationary state, characterized by a smooth loss landscape.


Theoretically, FedTOGA is capable of achieving a rapid convergence rate of $O(1/T)$ in non-convex settings.   Experimentally, extensive evaluations were conducted on the CIFAR10/100 datasets, demonstrating that FedTOGA achieves faster convergence rates and higher generalization accuracy in practice.    These results were obtained in comparison to 17 baseline methods, including FedAvg, FedAdam, FedYogi, SCAFFOLD, FedACG, FedCM, FedDyn, FedDC, FedRCL, FedSAM, MoFedSAM, FedGAMMA, FedSMOO, FedSpeed, FedLESAM, FedLESAM-D, and FedLESAM-S.
\begin{itemize}
    \item 
    We propose a new federated algorithm, FedTOGA, the first global perturbation using the merged global gradient approximation, and the first local dynamic regularizer using global gradient correction, which reduces the uplink communication overhead and reduces the lags in extreme environments while maintaining fast convergence and high generalization.
    \item To alleviate the local computational overhead, we propose for the first time neighborhood perturbation, which is merged or replaced in the local perturbation using gradient registers without increasing the overhead, and analyze the advantages and limitations.
    \item 
    We give a theoretical analysis of the convergence speed. In non-convex scenarios, FedTOGA achieves fast O(1/T) convergence speed. Meanwhile, we conduct extensive numerical studies on the CIFAR-10/100 dataset as well as different neural networks to verify the excellent performance of FedTOGA, especially on highly heterogeneous as well as barren participants far better than the existing methods.
\end{itemize}

\section{Preliminaries}
This section shows the preliminaries of FL and SAM, and related works can be found in Appendix \ref{A-1}.

\noindent\textbf{Federated Learning} The goal of the FL framework is to build global models that minimize the average experience loss of participating clients:
 \begin{align}
       \mathop{\arg\min}_{\theta}  &f(\theta) = \frac{1}{N} \sum_{i \in N} f_i(\theta) \nonumber\\
       &  f_i(\theta) \triangleq \mathbbm{E}_{\xi_{i \sim D_i}}f_i(\theta,\xi_i)
 \end{align}
Where $f:\mathbbm{R} \rightarrow \mathbbm{R}^d$ is denoted as the global objective function, $\theta$ is a model parameter, $N$ is the total number of all the participating clients, and $\xi_i$ is a randomly sampled data from the distribution $D_i$ subject to data heterogeneity. $f_i$ is the loss function for the $i$-th client.

\noindent\textbf{Sharpness Aware Minimization} 
Many studies\cite{NIPS1994_01882513,dinh2017sharpminimageneralizedeep} have pointed out that a flat minimum implies a better generalization performance, which possesses greater robustness to model perturbations. To minimize sharpness, \cite{keskar2017largebatchtrainingdeeplearning, foret2021sharpnessawareminimizationefficientlyimproving} proposed SAM:
\begin{equation}
     \mathop{\arg\min}_{\theta} \big\{F(\theta) = \mathop{\arg\max}_{\|\delta\| \leq \rho} f(\theta + \delta)\big\}
\end{equation}
SAM extends the search by a one-step gradient ascent perturbation, and a one-step gradient descent to reduce sharpness and loss. First, calculate the gradient ascent perturbation $\delta = \rho \frac{\nabla f(\theta)}{\|\nabla f(\theta)\|}$. The gradient is then computed after adding the perturbation using the model and updating the model $\tilde{g} = \nabla f(\theta + \delta); \theta = \theta -\eta \tilde{g}$, where $\eta$ is the learning rate.
\section{Rethink FedSAM and Others}
The limitations of SAM in FL systems\cite{pmlr-v162-qu22a} have been widely discussed \cite{sun2023dynamic,FedLESAM,lee2024fedsolstabilizedorthogonallearning}, with the main conflict coming mainly from centralized training vs. Distributed Computing Perturbation Differences. The centralized SAM\cite{pmlr-v162-qu22a}training objectives are as follows.
\begin{equation}
   \max_{\|\delta\| \leq \rho} \mathbbm{E}_{\xi \sim D}f(\theta+\delta,\xi)  =  \max_{\|\delta\| \leq \rho} \mathbbm{E}_i \mathbbm{E}_{\xi_i \sim D_i}f(\theta+\delta,\xi_i) 
\end{equation}
where $D = \mathbbm{E}_iD_i$, some work applies SAM directly to the FL paradigm \cite{sun2023fedspeed,10269141}, and reformulates its goal as.
\begin{equation}
    \max_{\|\delta_i\| \leq \rho} \mathbbm{E}_i \mathbbm{E}_{\xi_i \sim D_i}f_i(\theta_i+\delta_i,\xi_i) 
\end{equation}
where the local model $\theta$, and the perturbation $\delta$ are isolated due to the CTA of FL. As a result, maintaining consistency between the global model and the client model becomes more difficult as the local update interval and the degree of data heterogeneity increase \cite{FedLESAM}. In this case, minimizing local sharpness in isolation does not effectively achieve a global flat minimum.


Some recent studies, FedSAM\cite{caldarola2022}, MoFedSAM\cite{pmlr-v162-qu22a}, FedGAMMA\cite{10269141}, FedSpeed\cite{sun2023fedspeed} have not resolved the internal perturbation variance contradiction.MoFedSAM uses momentum to weigh the perturbation gradient against the global gradient to alleviate this problem, FedGAMMA\cite{10269141} uses variance reduction techniques, and FedSpeed\cite{sun2023fedspeed} uses dynamic regularization techniques to alleviate this contradiction. FedSMOO\cite{ sun2023dynamic} notices this contradiction for the first time and will use dynamic regularization to correct the discrepancy between local and global perturbations, FedSOL\cite{lee2024fedsolstabilizedorthogonallearning} employs perturbation orthogonality to find a consistent direction of perturbation, FedLESAM\cite{FedLESAM} believes that computing the perturbations requires additional computation and therefore opens up additional storage locally to approximate the estimated global perturbations. For a more detailed description of the limitations, please refer to Appendix \ref{A-2}.
\subsection{Motivation}
Obviously, the above study has some limitations. To increase the consistency of the global perturbations, FedSMOO\cite{sun2023dynamic} attempts to solve this problem by correcting for the local perturbations, whose local perturbation estimates are computed as $\delta = \rho \frac{\nabla f_i(\theta) - \mu_i -s}{\|\nabla f_i(\theta) - \mu_i -s\|}$. However, it introduces additional computation, which increases the overhead of the clients in FL. To efficiently optimize the global sharpness and reduce the computational burden on the client, FedLESAM\cite{FedLESAM} proposes a strategy of history model staging, which enables the client to add additional local variables to record the history model, thus estimating the global perturbation $\delta = \rho \frac{\theta_i^{old} -\theta_i^{t}}{\|\theta_i^{old} - \theta_i^{t}\|}$. However, as the set of activated clients $S_t$ decreases sharply, the perturbation estimation is more affected. FedTOGA estimates the global perturbation by merging in the local perturbation through receiving the proximity global update sent from the server, which does not require additional computation and storage space and ensures the timeliness of the estimation, and at the same time incorporating this global update into the dynamic regularizer \cite{ acar2021federated} to continuously find the global minimization.
\section{Methodology}
\label{sec:methodology}
This section introduces our method, FedTOGA, and provides a preliminary analysis of the three key techniques and demonstrates the flow of the proposed algorithm \ref{algorithm_toga}.
\begin{algorithm}[]
\small
\caption{FedTOGA Algorithm}
\label{algorithm_toga}
Initial model parameters $\theta^0$, initial global update $\Delta^{-1}$, local dual variable $h_i$, global dual variable $h$, local perturbation gradient $\tilde{g}_{i,-1}$, total communication rounds $T$, penalized coefficient for the quadratic term  $\alpha$, Correction coefficient for perturbation and dual term $\kappa, \beta$ \\
\textbf{Server execute:}\\
\For{each round $t \in [T] \triangleq \{0, 1, 2, \cdots, T-1 \}$}{
     Sample the active client set $S_t \subseteq [N]$.\\
    \For{$i \in S_t$ in parallel}{
      ${\theta}_i^{t+1} \leftarrow \textbf{Client Update}(\theta^{t},$\colorbox{cvprblue!60}{$\Delta^{t} $})\;
    \textbf{ communicate} ${\theta}_i^t$ to server \;    
        }

 $h^{t+1} = h^{t} - \frac{1}{\alpha M}\sum_{i \in S_t} ({\theta}_i^{t+1} - \theta^t); $\\
 \colorbox{cvprblue!60}{$\Delta^{t+1} = -\frac{1}{MK} \sum_{i \in S_t} ({\theta}_i^{t+1} - {\theta}^t)$}; \\ 
 $ \theta^{t+1} =  \frac{1}{M}\sum_{i \in  S_t} {\theta}_i^{t+1} - \alpha h^{t+1}$ 

}
\textbf{Client Update($ \theta_{t},\Delta_{t}$):} $\theta_{i,0}^t = \theta^{t}$\\
\For{local epoch $k \in [K] \triangleq \{0,1, 2, \cdots, K-1 \}$}{
    sample a  mini-batch data $\xi_{i,k}^t$;\\
   gradient estimate: $g_{i,k}^t =\tilde{\nabla} f_i(\theta_{i,k}^t;\xi_{i,k}^t)$\\
    Perturbation: \colorbox{cvprblue!60}{$\delta_{i,k}^t = \rho \frac{g_{i,k}^t[\tilde{g}_{i,k-1}^t]+\kappa \Delta^{t}}{\| g_{i,k}^t[\tilde{g}_{i,k-1}^t]+\kappa \Delta^{t}\|}$}\\
    extra-step:
    $\tilde{g}_{i,k}^t = \nabla f_i(\theta_{i,k}^{t} +\delta_{i,k}^t; \xi_{i,k}^t);\theta_{i,k+1}^t =$\\
    $ \theta_{i,k}^{t}  - \eta_l\big(\tilde{g}_{i,k}^t -h_i^{t} +\frac{1}{\alpha}\left( \theta_{i,k}^{t} - \theta_{i,0}^t\right) $+ \colorbox{cvprblue!60}{$\beta\Delta^{t}$}$\big)$
}
$ h_i^{t+1} = h_i^{t} - \frac{1}{\alpha}\left( \theta_{i,K}^{t} - \theta_{i,0}^t \right)$\\
\textbf{return} ${\theta}_i^{t+1} = \theta_{i,K}^{t} $\\

\end{algorithm}
\subsection{Estimate Global Perturbation}
\label{Sec.4.1}
As mentioned above, our goal is to efficiently estimate the global perturbations(G-perturbations) of each client without incurring additional storage or computational overhead. To achieve this, we first recall the definition of sharpness-aware minimization in FL: 
\begin{equation}
    \min_{\theta}\big\{f = \frac{1}{N} \sum_{i \in N} \max_{\|\delta_i\| \leq \rho} \mathbbm{E}_i \mathbbm{E}_{\xi_i \sim D_i}f_i(\theta_i+\delta_i,\xi_i) \big\}
\end{equation}
Therefore, we can obtain that at $t$ round, $k$ moments, the virtual global perturbation variable $\delta^t_{k} = \rho\frac{\nabla f(\theta^t)}{\|\nabla f(\theta^t)\|} =\rho \frac{\sum_{i \in N}\nabla f_i(\theta^t_k)}{\|\sum_{i \in N}\nabla f_i(\theta^t_k)\|}\simeq \rho \frac{\sum_{i \in S}\nabla f_i(\theta^t_k)}{\|\sum_{i \in S}\nabla f_i(\theta^t_k)\|}$. The $\theta_k^t$ denotes the global model at virtual moment $k$, which is computed as $\theta^t_k = \frac{1}{M}\sum_{i \in S_t}\theta^t_{i,k}$.
However, due to the CTA strategy in the FL paradigm, the set of clients does not have effective access to the global model $\theta_k^t$ at each moment in time, and thus the global perturbation $\delta$ cannot be computed correctly. Inspired by the FedCM\cite{xu2021fedcmfederatedlearningclientlevel}, strategy, we estimate the global update $\Delta^t \simeq \nabla f(\theta^t)$ by passing the global update variable. Finally, we define the update strategy for the global perturbation SAM of FedTOGA as follows:
$\delta_{k}^t = \rho\frac{\nabla f(\theta^t)}{\|\nabla f(\theta^t)\|}\simeq  \rho \frac{g_{i,k}^t+\kappa \Delta^{t}}{\|g_{i,k}^t+\kappa \Delta^{t}\|}; \theta_{i,k}^t = \theta_{i,k-1}^t -\eta_l \nabla F_i( \theta_{i,k}^t + \rho \delta_k^t)$. The differences between the FedTOGA perturbation strategy and the rest of the similar works can be viewed in Appendix \ref{A-2} Tab.\ref{tab-abs-p}.

\subsection{Utilize Neighbourhood Perturbation}
Besides, as stated by FedLESAM\cite{FedLESAM}, local perturbations require additional gradient ascent computation, which may consume additional computational overhead. Therefore, how to estimate the local perturbation without utilizing additional computation? We propose neighborhood perturbation(N-perturbation) for the first time. Specifically, when the client's local iteration interval exceeds one, the local perturbation gradient $\tilde{g}_{i,k-1}^t$ will be recorded by the cache without opening additional storage space. We can get $g_{i,k} \simeq \tilde{g}_{i,k-1}^t$. We can further replace the perturbation term in the local SAM optimization and get: $\delta_{i,k}^t = \rho \frac{\tilde{g}_{i,k-1}^t+\kappa \Delta^{t}}{\| \tilde{g}_{i,k-1}^t+\kappa \Delta^{t}\|}$. This operation allows approximate estimation of local perturbations in environments where the client-side resource is scarce. 

\textbf{Perturbation Fusion?}
In the FL paradigm, the edge client SAM only captures the sharpness of a specific small batch of data, which is mitigated effect by the G-Perturbation technique described above to enhance generalization. Let's further think about whether N-Perturbation may bring additional benefits in addition to alleviating computational overhead. Similar to LookAhead\cite{zhang2019lookaheadoptimizerksteps}, it backtracks by perturbing ascent after each gradient descent. Then, our perturbation calculation can be rewritten:
$\delta_{i,k}^t = \rho \frac{g_{i,k}+\tilde{g}_{i,k-1}^t+\kappa \Delta^{t}}{\| g_{i,k}+\tilde{g}_{i,k-1}^t+\kappa \Delta^{t}\|}$.
A more in-depth discussion of N-perturbation can be found in Appendix \ref{A-4}.
\subsection{Global Correction in Dynamic Regularizer}
In order to effectively avoid performance degradation and further improve the optimization objective consistency, we also adopt dynamic regularization\cite{acar2021federated} that merges the global update $\Delta$ correction on each local client, which takes the form of an ADMM-like method in order to effectively minimize the global objective $f$. This is the first FL framework that considers local dual variable corrections. First, we consider a centralized global Augmented Lagrangian(AL) function $f_{cen}$ for which the correction introduces a penalty term $\theta = \theta_i$ constraint as:
\begin{small}
    \begin{equation}
    \label{eq_f_dyn_cen}
    f_{cen}:\frac{1}{N}\sum_{i \in N}\left\{f_i + \left\langle h^t, \theta^t -\theta_i^t\right\rangle + \frac{1}{2\alpha}\|\theta^t - \theta^t_i\|^2\right\}
\end{equation}
\end{small}
In the case of centralized learning, the dual variables are updated promptly, via $h^t = h^{t-1} - \frac{1}{\alpha N}\sum_i(\theta^t_i - \theta^{t-1} )$. However, FL's CTA policy makes the dual variables forced to decompose to update locally as follows:
\begin{small}
\begin{equation}\label{eq_f_dyn_fed}
    f_{fed}:\frac{1}{N}\sum_{i \in N}\left\{f_i + \left\langle h_i^t, \theta^t -\theta_i^t\right\rangle + \frac{1}{2\alpha}\|\theta^t - \theta^t_i\|^2\right\}
\end{equation}
\end{small}
where the global dual variable $h$ is updated at each communication and the local dual variable $h_i$ is locally staged. Thus dual variable differences are amplified as the local interval expands, which has not been addressed in previous studies\cite{acar2021federated,sun2023dynamic}. Thus, similar to the strategy for correcting global perturbations (in Sec. \ref{Sec.4.1}), we employ proximity global update correction to mitigate the difference between the local dual variable and the global dual variable. Specifically, we cause the local dual variables to approximate the global dual variables by adding corrections and obtaining $h_i - \beta \Delta^t \simeq h$. Locally in each subproblem, we solve the local model $\theta_i$ by minimizing the corrected AL function:
\begin{small}
\begin{equation}
\label{eq_f_dyn_clnt}
    \theta_{i,K}^t = \mathop{\min}_{\theta_i}\left\{f_i - \left\langle h_i^t - \beta\Delta^t, \theta^t_i\right\rangle + \frac{1}{2\alpha}\|\theta^t - \theta^t_i\|^2\right\}
\end{equation}
\end{small}
Again, in order not to affect the original SAM we use SGD to solve this problem \cite{sun2023dynamic}. We then update the dual variables locally $h_i^{t+1} = h_i^{t} - \frac{1}{\alpha}\left( \theta_{i,K}^{t} - \theta_{i,0}^t \right)$. After finishing the local training, update $\theta^t$ to $\theta^{t+1}$ by solving the equation \ref{eq_f_dyn_fed} and start the next iteration.

\subsection{Overview of FedTOGA}
Algorithm\ref{algorithm_toga} shows the detailed flow of FedTOGA. First initialize the server-side global model $\theta$. In the global synchronization round $t$, a set $S_t$ containing $M$ clients is randomly selected from all clients $N$, and the global model $\theta_t$ is sent to the set of authorized clients $S_t$ with the global update $\Delta^t$ of the $t-1$ round. The client first computes the original gradient $g_{i,k}^t$ according to line 16 of the algorithm, and subsequently, computes the SAM gradient $\delta_{i,k}^t$ corrected by $\Delta^t$ in line 17, with the neighborhood perturbation variable $\tilde{g_i}$ being optional. In line 19, we use the formula \ref{eq_f_dyn_clnt} for local dual variable correction to update the local model $\theta_i$, followed by updating the local dual  variable $h_i$ via line 21. After local training is complete, FedTOGA sends only $\theta_i^t$ to the server for aggregation. In lines 9-11 of the algorithm, the server updates the global model from $\theta^t$ to $\theta^{t+1}$ by minimizing the formula \ref{eq_f_dyn_fed}. This process is repeated until $T-1$.

\section{Theoretical Analysis}

\begin{assumption}
\label{assumption_1}
The loss function  $f_i$ is $L$-Smooth, i.e.,$f_i(y) - f_i(x) \leq \langle \nabla f_i(x),y-x\rangle + \frac{L}{2}\|y-x\|^2.$
\end{assumption}
\begin{assumption}
\label{assumption_2}
Unbiased and bounded variance of stochastic gradient.  The stochastic gradient $\tilde{\nabla} f_i(x) = \nabla f_i(x, \xi_i)$ computed by the $i$-th client using mini-batch $\xi$ is an unbiased estimator of $\nabla f_i(x)$, i.e.
$\mathbb{E}[\tilde{\nabla} f_i(x)] =\nabla f_i(x), \mathbb{E}{\| \tilde{\nabla} f_i(x)- \nabla  f_i( x)\|}^2 \leq \sigma^2_l.$
\end{assumption}
\begin{assumption}
\label{assumption_3}
Bounded Heterogeneity, for all $x \in \mathbb{R}^d$, we establish the following inequality:$\mathbbm{E}\|\nabla f_i(x) - \nabla f(x)\| \leq \sigma_g$
Besides, the variance of unit gradient is bounded:
$\mathbbm{E}\|\frac{\nabla f_i(x)}{\|\nabla f_i(x)\|} - \frac{\nabla f(x)}{\|\nabla f(x)\|} \|\leq \sigma_{g}'.$
\end{assumption}
\begin{theorem}
\label{theorem_1}
    Under Assumption \ref{assumption_1}-\ref{assumption_3}, For any training interval $t$ on $i$-th client, model divergence satisfies:
   \begin{align}
     \|\theta_{i,k}^t - v_{k}^t\|^2 \leq  H_i(k) 
   \end{align}
where $H_i(\tau) \leq \frac{L^2\rho^2\sigma_g^{'2}+\sigma_g^2}{2L^2}((1+2\eta_l^2L^2)^{\tau}-1)$, $\{v^t\}$is a virtual sequence representing the global model.
More Details can be referred to the Appendix.
\end{theorem}

\begin{remark}
 The difference between the local model and the global model will be geometrically amplified as the local interval expands, mainly from the model perturbation error and update error, and thus it is reasonable to enhance the   consistency 
 of optimization and generalization objective (in Sec. \ref{sec:methodology}).
\end{remark}
\begin{theorem}
\label{theorem_2}
Under Assumption \ref{assumption_1}-\ref{assumption_3}. When $\eta_l \leq \min\{\frac{c_1}{\sqrt{1008L^2K + 72 L^2 K\beta^2}},2\alpha\}$, $c_1 = \sqrt{1/2 - L\alpha \beta^2} > 0$ and the perturbation learning rate satisfies $\rho = O(1/\sqrt{T})$, and local interval $K >\frac{\alpha}{\eta_l}$, let $\Gamma =\frac{1}{2} -1008\eta_l^2L^2K - 72\eta_l^2L^2K\beta^2 -L\alpha\beta^2$ is a positive constant with select the suitable $\eta_l$, the auxiliary sequence $\{z^t\}$ generated by executing the Algorithm \ref{algorithm_toga} satisfies:
    \begin{align}
    \frac{1}{T}\sum_{t=0}^{T-1} \leq& \frac{1}{T\alpha \Gamma }\left(\nabla f(z^0) - f^*\right)  +\frac{16\alpha^3L^2}{T\Gamma} \mathbbm{E}_t\left\|\frac{1}{N} \sum_{i \in N } h_i^{0}\right\|^2 \nonumber\\ 
    & +\frac{144\alpha L^2\eta_l^2K}{ T  N \Gamma}\sum_{i \in N}\mathbbm{E}\|h_i^0\|^2 +\Upsilon
\end{align}
where the $f^*$ is the optimal of non-convex function $f$, and the term $\Upsilon$ is: 
$$\Upsilon =  \frac{1}{\alpha \Gamma} \left(72\eta_l^2L^2K(24\sigma_l^2 + 2\sigma_l^2 +6L^2\rho^2)+ 3\alpha L^2\rho^2\right)$$
More Details can be referred to the Appendix.
\end{theorem}
\begin{remark}
  When we set the local learning rate $\eta_l$ to satisfy $\eta_l = O(1/K)$, $\eta_l \geq \alpha/K$, and the perturbation learning rate to be $O(1/T)$, FedTOGA can achieve a fast convergence rate of $O(1/T)$ when the local interval $K$ satisfies $K = O(T)$.
\end{remark}
\begin{remark}
  Inspired by the FedSpeed\cite{sun2023fedspeed}, the FedTOGA can speed up convergence by increasing the setting of the local interval $K$, which is useful for bandwidth-constrained FL systems. However, the local perturbation learning rate $\rho$ in FedSpeed restricts the upper bound, and our proof slightly relaxes the limitation so that the perturbation learning rate just satisfies $O(1/T)$.
\end{remark}

\begin{table*}[]
\centering
\caption{Dirichlet coefficients $u$ selected from $\{0.1,0.6\}$, and $c$ is the Pathological coefficient, i.e., the number of active categories in each client. The two datasets have 100 clients in the upper part with 10\% active in each round, 200 clients in the lower part with 5\% active in each round.(LeNet)}
\label{tab_exp_lenet}
\scalebox{0.85}{
\begin{tabular}{l|cccc|cccc}
{Method}&\multicolumn{4}{c|}{CIFAR10}&\multicolumn{4}{c}{CIFAR100}\\ \cline{2-9}
 Partition&\multicolumn{2}{c}{Dirichlet}&\multicolumn{2}{c|}{Pathological}&\multicolumn{2}{c}{Dirichlet}&\multicolumn{2}{c}{Pathological}\\
Coefficient&{$u=0.6$}&{$u=0.1$}&{$c=6$}&{$c=3$}&{$u=0.6$}&{$u=0.1$}&{$c=20$}&{$c=10$}\\\hline
FedAvg \cite{DBLP:conf/aistats/McMahanMRHA17}  &$80.28^{\pm 0.14}$&$74.68^{\pm 0.19}$&$80.59^{\pm 0.18}$&$78.10^{\pm 0.23}$&$47.35^{\pm 0.16}$&$45.56^{\pm 0.20}$&$46.46^{\pm 0.20}$&$43.43^{\pm 0.27}$\\
FedAdam&$80.39^{\pm 0.17}$&$71.52^{\pm 0.29}$&$81.02^{\pm 0.20}$&$77.88^{\pm 0.23}$&$48.94^{\pm 0.21}$&$43.62^{\pm 0.25}$&$44.86^{\pm 0.25}$&$41.58^{\pm 0.27}$\\
FedYogi \cite{reddi2021adaptivefederatedoptimization} &$80.11^{\pm 0.19}$&$73.58^{\pm 0.25}$&$81.08^{\pm 0.21}$&$78.10^{\pm 0.20}$&$48.41^{\pm 0.21}$&$45.44^{\pm 0.22}$&$46.18^{\pm 0.22}$&$42.07^{\pm 0.25}$\\
SCAFFOLD\cite{pmlr-v119-karimireddy20a}  &$82.87^{\pm 0.12}$&$78.00^{\pm 0.16}$&$83.31^{\pm 0.10}$&$80.29^{\pm 0.15}$&$53.68^{\pm 0.21}$&$50.33^{\pm 0.24}$&$51.30^{\pm 0.22}$&$47.71^{\pm 0.22}$\\
FedACG\cite{kim2024communication}  &$82.87^{\pm 0.14}$&$77.51^{\pm 0.16}$&$82.86^{\pm 0.12}$&$80.84^{\pm 0.17}$&$52.88^{\pm 0.20}$&$48.72^{\pm 0.23}$&$50.24^{\pm 0.21}$&$46.08^{\pm 0.24}$\\
FedCM\cite{xu2021fedcmfederatedlearningclientlevel}  &$77.04^{\pm 0.30}$&$62.75^{\pm 0.31}$&$66.58^{\pm 0.29}$&$71.20^{\pm 0.33}$&$43.08^{\pm 0.19}$&$34.69^{\pm 0.26}$&$36.27^{\pm 0.18}$&$28.48^{\pm 0.30}$\\
FedDyn\cite{acar2021federated}  &$82.31^{\pm 0.13}$&$78.05^{\pm 0.19}$&$83.13^{\pm 0.18}$&$79.96^{\pm 0.19}$&$49.97^{\pm 0.19}$&$45.85^{\pm 0.29}$&$47.41^{\pm 0.21}$&$43.29^{\pm 0.19}$\\
FedDC\cite{gao2022federated}  &$83.58^{\pm 0.14}$&$78.50^{\pm 0.19}$&$84.00^{\pm 0.16}$&$81.72^{\pm 0.17}$&$51.99^{\pm 0.15}$&$48.75^{\pm 0.21}$&$49.53^{\pm 0.19}$&$44.82^{\pm 0.23}$\\
FedRCL \cite{seo2024relaxed}&$77.62^{\pm 0.11}$&$68.79^{\pm 0.16}$&$78.28^{\pm 0.15}$&$76.04^{\pm 0.19}$&$46.34^{\pm 0.24}$&$42.28^{\pm 0.17}$&$44.06^{\pm 0.19}$&$39.64^{\pm 0.21}$\\
FedSAM &$81.58^{\pm 0.15}$&$77.67^{\pm 0.15}$&$82.15^{\pm 0.17}$&$79.23^{\pm 0.23}$&$48.08^{\pm 0.21}$&$46.86^{\pm 0.26}$&$46.71^{\pm 0.25}$&$43.41^{\pm 0.22}$\\
MoFedSAM \cite{pmlr-v162-qu22a} &$77.17^{\pm 0.12}$&$66.24^{\pm 0.15}$&$77.44^{\pm 0.15}$&$72.15^{\pm 0.19}$&$43.30^{\pm 0.18}$&$34.43^{\pm 0.21}$&$36.50^{\pm 0.19}$&$29.92^{\pm 0.24}$\\
FedGAMMA\cite{10269141}&$83.88^{\pm 0.13}$&$78.61^{\pm 0.15}$&$83.79^{\pm 0.14}$&$79.68^{\pm 0.15}$&$53.94^{\pm 0.20}$&$49.95^{\pm 0.24}$&$51.20^{\pm 0.22}$&$48.11^{\pm 0.29}$\\
FedSMOO \cite{sun2023dynamic} &\textbf{84.82}$^{\pm 0.15}$&\textbf{80.06}$^{\pm 0.16}$&\textbf{85.07}$^{\pm 0.17}$&$81.26^{\pm 0.19}$&\textbf{56.57}$^{\pm 0.18}$&$52.17^{\pm 0.17}$&$53.42^{\pm 0.21}$&$48.12^{\pm 0.19}$\\
FedSpeed \cite{sun2023fedspeed} &$84.14^{\pm 0.15}$&$80.16^{\pm 0.16}$&$84.74^{\pm 0.14}$&\textbf{82.20}$^{\pm 0.19}$&$53.96^{\pm 0.19}$&\textbf{52.29}$^{\pm 0.21}$&{53.78}$^{\pm 0.18}$&\textbf{48.33}$^{\pm 0.20}$\\
FedLESAM &$80.94^{\pm 0.18}$&$77.02^{\pm 0.15}$&$81.79^{\pm 0.18}$&$78.85^{\pm 0.15}$&$48.13^{\pm 0.18}$&$46.55^{\pm 0.21}$&$46.08^{\pm 0.23}$&$43.57^{\pm 0.17}$\\
FedLESAM-D&$83.28^{\pm 0.15}$&$79.12^{\pm 0.18}$&$84.20^{\pm 0.19}$&$80.91^{\pm 0.16}$&$54.88^{\pm 0.18}$&$52.08^{\pm 0.22}$&\textbf{54.14}$^{\pm 0.19}$&$48.28^{\pm 0.22}$\\
FedLESAM-S\cite{FedLESAM}&$83.39^{\pm 0.12}$&$78.23^{\pm 0.17}$&$83.99^{\pm 0.19}$&$81.20^{\pm 0.15}$&$53.29^{\pm 0.15}$&$50.12^{\pm 0.21}$&$52.20^{\pm 0.20}$&$47.29^{\pm 0.17}$\\
FedTOGA(ours)&\textcolor{cvprblue}{\textbf{86.01}$^{\pm 0.12}$}&\textcolor{cvprblue}{\textbf{82.05}$^{\pm 0.11}$}&\textcolor{cvprblue}{\textbf{85.71}$^{\pm 0.13}$}&\textcolor{cvprblue}{\textbf{84.00}$^{\pm 0.12}$}&\textcolor{cvprblue}{\textbf{57.25}$^{\pm 0.13}$}&\textcolor{cvprblue}{\textbf{53.45}$^{\pm 0.13}$}&\textcolor{cvprblue}{\textbf{55.49}$^{\pm 0.13}$}&\textcolor{cvprblue}{\textbf{51.27}$^{\pm 0.18}$}\\\hline
FedAvg \cite{DBLP:conf/aistats/McMahanMRHA17}  &$77.53^{\pm 0.17}$&$74.60^{\pm 0.23}$&$79.21^{\pm 0.25}$&$76.20^{\pm 0.23}$&$43.86^{\pm 0.21}$&$42.70^{\pm 0.24}$&$42.94^{\pm 0.25}$&$42.28^{\pm 0.29}$\\
FedAdam&$79.39^{\pm 0.19}$&$74.49^{\pm 0.31}$&$79.53^{\pm 0.23}$&$76.09^{\pm 0.25}$&$45.34^{\pm 0.25}$&$42.79^{\pm 0.23}$&$43.57^{\pm 0.25}$&$40.66^{\pm 0.29}$\\
FedYogi \cite{reddi2021adaptivefederatedoptimization} &$79.95^{\pm 0.21}$&$75.29^{\pm 0.25}$&$79.73^{\pm 0.22}$&$77.64^{\pm 0.23}$&$46.67^{\pm 0.25}$&$43.02^{\pm 0.24}$&$44.70^{\pm 0.27}$&$41.33^{\pm 0.30}$\\
SCAFFOLD\cite{pmlr-v119-karimireddy20a}  &$81.18^{\pm 0.15}$&$76.11^{\pm 0.19}$&$82.44^{\pm 0.17}$&$78.52^{\pm 0.17}$&$51.45^{\pm 0.25}$&$47.19^{\pm 0.27}$&$48.26^{\pm 0.28}$&$46.82^{\pm 0.26}$\\
FedACG\cite{kim2024communication}  &$82.57^{\pm 0.17}$&$78.47^{\pm 0.20}$&$82.09^{\pm 0.16}$&$80.50^{\pm 0.19}$&$51.96^{\pm 0.24}$&$49.34^{\pm 0.26}$&$50.01^{\pm 0.27}$&$46.82^{\pm 0.25}$\\
FedCM\cite{xu2021fedcmfederatedlearningclientlevel}  &$76.08^{\pm 0.30}$&$64.33^{\pm 0.31}$&$76.64^{\pm 0.29}$&$68.61^{\pm 0.33}$&$40.32^{\pm 0.19}$&$33.05^{\pm 0.26}$&$34.19^{\pm 0.18}$&$27.88^{\pm 0.30}$\\
FedDyn\cite{acar2021federated}  &$80.60^{\pm 0.17}$&$77.53^{\pm 0.21}$&$81.54^{\pm 0.22}$&$79.39^{\pm 0.24}$&$48.40^{\pm 0.20}$&$45.04^{\pm 0.31}$&$46.87^{\pm 0.24}$&$43.04^{\pm 0.29}$\\
FedDC\cite{gao2022federated}  &$81.83^{\pm 0.17}$&$78.87^{\pm 0.21}$&$82.44^{\pm 0.17}$&$80.93^{\pm 0.19}$&$48.74^{\pm 0.19}$&$45.11^{\pm 0.26}$&$45.94^{\pm 0.22}$&$43.94^{\pm 0.27}$\\
FedRCL \cite{seo2024relaxed}&$76.06^{\pm 0.15}$&$66.88^{\pm 0.19}$&$76.51^{\pm 0.19}$&$72.28^{\pm 0.23}$&$42.05^{\pm 0.27}$&$38.60^{\pm 0.20}$&$40.56^{\pm 0.24}$&$37.28^{\pm 0.26}$\\
FedSAM &$79.74^{\pm 0.18}$&$74.69^{\pm 0.19}$&$79.87^{\pm 0.18}$&$76.90^{\pm 0.23}$&$44.78^{\pm 0.25}$&$43.50^{\pm 0.24}$&$44.14^{\pm 0.29}$&$43.36^{\pm 0.25}$\\
MoFedSAM \cite{pmlr-v162-qu22a} &$76.36^{\pm 0.15}$&$65.74^{\pm 0.19}$&$76.74^{\pm 0.17}$&$70.74^{\pm 0.21}$&$41.07^{\pm 0.19}$&$34.11^{\pm 0.23}$&$35.91^{\pm 0.17}$&$28.55^{\pm 0.27}$\\
FedGAMMA\cite{10269141}&$80.89^{\pm 0.17}$&$75.34^{\pm 0.19}$&$81.73^{\pm 0.16}$&$78.74^{\pm 0.19}$&$49.78^{\pm 0.25}$&$46.31^{\pm 0.27}$&$47.91^{\pm 0.26}$&$45.26^{\pm 0.33}$\\
FedSMOO \cite{sun2023dynamic} &\textbf{84.17}$^{\pm 0.19}$&\textbf{80.92}$^{\pm 0.17}$&\textbf{84.78}$^{\pm 0.19}$&\textbf{82.79}$^{\pm 0.21}$&\textbf{52.31}$^{\pm 0.24}$&\textbf{49.42}$^{\pm 0.20}$&$50.59^{\pm 0.21}$&$46.08^{\pm 0.25}$\\
FedSpeed \cite{sun2023fedspeed} &$82.76^{\pm 0.19}$&$79.95^{\pm 0.19}$&$83.36^{\pm 0.18}$&$80.72^{\pm 0.22}$&$49.93^{\pm 0.23}$&$49.04^{\pm 0.24}$&\textbf{50.61}$^{\pm 0.23}$&\textbf{46.85}$^{\pm 0.25}$\\
FedLESAM &$80.11^{\pm 0.23}$&$74.35^{\pm 0.22}$&$78.35^{\pm 0.21}$&$71.23^{\pm 0.25}$&$44.35^{\pm 0.19}$&$43.75^{\pm 0.21}$&$43.97^{\pm 0.23}$&$43.21^{\pm 0.22}$\\
FedLESAM-D&$83.26^{\pm 0.19}$&$79.89^{\pm 0.20}$&$83.99^{\pm 0.23}$&$81.89^{\pm 0.21}$&$49.77^{\pm 0.20}$&$45.35^{\pm 0.22}$&$50.58^{\pm 0.19}$&$46.55^{\pm 0.21}$\\
FedLESAM-S\cite{FedLESAM}&$83.76^{\pm 0.17}$&$79.02^{\pm 0.18}$&$83.12^{\pm 0.20}$&$81.57^{\pm 0.21}$&$49.52^{\pm 0.19}$&$47.83^{\pm 0.22}$&$48.21^{\pm 0.20}$&$45.75^{\pm 0.24}$\\
FedTOGA(ours)&\textcolor{cvprblue}{\textbf{84.91}$^{\pm 0.15}$}&\textcolor{cvprblue}{\textbf{81.78}$^{\pm 0.17}$}&\textcolor{cvprblue}{\textbf{84.90}$^{\pm 0.19}$}&\textcolor{cvprblue}{\textbf{83.49}$^{\pm 0.14}$}&\textcolor{cvprblue}{\textbf{54.90}$^{\pm 0.16}$}&\textcolor{cvprblue}{\textbf{51.00}$^{\pm 0.15}$}&\textcolor{cvprblue}{\textbf{53.25}$^{\pm 0.17}$}&\textcolor{cvprblue}{\textbf{49.90}$^{\pm 0.21}$}\\\hline
\end{tabular}}
\end{table*}

\begin{table*}[]
\centering
\caption{Dirichlet coefficients $u$ selected from $\{0.1,0.6\}$, and $c$ is the Pathological coefficient, i.e., the number of active categories in each client. The CIFAR10 has 100 clients in the left part with 10\% active in each round,  and 200 clients in the right part with 5\% active in each round.(ResNet18)}
\label{tab_exp_resnet}
\scalebox{0.85}{
\begin{tabular}{l|cccc|cccc}
{Method}&\multicolumn{8}{c}{CIFAR10}\\ \cline{2-9}
 Partition&\multicolumn{2}{c}{Dirichlet}&\multicolumn{2}{c|}{Pathological}&\multicolumn{2}{c}{Dirichlet}&\multicolumn{2}{c}{Pathological}\\
Coefficient&{$u=0.6$}&{$u=0.1$}&{$c=6$}&{$c=3$}&{$u=0.6$}&{$u=0.1$}&{$c=6$}&{$c=3$}\\\hline
FedAvg \cite{DBLP:conf/aistats/McMahanMRHA17}  &$79.52^{\pm 0.13}$&$76.00^{\pm 0.18}$&$79.91^{\pm 0.17}$&$74.08^{\pm 0.22}$&$75.90^{\pm 0.21}$&$72.93^{\pm 0.19}$&$77.47^{\pm 0.34}$&$71.68^{\pm 0.34}$\\
FedAdam\cite{reddi2021adaptivefederatedoptimization} &$77.08^{\pm 0.31}$&$73.41^{\pm 0.33}$&$77.05^{\pm 0.26}$&$72.44^{\pm 0.20}$&$75.55^{\pm 0.38}$&$69.70^{\pm 0.32}$&$75.74^{\pm 0.22}$&$70.49^{\pm 0.26}$\\
SCAFFOLD\cite{pmlr-v119-karimireddy20a} &$81.81^{\pm 0.17}$&$78.57^{\pm 0.14}$&$83.07^{\pm 0.10}$&$77.02^{\pm 0.18}$&$79.00^{\pm 0.26}$&$76.15^{\pm 0.15}$&$80.69^{\pm 0.21}$&$74.05^{\pm 0.31}$\\
FedCM\cite{xu2021fedcmfederatedlearningclientlevel}  &$82.97^{\pm 0.21}$&$77.82^{\pm 0.16}$&$83.44^{\pm 0.17}$&$77.82^{\pm 0.19}$&$80.52^{\pm 0.29}$&$77.28^{\pm 0.22}$&$81.76^{\pm 0.24}$&$76.72^{\pm 0.25}$\\
FedDyn\cite{acar2021federated}  &$83.22^{\pm 0.18}$&$78.08^{\pm 0.19}$&$83.18^{\pm 0.17}$&$77.63^{\pm 0.14}$&$80.69^{\pm 0.23}$&$76.82^{\pm 0.17}$&$82.21^{\pm 0.18}$&$74.93^{\pm 0.22}$\\
FedSAM &$81.46^{\pm 0.12}$&$77.03^{\pm 0.17}$&$81.13^{\pm 0.23}$&$78.30^{\pm 0.24}$&$78.32^{\pm 0.16}$&$74.00^{\pm 0.14}$&$78.75^{\pm 0.27}$&$75.12^{\pm 0.29}$\\
MoFedSAM \cite{pmlr-v162-qu22a} &$85.29^{\pm 0.13}$&$80.25^{\pm 0.17}$&$84.74^{\pm 0.16}$&$83.09^{\pm 0.24}$ &$84.76^{\pm 0.20}$&\textbf{80.10}$^{\pm 0.14}$&85.00$^{\pm 0.27}$&\textbf{82.13}$^{\pm 0.23}$\\
FedGAMMA\cite{10269141}&$82.82^{\pm 0.16}$&$79.91^{\pm 0.15}$&$83.51^{\pm 0.18}$&$77.11^{\pm 0.14}$&$80.72^{\pm 0.19}$&$76.70^{\pm 0.14}$&$81.81^{\pm 0.27}$&$77.44^{\pm 0.29}$\\
FedSMOO \cite{sun2023dynamic} &\textbf{86.08}$^{\pm 0.14}$&\textbf{81.80}$^{\pm 0.18}$&\textbf{86.38}$^{\pm 0.15}$&\textbf{82.79}$^{\pm 0.16}$ &\textbf{84.96}$^{\pm 0.19}$&$79.76^{\pm 0.19}$&$84.82^{\pm 0.18}$&$81.01^{\pm 0.19}$\\
FedSpeed \cite{sun2023fedspeed} &$86.01^{\pm 0.16}$&$81.02^{\pm 0.16}$&$86.09^{\pm 0.19}$&$82.50^{\pm 0.16}$&$84.12^{\pm 0.18}$&$76.74^{\pm 0.14}$&$84.78^{\pm 0.27}$&$79.09^{\pm 0.29}$\\
FedLESAM &$81.04^{\pm 0.19}$&$76.92^{\pm 0.16}$&$81.37^{\pm 0.17}$&$78.21^{\pm 0.21}$&$77.80^{\pm 0.18}$&$73.73^{\pm 0.22}$&$78.44^{\pm 0.20}$&$74.53^{\pm 0.19}$\\
FedLESAM-D&$84.27^{\pm 0.14}$&$80.08^{\pm 0.19}$&$85.62^{\pm 0.18}$&$83.00^{\pm 0.22}$&$82.53^{\pm 0.19}$&$79.56^{\pm 0.27}$&\textbf{85.04}$^{\pm 0.21}$&{81.10}$^{\pm 0.19}$\\
FedLESAM-S\cite{FedLESAM}&$84.94^{\pm 0.12}$&$79.52^{\pm 0.17}$&$85.88^{\pm 0.19}$&$82.18^{\pm 0.15}$&$83.22^{\pm 0.22}$&$78.69^{\pm 0.17}$&$85.02^{\pm 0.24}$&$80.57^{\pm 0.17}$\\
FedTOGA(ours)&\textcolor{cvprblue}{\textbf{86.99$^{\pm 0.13}$}}&\textcolor{cvprblue}{\textbf{83.16$^{\pm 0.17}$}}&\textcolor{cvprblue}{\textbf{87.21$^{\pm 0.18}$}}&\textcolor{cvprblue}{\textbf{84.55$^{\pm 0.15}$}}&\textcolor{cvprblue}{\textbf{85.21}$^{\pm 0.17}$}&\textcolor{cvprblue}{\textbf{81.60}$^{\pm 0.16}$}&\textcolor{cvprblue}{\textbf{85.24}$^{\pm 0.19}$}&\textcolor{cvprblue}{\textbf{83.25}$^{\pm 0.20}$}\\\hline
\multicolumn{9}{l}{\textbf{Note}: The extended table sees Tab. \ref{tab_exp_resnet_extend}}
\end{tabular}}
\end{table*}

\begin{table}[htb]

\centering
\caption{WALL-CLOCK Time Comparison}
\label{tab_time}
\scalebox{0.85}{
\begin{tabular}{lclcl}\hline
Method&R(80\%)&Cost&R(82\%)&Cost\\\hline
FedSAM&481&3.6$\times$&800+&4.7$\times$\\\hline
MoFedSAM&167&1.2$\times$&270&1.6$\times$\\\hline
FedGAMMA&458&3.4$\times$&630&3.7$\times$\\\hline
FedSpeed&262&1.9$\times$&318&1.9$\times$\\\hline
FedSMOO&190&1.4$\times$&253&1.5$\times$\\\hline
FedLESAM-D&248&1.8$\times$&418&2.5$\times$\\\hline
FedTOGA&135&1.0$\times$&170&1.0$\times$\\\hline
\multicolumn{5}{l}{\textbf{Note}: The extended table sees Tab. \ref{tab_time_extend}}
\end{tabular}}
\end{table}

\section{Experiments}
This section describes part of the experimental setup, including the baseline algorithm, dataset, segmentation, and experimental details. We then show performance results on the baseline dataset, followed by extensive further analysis such as visualization as well as parameter sensitivity analysis experiments.
\subsection{Experimental Setups.}
\textbf{Baselines.} We compare FedTOGA with the baseline FedAvg\cite{DBLP:conf/aistats/McMahanMRHA17} and existing SAM-base FL methods, including FedSAM, MoFedSAM\cite{pmlr-v162-qu22a}, FedGAMMA\cite{10269141}, FedSMOO\cite{sun2023dynamic}, FedSpeed\cite{sun2023fedspeed},  with the recent FedLESAM\cite{FedLESAM}. Also, we compare with momentum-based FL algorithms, for example, FedAdam, FedYogi\cite{reddi2021adaptivefederatedoptimization}, FedACG\cite{kim2024communication}, FedCM\cite{xu2021fedcmfederatedlearningclientlevel}. In addition, methods based on local consistency are also included in the comparison, including FedDyn\cite{acar2021federated}, SCAFFOLD\cite{pmlr-v119-karimireddy20a} , FedDC\cite{gao2022federated} with FedRCL \cite{seo2024relaxed}.

\noindent\textbf{Dataset and Splits.}
The benchmark datasets CIFAR10 and CIFAR100 are utilized in the experiments. We follow the methodologies outlined in \cite{dai2023fedgamma,sun2024understanding,fan2024locally} to simulate client data using Dirichlet and Pathological splits in non-IID scenarios. To emulate real-world conditions where a large number of clients are involved, some of whom may lag, we conducted three experimental setups: First, data is distributed to 100 clients, with 10\% of clients participating in training in each round. Second, data is distributed to 200 clients, with 5\% of clients participating in each training round.

\noindent\textbf{Experimental Details.} 
For a fair comparison, we adopt the same experimental setup as in \cite{sun2023dynamic, FedLESAM}. We use SGD as the optimizer, with a client learning rate $\eta_l$ set to 0.1 and a global learning rate of 1. The weight decay is fixed at $1e^{-3}$. To further assess the generalization capability of our method, we conduct experiments with two models, LeNet and ResNet18\cite{7780459}. For LeNet, the learning rate decays by 0.997 per epoch, whereas for ResNet18\cite{7780459}, it decays by 0.998. For CIFAR10, the batch size is set to 50, and the number of local epochs is set to 5. For CIFAR100, the batch size is set to 20, and the number of local epochs is set to 2. In FedTOGA, the local perturbation coefficient $\kappa$ is set to 1, the dual variable coefficient $\beta$ is set to 0.9, and the penalized coefficient $\alpha$ is set to 0.1. Consistent with several other algorithms, the perturbation magnitude $\rho$ is set to 0.1, except for FedSAM, MoFedSAM, and FedLESAM, where it is set to 0.01. Detailed information about the experimental setup can be found in Appendices \ref{hyperparameters-set}.

\subsection{Performance Evaluation}
\begin{figure*}[htb]
\centering 
 \subcaptionbox*{}{\includegraphics[width=0.19\textwidth]{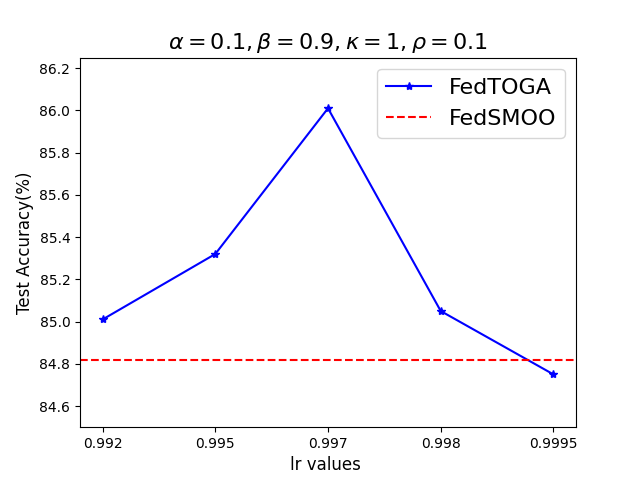}}
 \subcaptionbox*{}{\includegraphics[width=0.19\textwidth]{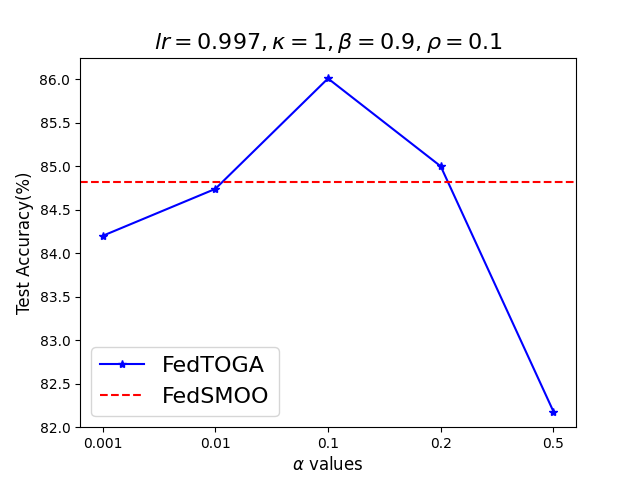}}
 \subcaptionbox*{}{\includegraphics[width=0.19\textwidth]{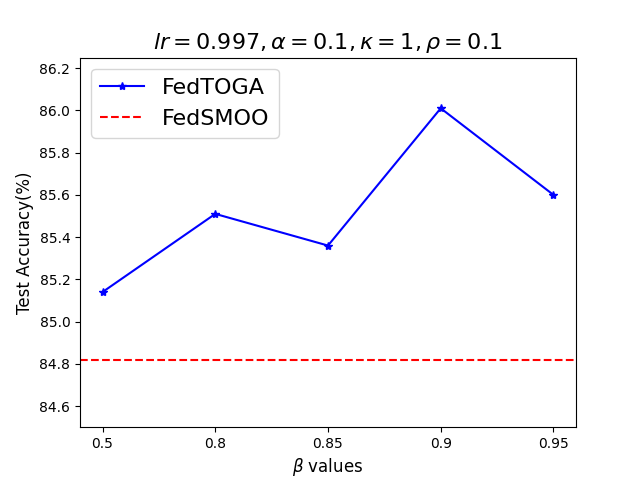}}
 \subcaptionbox*{}{\includegraphics[width=0.19\textwidth]{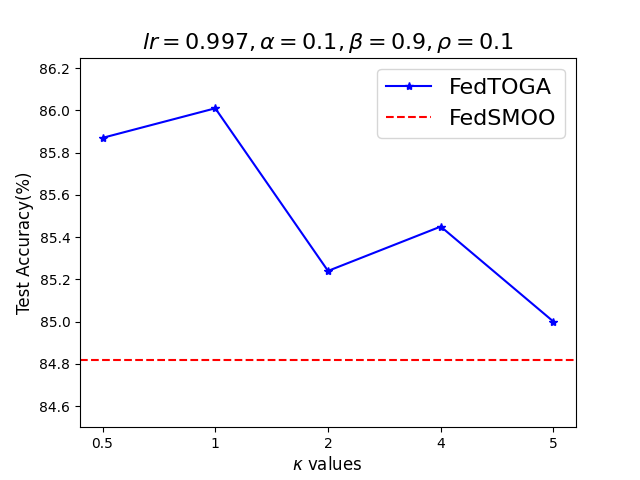}}
 \subcaptionbox*{}{\includegraphics[width=0.19\textwidth]{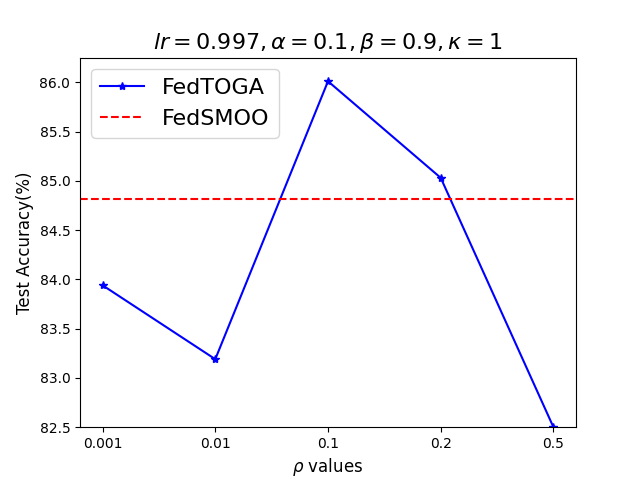}}
\caption{Hyperparameters sensitivity studies of lr decay, penalized coeficient $\alpha$, Correction coeficient $\beta, \kappa$ and perturbations  coefficient $\rho$ on CIFAR-10.}
\label{sensitivity}
\end{figure*}

\textbf{Performance with compared benchmarks.}
As shown in Tables \ref{tab_exp_lenet} and \ref{tab_exp_resnet}, the proposed FedTOGA algorithm shows excellent performance on various heterogeneous datasets in terms of convergence speed and final achieved accuracy. Table \ref{tab_exp_lenet}, which details the test accuracy of the LeNet model, demonstrates that FedTOGA significantly outperforms other algorithms with different heterogeneous data conditions. Specifically, under the Dirichlet-0.1 setting on the CIFAR10 dataset, FedTOGA attains an accuracy of 82.05\%, marking a significant improvement of over 7.37\% compared to vanilla FedAvg and a 1.99\% increase over the second-highest baseline accuracy. Similar results are observed in Table \ref{tab_exp_resnet} for the ResNet18 model, FedTOGA outperforms all current baseline algorithms.

As seen in Table \ref{tab_time}, FedTOGA also exhibits a significant advantage in terms of convergence speed. When reaching 80\% accuracy, FedTOGA converges 3.6× faster compared to FedSAM and 1.2× faster compared to the second-best algorithm. Similarly, when reaching 82\% accuracy, FedTOGA converges 4.7× faster compared to FedSAM and 1.5× faster compared to the second-best algorithm. This indicates that FedTOGA achieves the target accuracy with significantly reduced computation and communication overhead compared to other methods. The exceptional global consensus capability of FedTOGA enables it to more effectively mitigate the impact of data heterogeneity and enable faster convergence.
 
\noindent\textbf{Impact of heterogeneity.} We use the Dirichlet and Pathological methods for data partitioning. For the Dirichlet distribution, we adopt with variance coefficients $u$ of 0.1 and 0.6. For the Pathological distribution, we use coefficients $c$ of 3 and 6. As shown in Tables \ref{tab_exp_lenet} and \ref{tab_exp_resnet}, increased data heterogeneity leads to a decrease in accuracy across all algorithms. However, FedTOGA exhibits the smallest accuracy drop. Specifically, for the Resnet18 model, as $u$ changes from 0.6 to 0.1 under the Dirichlet distribution on the CIFAR10 dataset, FedSAM's accuracy decreases from 81.46\% to 77.03\%, a 4.43\% reduction, while the second-best algorithm, FedSMOO, shows a drop from 86.08\% to 81.80\%, a 4.28\% reduction. In contrast, FedTOGA's accuracy declines from 86.99\% to 83.16\%, a 3.83\% drop. Similar trends are observed under the Pathological distribution, underscoring FedTOGA's superior stability and accuracy across varying levels of data heterogeneity.

\noindent\textbf{Impact of partial participation.}
We fix all hyperparameters except for the client participation rate to assess its effect on accuracy. As illustrated in Table \ref{tab_exp_resnet}, a reduction in the client participation rate from 10\% to 5\% results in a modest decline in accuracy across all algorithms. For instance, on the CIFAR10 dataset, under the challenging pathological distribution with $c=3$, FedTOGA's accuracy decreases marginally from 84.55\% to 83.35\%, a reduction of just 1.40\%, while FedSMOO experiences a sharper decline from 82.79\% to 81.01\%, a reduction of 1.78\%. Similarly, under the Dirichlet distribution with $u=0.1$, FedTOGA's accuracy decreases from 86.99\% to 85.21\%, a decrease of 1.78\%, whereas FedSMOO’s accuracy drops from 86.08\% to 84.96\%, a reduction of 1.12\%. Notably, despite these reductions, FedTOGA consistently outperforms other algorithms in terms of accuracy, highlighting its robust generalization capability and stability. 

\noindent\textbf{Hyperparameters Sensitivity.}  
We study the sensitivity of the hyperparameters: learning rate decay, penalty coefficient $\alpha$, correction coefficients $\beta$ and $\kappa$, and perturbation coefficient $\rho$. As shown in Figure \ref{sensitivity}, our extensive experiments demonstrate FedTOGA's resilience to variations in these hyperparameters. By systematically adjusting each parameter while holding the others constant, we found that FedTOGA remains remarkably stable under changes in learning rate decay and the correction coefficients $\beta$ and $\kappa$. Additionally, the penalty coefficient $\alpha$ and perturbation coefficient $\rho$ effectively maintain robust performance when appropriately selected.

\section{Conclusion}
In this paper, we propose a novel federated learning algorithm, FedTOGA, which for the first time estimates global perturbations by combining global training gradients, and corrects the local dynamic regularizer. This ensures that local optima can be effectively aligned with the global generalization and optimization objectives. FedTOGA facilitates the efficient search for globally consistent flat minima and accelerates convergence without incurring additional local storage or uplink communication overhead. Theoretical analysis guarantees that FedTOGA achieves a fast convergence rate of $O(1/T)$. Extensive experiments were conducted to verify its efficiency and remarkable performance.

\newpage
{ \small
    \bibliographystyle{ieeenat_fullname}
    \bibliography{main}
}

\newpage
\onecolumn
\appendix
\section{Related Works}
\label{A}
\subsection{Literature review}
\label{A-1}
In this section, we review the contributions of related works.


\noindent\textbf{Federated Learning } 
Federated learning gained widespread attention upon its introduction due to its data-exchange-free nature. FedAvg\cite{DBLP:conf/aistats/McMahanMRHA17} As its foundational framework, it allows for collaborative modeling by passing models around without exchanging data\cite{ stich2019localsgdconvergesfast}. However, due to various irresistible factors, the data of cooperative devices show heterogeneous distribution, which makes the modeling effectiveness suffer. Therefore, many studies based on empirical loss minimization have been proposed to solve this problem. FedProx\cite{li2020federatedoptimizationheterogeneousnetworks} employs a simple and intuitive practice, namely, paradigm constraints, which is used to ensure that the local model is never far from the global model. Specifically, it introduces in regular terms during local training to limit the distance of its local model from the global model. SCAFFOLD \cite{pmlr-v119-karimireddy20a}, Mime \cite{DBLP:conf/nips/KarimireddyJKMR21} uses control variables for local updates, however, its requires greater communication overhead. FedDyn \cite{acar2021federated}, FedPD \cite{Zhang2020FedPDAF} considers the inconsistency of the local optimal point with the global optimal point to be a fundamental dilemma, which aligns the locally optimal solution to the global optimal solution via a dynamic regularizer. FedPA \cite{alshedivat2021federated} removes bias from client updates by estimating a global posterior. FedDC \cite{gao2022federated} takes decoupled local as well as updated quantities to mitigate heterogeneity. Furthermore, recent research has shown that this model bias is similar to catastrophic forgetting in continuous learning \cite{NEURIPS2022_fadec8f2,lee2024fedsolstabilizedorthogonallearning,shoham2019overcomingforgettingfederatedlearning,wang2023comprehensivesurveyforgettingdeep}, Clients overriding previously important parameters to learn a new task resulted in disrupting pretask performance. Some studies have mitigated global knowledge collapse by class task recall\cite{rebuffi2017icarlincrementalclassifierrepresentation,dong2022federatedclassincrementallearning}. Server momentum\cite{sun2023roleservermomentumfederated} algorithms also play an important role in FL. \cite{DBLP:conf/nips/ZaheerRSKK18} investigates the convergence failure of ADAM in certain non-convex settings and develops an adaptive optimizer, YOGI, which aims to improve convergence. \cite{reddi2021adaptivefederatedoptimization} integrates it in a FL framework. FedAvgM \cite{DBLP:journals/corr/abs-1909-06335} using Momentum \cite{DBLP:journals/nn/Qian99}, while FedACG \cite{kim2024communication} utilizes NAG \cite{Nesterov1983AMF}. And FedCM\cite{xu2021fedcmfederatedlearningclientlevel} mitigates local heterogeneity by using the proximity global update gradient applied to the client momentum. Inspired by FedCM\cite{xu2021fedcmfederatedlearningclientlevel}, we approximate the global update gradient $\Delta$ as $\nabla f(\theta)$,to estimate the global perturbation range.


\noindent\textbf{Sharpness-aware Minimization}.  
Many studies\cite{NIPS1994_01882513,dinh2017sharpminimageneralizedeep} have pointed out that flat minima imply superior generalization performance, which possesses greater robustness to model perturbations. In order to minimize sharpness\cite{keskar2017largebatchtrainingdeeplearning}, \cite{foret2021sharpnessawareminimizationefficientlyimproving, becker2024momentumsamsharpnessawareminimization} proposed sharpness-aware minimization (SAM), and many works \cite{li2023enhancingsharpnessawareoptimizationvariance, mueller2023normalizationlayerssharpnessawareminimization} were carried out. Specifically, SAM's only capture the sharpness of specific small batches of data, and VaSSO\cite{mueller2023normalizationlayerssharpnessawareminimization} aims to address this issue. Our FedTOGA can also help solve this problem to some extent, first we add neighborhood perturbations $\tilde{g}_{i,k-1}^t$ to help the local SAM optimizer perceive the amount of neighboring batch data perturbations (optionally), and global update perturbations via $\Delta^{t-1}$. In addition, m-SAM \cite{pmlr-v162-andriushchenko22a} can be considered to be closely related to federated sharpness minimization. m-SAM \cite{pmlr-v162-andriushchenko22a} quantifies the sharpness of batches across m training point batches, averaging out multiple disjoint batches in the generated Updates. The neighborhood global perturbation proposed by our FedTOGA alleviates the problem of local perturbation isolation in the FL paradigm and can be applied to all existing algorithms.

\noindent\textbf{SAM in Federated Learning} 
In order to extend the generalizability of local models in FL, \cite{pmlr-v162-qu22a,caldarola2022} introduced SAM into the FL paradigm to propose FedSAM. further, \cite{pmlr-v162-qu22a} combined with FedCM\cite{ xu2021fedcmfederatedlearningclientlevel} to propose MoFedSAM. FedGAMMA\cite{10269141} introduces the variance reduction technique of SCAFFOLD\cite{pmlr-v119-karimireddy20a} into FedSAM and gets some results. And FedSpeed \cite{sun2023fedspeed} uses SAM to optimize FedDyn\cite{acar2021federated}. FedSMOO\cite{sun2023dynamic} builds on FedSpeed to use dynamic regularization to SAM to estimate global disturbances. FedSOL\cite{lee2024fedsolstabilizedorthogonallearning} uses the orthogonal idea of continuous learning to make local perturbations close to the global. \cite{FedLESAM} proposes an efficient algorithm, the Local Estimation of Global Perturbations SAM (FedLESAM), which optimizes global sharpness and reduces computation. As we have seen, FedSAM\cite{pmlr-v162-qu22a,caldarola2022}, MoFedSAM, and FedGAMMA\cite{10269141} compute local perturbations and optimize sharpness on client data, which may result the local SAM does not reach the global flat minimum. Several studies have identified this drawback and attempted to address it. FedSOL\cite{lee2024fedsolstabilizedorthogonallearning} uses local orthogonal solving to limit the range of local perturbations, which can lead to perturbation absences. FedSMOO\cite{sun2023dynamic} uses dynamic regularity to compute and add corrections, however, requires additional communication and storage overheads. FedLESAM\cite{ FedLESAM} believes that additional computation would be burdensome and therefore uses historical storage parameters to estimate global perturbations. However the above solutions may have limitations due to network fluctuations, which we discuss in detail in \ref{A-2}. Therefore, we propose FedTOGA to estimate the global perturbation using the global update.

\newpage

\subsection{Existing limitations}
\label{A-2}
\textbf{FedSMOO}\cite{sun2023dynamic}  In the algorithm \ref{algorithm_smoo}, to solve the model perturbation problem, \cite{sun2023dynamic} utilizes the local Augmented Lagrangian function to penalize the deviation of the local perturbation from the global perturbation. As training proceeds, the local perturbation is made to approach the global perturbation gradually. However, it needs to open extra storage space on the client side to record $\mu_i,\tilde{s}_i$. Meanwhile, $\tilde{s}_i$ needs to be synchronously uploaded to update the global perturbation variable $s$ at the time of aggregation during the communication process, which doubles the communication overhead. Further, this estimation bias will be exacerbated by the server's strategy of randomly selecting the set of clients $S_t$ to mitigate the communication overhead due to communication bottlenecks.
\begin{small}

\begin{algorithm}[H]
\caption{FedSMOO Algorithm}
\label{algorithm_smoo}
Initial ${\theta^0,\theta_i,s_i,s,\lambda_i,\lambda,\mu_i}$\\
\For{each round $t \in [T] \triangleq \{0, 1, 2, \cdots, T-1 \}$}{
     Sample the active client set $S_t \subseteq [N]$.\\
    \For{$i \in S_t$ in parallel}{
      ${\theta}_i^t,\tilde{s}_i \leftarrow \textbf{Client Update}(\theta^{t},s^t$); \quad
    \textbf{ communicate} ${\theta}_i^t,\tilde{s}_i$ to server \;    
        }
   
  $\mathcal{S}^t = \frac{1}{M}\sum_{i \in S_t}\tilde{s}_i ;s^t = \rho \frac{\mathcal{S}^t}{\|\mathcal{S}^t\|}$;
 $\quad h^{t+1} = h^{t} - \frac{1}{\alpha N}\sum_{i \in S_t} ({\theta}_i^t - \theta^t); \quad \theta^{t+1} =  \frac{1}{M}\sum_{i \in  S_t} {\theta}_i^t - \alpha h^{t+1}$;\\ 

}
\textbf{Client Update($ \theta_{t},s^t$):} $\theta_{i,0}^t = \theta^{t};s=s^t$\\
\For{local epoch $k \in [K] \triangleq t\{0,1, 2, \cdots, K-1 \}$}{
    sample a  mini-batch data $\xi_{i,k}^t$;
    gradient estimate: $g_{i,k}^t = \nabla f_i(\theta_{i,k}^t;\xi_{i,k}^t)$;\\
    Perturbation: $ \mathcal{S}_{i,k}^t = g_{i,k}^t - \mu_i - s;\hat{s}_{i,k}^t = \rho \frac{\mathcal{S}_{i,k}^t}{\| \mathcal{S}_{i,k}^t\|}; \mu_i = \mu_i + (\hat{s}_{i,k}^t  -s)$\\
    extra-step:
    $\tilde{g}_{i,k}^t = \nabla f_i(\theta_{i,k}^{t} +\hat{s}_{i,k}^t; \xi_{i,k}^t);\quad$
    $ \theta_{i,k+1}^t =\theta_{i,k}^{t}  - \eta_l\left(\tilde{g}_{i,k}^t -h_i^{t} +\frac{1}{\alpha}\left( \theta_{i,k}^{t} - \theta_{i,0}^t\right)\right) $
}
$\tilde{s}_i = \mu_i - \hat{s}_{i,K}^t; h_i^{t+1} = h_i^{t} - \frac{1}{\alpha}\left( \theta_{i,K}^{t} - \theta_{i,0}^t \right)$\\
\textbf{return} ${\theta}_i^t = \theta_{i,K}^{t};\tilde{s}_i$\\

\end{algorithm}
\end{small}
\noindent\textbf{FedLESAM}\cite{FedLESAM} 
In the algorithm \ref{algorithm_lesam}, to reduce the computational overhead and estimate the global perturbations, \cite{FedLESAM} utilizes the historical global model record values $\theta^{old}_i$ to compare with the latest round's global model $\theta^t$ to estimate the global perturbations. This poses the same problem as the algorithm \ref{algorithm_smoo} described above, specifically, in the face of an extreme case where participants will only participate in one global aggregation, FedLESAM will not be able to efficiently estimate the global perturbation variables. Meanwhile, the perturbation scales will vary when the frequency of client participation is different. In addition, since the perturbation estimation does not include the current perturbation computation it may not be possible to accurately estimate the current perturbation direction.

\begin{algorithm}[H]
\small
\caption{FedLESAM-D Algorithm}
\label{algorithm_lesam}
Initial ${\theta^0,\theta_i^{old}, h_i,h}$\\
\For{each round $t \in [T] \triangleq \{0, 1, 2, \cdots, T-1 \}$}{
     Sample the active client set $S_t \subseteq [N]$.\\
    \For{$i \in S_t$ in parallel}{
      ${\theta}_i^t \leftarrow \textbf{Client Update}(\theta^{t}$); \quad
    \textbf{ communicate} ${\theta}_i^t$ to server \;    
        }

 $h^{t+1} = h^{t} - \frac{1}{\alpha N}\sum_{i \in S_t} ({\theta}_i^t - \theta^t); \quad \theta^{t+1} =  \frac{1}{M}\sum_{i \in  S_t} {\theta}_i^t - \alpha h^{t+1}$;\\ 

}
\textbf{Client Update($ \theta_{t}$):} $\theta_{i,0}^t = \theta^{t}$\\
\For{local epoch $k \in [K] \triangleq \{0,1, 2, \cdots, K-1 \}$}{
    sample a  mini-batch data $\xi_{i,k}^t$;
    Perturbation: $\delta_{i,k}^t = \rho \frac{\theta^{old}_i  - \theta^t}{\| \theta^{old}_i  - \theta^t\|}$\\
    extra-step:
    $\tilde{g}_{i,k}^t = \nabla f_i(\theta_{i,k}^{t} +\delta_{i,k}^t; \xi_{i,k}^t);\quad$
    $ \theta_{i,k+1}^t =\theta_{i,k}^{t}  - \eta_l\left(\tilde{g}_{i,k}^t -h_i^{t} +\frac{1}{\alpha}\left( \theta_{i,k}^{t} - \theta_{i,0}^t\right)\right) $
}
$ h_i^{t+1} = h_i^{t} - \frac{1}{\alpha}\left( \theta_{i,K}^{t} - \theta_{i,0}^t \right); \quad \theta^{old}_i = \theta^t$\\
\textbf{return} ${\theta}_i^t = \theta_{i,K}^{t}$\\

\end{algorithm}


\begin{table*}[htb]
\centering
\caption{Abstract for the SAM-based FL algorithms for solving data heterogeneity, focusing on the basic algorithm, sharpness minimization objective, perturbation computation strategy, additional communication, and storage overhead comparison.}
\label{tab-abs-p}
\scalebox{0.85}{
\begin{tabular}{l|ccccc}\hline
 \small
Works &Base Algorithm &Minimizing Target&Local Perturbation& Extra Storage &Extra  Communication \\\hline
FedSAM &FedAvg&Local Sharpness&$\rho \frac{g_{i,k}^t}{\|g_{i,k}^t\|}$&$1\times$&$1\times$\\
MoFedSAM \cite{pmlr-v162-qu22a} &FedCM&Local Sharpness&$\rho \frac{g_{i,k}^t}{\|g_{i,k}^t\|}$&$1\times$&$1\times$\\
FedSpeed \cite{sun2023fedspeed} &FedDyn&Local Sharpness&$\rho \frac{g_{i,k}^t}{\|g_{i,k}^t\|}$&$2\times$&$1\times$\\
FedGAMMA\cite{10269141}&SCAFFOLD&Local Sharpness& $\rho \frac{g_{i,k}^t}{\|g_{i,k}^t\|}$&$2\times$&$2\times$\\
FedSMOO \cite{sun2023dynamic} &FedDyn&\makecell[c]{Local Sharpness\\With Correction}&$\rho \frac{g_{i,k}^t -\mu_i -s}{\|g_{i,k}^t-\mu_i -s\|}$&$3\times$&$2\times$\\
FedLESAM(S-D)\cite{FedLESAM}&\makecell[c]{FedDyn\\SCAFFOLD } &\makecell[c]{Global Sharpness\\Estimate}&$\rho \frac{\theta_i^{old} - \theta_t}{\|\theta_i^{old} - \theta_t\|}$&$3\times$&$1\times$\\
\textbf{FedTOGA(ours)}&FedDyn FedCM&\makecell[c]{Local With Global \\Sharpness Estimate}&$\rho \frac{g_{i,k}^t [\tilde{g}_{i,k-1}^t] + \kappa \Delta^{t}}{\|g_{i,k}^t[\tilde{g}_{i,k-1}^t] + \kappa \Delta^{t}\|}$&$2\times$&$1\times$\\\hline

\end{tabular}}
\end{table*}
\subsection{Neighbourhood Perturbation strategy}
\label{A-4}

\begin{figure}[H]
\centering
\includegraphics[scale=0.45]{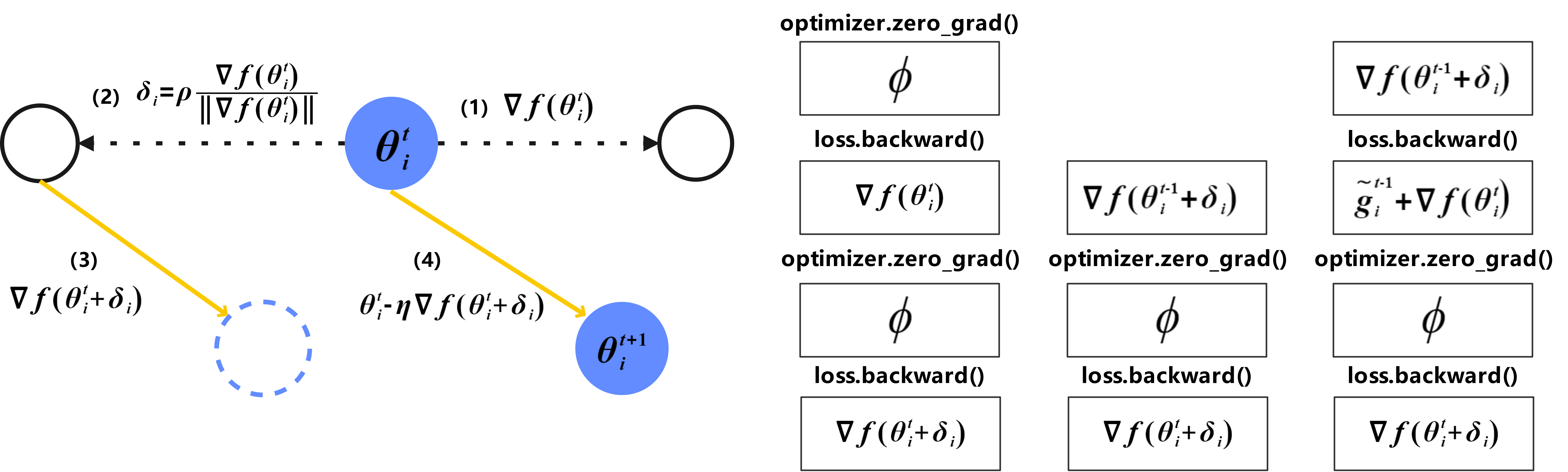}
\caption{Illumination of the perturbation technique and its variants}
\label{fig-ngp}
\end{figure}

\noindent\textbf{How Neighbourhood Perturbation works?} 
Let's take Pytorch as an example. Recall the fact that when the local iteration interval is greater than 1, the gradient register needs to be cleared by (\texttt{optimizer.zero\_grad()}) each time the gradient computation is performed, this is to prevent the accumulating gradient  to causing errors. Recognizing that using the neighborhood gradient variables in the registers does not need any substantial additional overhead, we use the neighboring gradients temporarily stored in the registers to simulate the current perturbation gradient. We give an example of a local gradient computation to help better understand the workflow. We initialize the model $\theta_0$ and a gradient register $G=[\emptyset]$. After the first calculation of the gradient (\texttt{loss.backward()}), the register is updated $G=[g_1=\nabla f(\theta_0)]$.
If SGD is used, then $\theta_1 =\theta_0-\eta g_1 $, followed by clearing the register $G=[\emptyset]$ before computing the second gradient. If SAM is used, then the perturbation is computed as $\delta_1 =\rho\frac{g_1}{\|g_1\|}$, then the gradient register is emptied, and after the gradient is computed again as $\tilde{g}_1 =\nabla f(\theta +\delta_1)$, perform model update $\theta_1 =\theta_0-\eta \tilde{g}_1 $. So the register status changes to $G=[\emptyset]\rightarrow[g_1]\rightarrow[\emptyset]\rightarrow[\tilde{g}_1]$. When neighborhood perturbation is enabled, gradient calculation and clearing before SAM  are no longer required. Therefore, the register status changes to $G=[{g}_0]\rightarrow[\emptyset]\rightarrow[{g}_1]$, The client will directly use the previously calculated gradient of the gradient register as the perturbation variable. You can observe the gradient calculation and cache change process in Fig. \ref{fig-ngp}.


\noindent\textbf{How Perturbation Fusion works?} 
With the above technical means of neighborhood perturbation, we can easily merge it in the perturbation computation by not emptying the gradient cache before computing the perturbation rise, then the gradient cache state will change to $G=[\tilde{g}_0]\rightarrow[\tilde{g}_0 + g_1]\rightarrow[\emptyset]\rightarrow[\tilde{g}_1]$.


\noindent\textbf{Waht is LOOKAHEAD?} LOOKAHEAD\cite{zhang2019lookaheadoptimizerksteps} uses a fast-slow-step mechanism, where a retrospective is performed every $K$ steps forward. The idea behind this is to take a step in the direction of the current gradient update, and then use a set of additional weights (called “slow weights”) to take a step in the same direction, but on a longer time scale. These slow weights are updated less frequently than the original weights, effectively creating a “look ahead” into the future of the optimization process. By incorporating N-perturbation techniques, it forms a lookahead-like updating mechanism that helps the optimizer escape local minima and saddle points more efficiently, leading to faster convergence.

\newpage
\section{ Experiments}
\subsection{Introduction of Datasets}
\label{exp-set}
\begin{table*}[htb]
\centering
\caption{ Summary of CIFAR10/100}
\label{tab-dataset}
\begin{tabular}{l|ccccc}\hline
Dataset&Total Number& Train Data& Test Data& Class & Size\\\hline
CIFAR10&60,000&50,000&10,000&10&3$\times$32$\times $32\\
CIFAR100&60,000&50,000&10,000&100&3$\times$32$\times$32\\\hline
\end{tabular}
\end{table*}
CIFAR10 and CIFAR100 are two datasets widely used in image classification and machine learning research. As shown in Table \ref{tab-dataset}, CIFAR10 contains 60,000 color images in 10 categories with 6,000 images in each category and image size of $32\times32$ pixels. CIFAR100 is similar to CIFAR10, but it contains 100 categories with 600 images per category, 500 of which are used for training and 100 for testing. These categories are further categorized into 20 major categories, each of which contains five subcategories.

\subsection{Detailed Hyperparameters Selection}
\label{hyperparameters-set}
To ensure a fair comparison across different datasets, we employed an experimental design consistent with  FedGAMMA \cite{10269141}, FedSMOO \cite{sun2023dynamic}, and FedLESAM \cite{FedLESAM}. ResNet18 \cite{7780459} was selected as the backbone model, utilizing group normalization \cite{wu2018groupnormalization} and stochastic gradient descent (SGD). A total of 800 training rounds were conducted, with the initial local learning rate set to $\eta_l = 0.1$. The global learning rate was maintained at $\eta_g = 1.0$ for most experiments, except for FedAdam and FedYOGI \cite{reddi2021adaptivefederatedoptimization} were adjusted to $0.01$. The penalty coefficients $\alpha$ for FedDC \cite{gao2022federated} and FedDyn \cite{acar2021federated} were uniformly set to $0.01$ in the LeNet, consistent with \cite{gao2022federated}, but were increased to $0.1$ for ResNet18. In FedACG \cite{kim2024communication}, following its prescribed settings, the local penalty coefficient was set to $\mu=0.01$, and the server momentum coefficient $\lambda$ was set to $0.85$. For FedAdam and FedYOGI \cite{reddi2021adaptivefederatedoptimization}, the parameters were set as $\beta_1=0.9$, $\beta_2=0.99$, and $\tau=1e^{-3}$. The momentum trade-off coefficient for FedCM \cite{xu2021fedcmfederatedlearningclientlevel} was configured as $\alpha=0.1$. In the SAM-based algorithms, the penalty coefficients for FedSpeed, FedSMOO, FedLESAM-D, and FedTOGA were uniformly set to $\alpha=0.1$. The perturbation coefficients for FedGAMMA, FedSpeed, FedSMOO, FedLESAM(S-D), and FedTOGA were consistently set to $\rho=0.1$ for ResNet18, except for FedSAM and MoFedSAM \cite{pmlr-v162-qu22a} and the vanilla FedLESAM coefficient were set to $0.01$. In the LeNet experiments, the perturbation coefficients $\rho$ for FedGAMMA, FedLESAM, and its variants were set to $0.01$, though in some cases $0.1$ yielded better performance. Weight decay was uniformly set to $1e^{-3}$ across all experiments.In the ResNet18 experiments, the learning rate decay were set to $0.998$ for most methods, except for FedSMOO, FedLESAM and its variants which were set to $0.9995$. In the 200-client case the learning rate decay was set to $0.9995$ (This is not always the case, in some cases a learning rate decay of $0.998$ works better, and we kept only the best results).
In most scenarios, the local perturbation correction coefficient $\kappa$ for FedTOGA was set to 1; however, in cases of increased heterogeneity, $\kappa$ could be slightly enlarged, but not beyond the local interval value $K$. The local dual variable correction coefficient $\beta$ for FedTOGA was chosen within the range of 0 to 1, with $0.8$ or $0.9$ typically performing best on CIFAR10. Generally, the parameter selection range can be determined according to Table \ref{tab-hpselect}.

\begin{small}
\begin{table*}[htb]
\centering
\caption{ Hyperparameters Selection.}
\label{tab-hpselect}
\begin{tabular}{l|cccc}\hline
Options&SGD-type & Best Selection & proxy-Type&  Best Selection\\\hline
Local Learning Rate&\{0.01,0.1,0.5\}&0.1&\{0.01,0.1,0.5\}&0.1\\
Global Learning Rate&\{0.01,0.1,1.0\}&1.0&\{0.01,0.1,1.0\}&1.0\\
Learning Rate Decay&\{0.995,0.998,0.9995\}&0.998&\{0.997,0.998,0.9995\}&0.9995\\
SAM Learning Rate&\{0.001,0.01,0.1\}&0.01&\{0.001,0.01,0.1\}&0.1\\
penalized coefficient $\alpha$&\{0.01,0.1,0.2\}&0.1&\{0.01,0.1,0.2\}&0.1\\
client-level momentum  $\alpha$&\{0.01,0.05,0.1\}&0.1&-&-\\
SAM Perturbation Correction  $\kappa$&-&-&\{1,2,4\}&1\\
Dual variable Correction  $\beta$&-&-&\{0.8,0.85,0.9\}&0.9\\
Server-level momentum $\lambda$&\{0.8,0.85,0.9\}&0.85&-&-\\\hline
\end{tabular}
\end{table*}
\end{small}

\textbf{Test Experiments}: Quadro RTX 6000; NVIDIA-SMI 515.76;  Driver Version 515.76; CUDA Version 11.7 
\newpage
\subsection{Distributions of Dirichlet and Pathological Split}
\begin{figure*}[htb]
\centering 
 \subcaptionbox{CIFAR10  $\alpha=0.6$}{\includegraphics[width=0.22\textwidth]{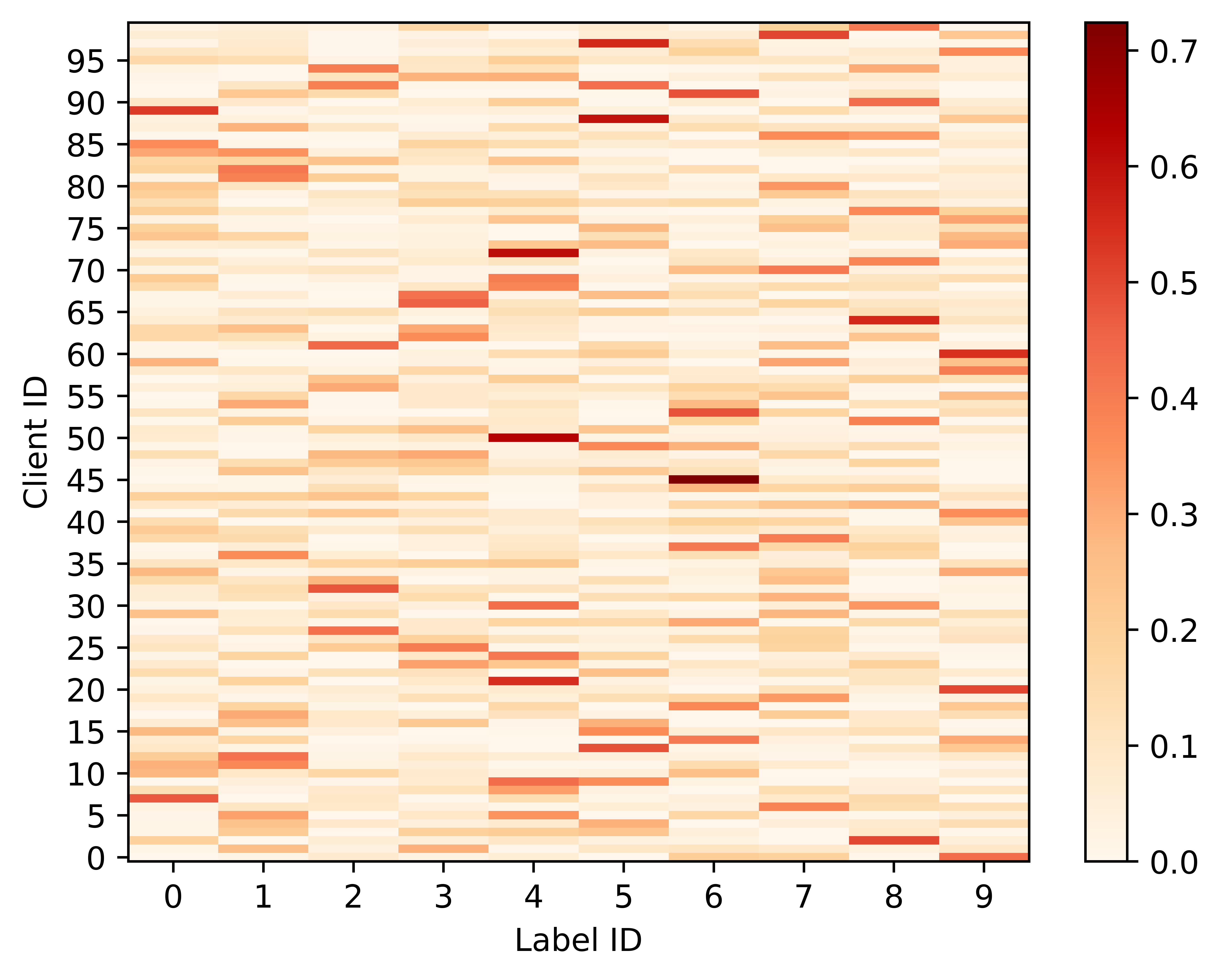}}
 \subcaptionbox{CIFAR10  $\alpha=0.1$}{\includegraphics[width=0.22\textwidth]{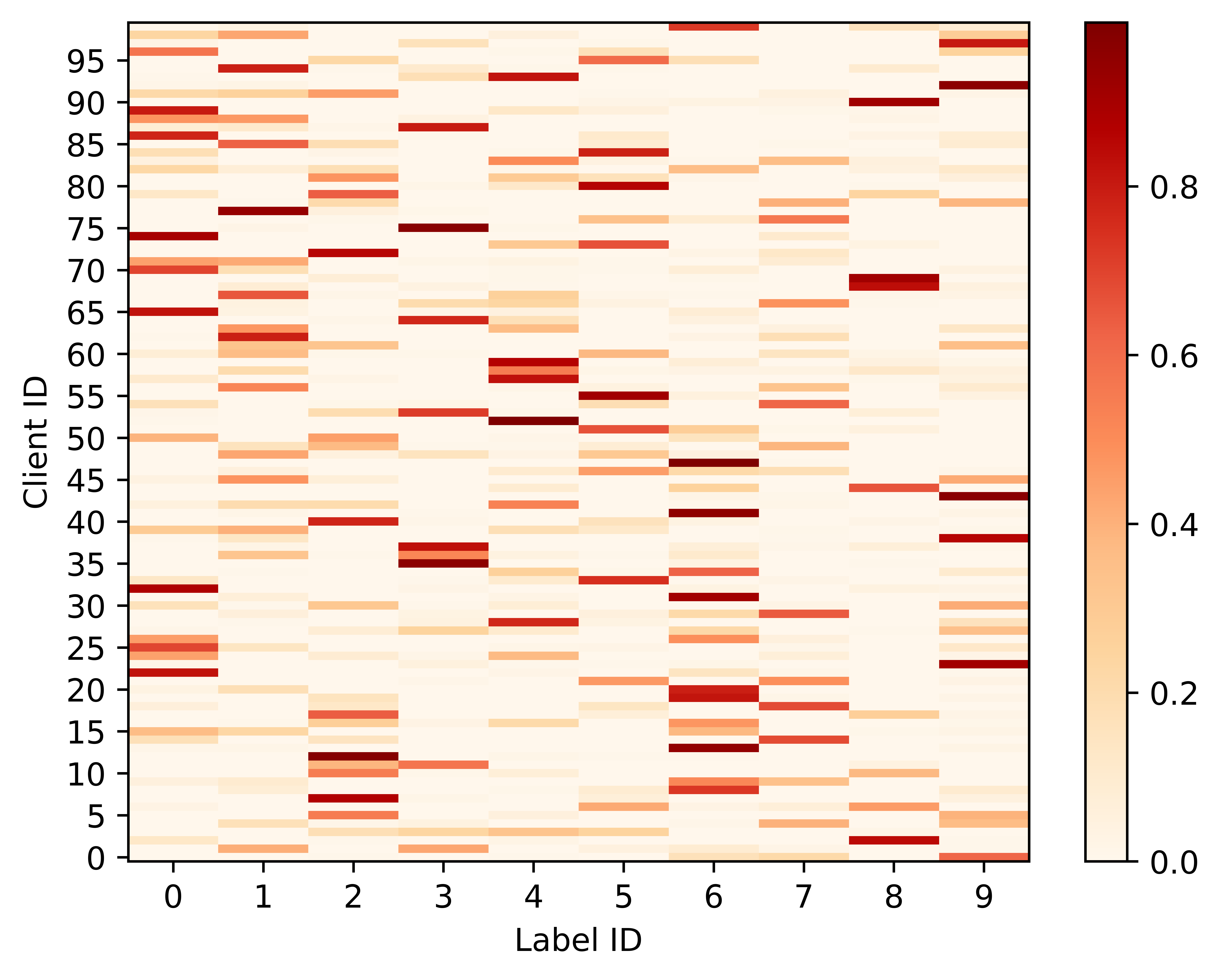}}
 \subcaptionbox{CIFAR10  $\beta=6$}{\includegraphics[width=0.22\textwidth]{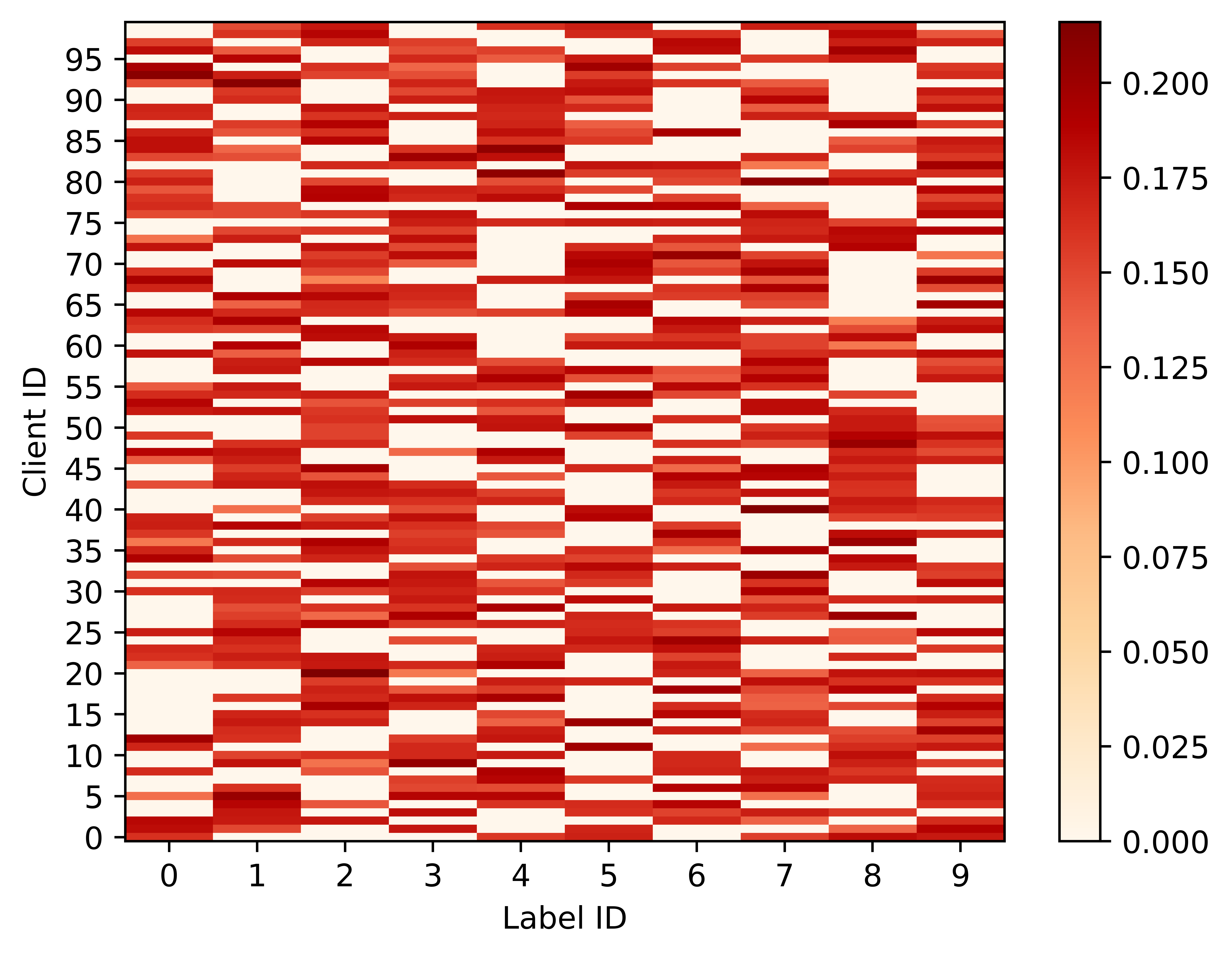}}
 \subcaptionbox{CIFAR10  $\beta=3$}{\includegraphics[width=0.22\textwidth]{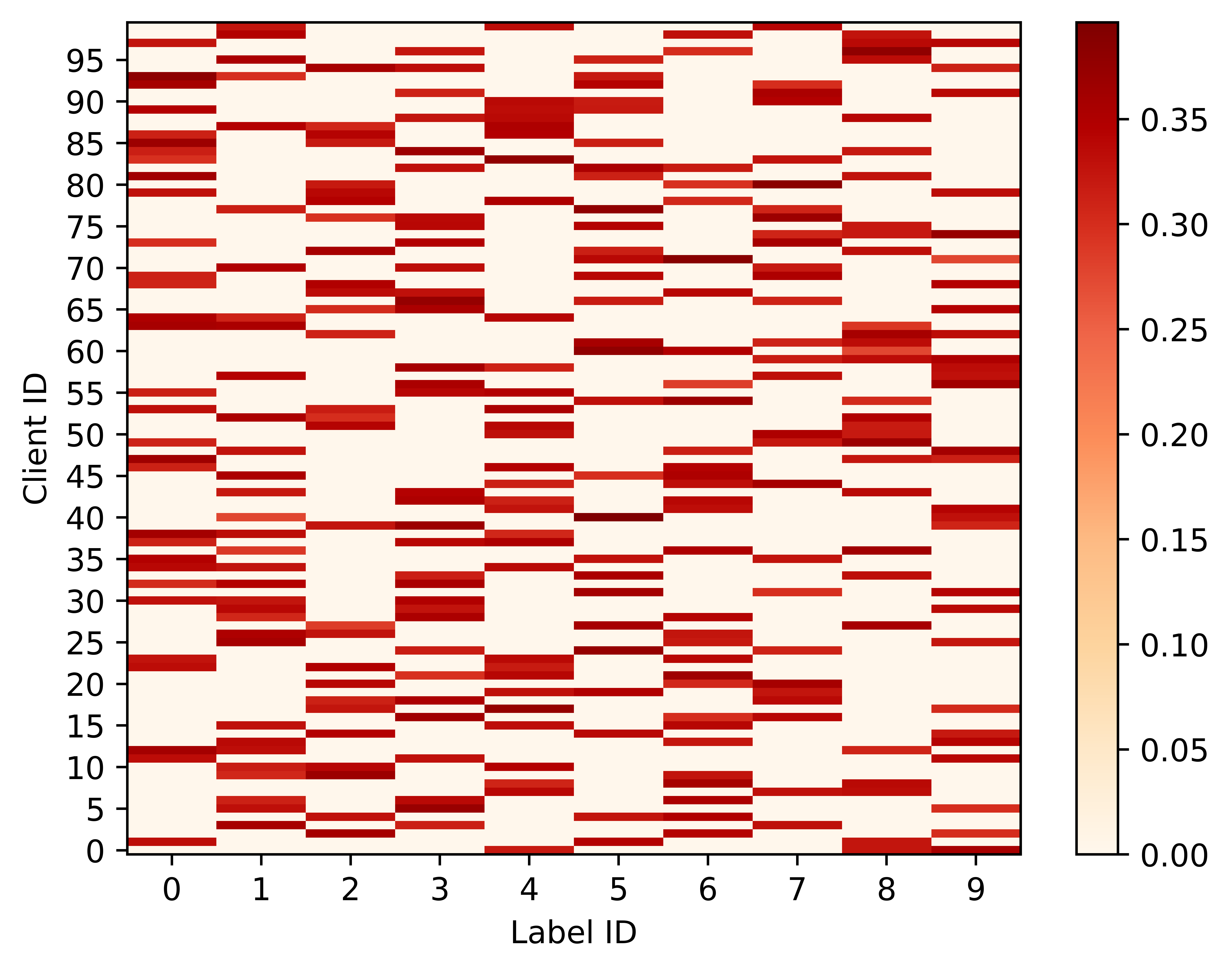}}

\subcaptionbox{CIFAR100  $\alpha=0.6$}{\includegraphics[width=0.22\textwidth]{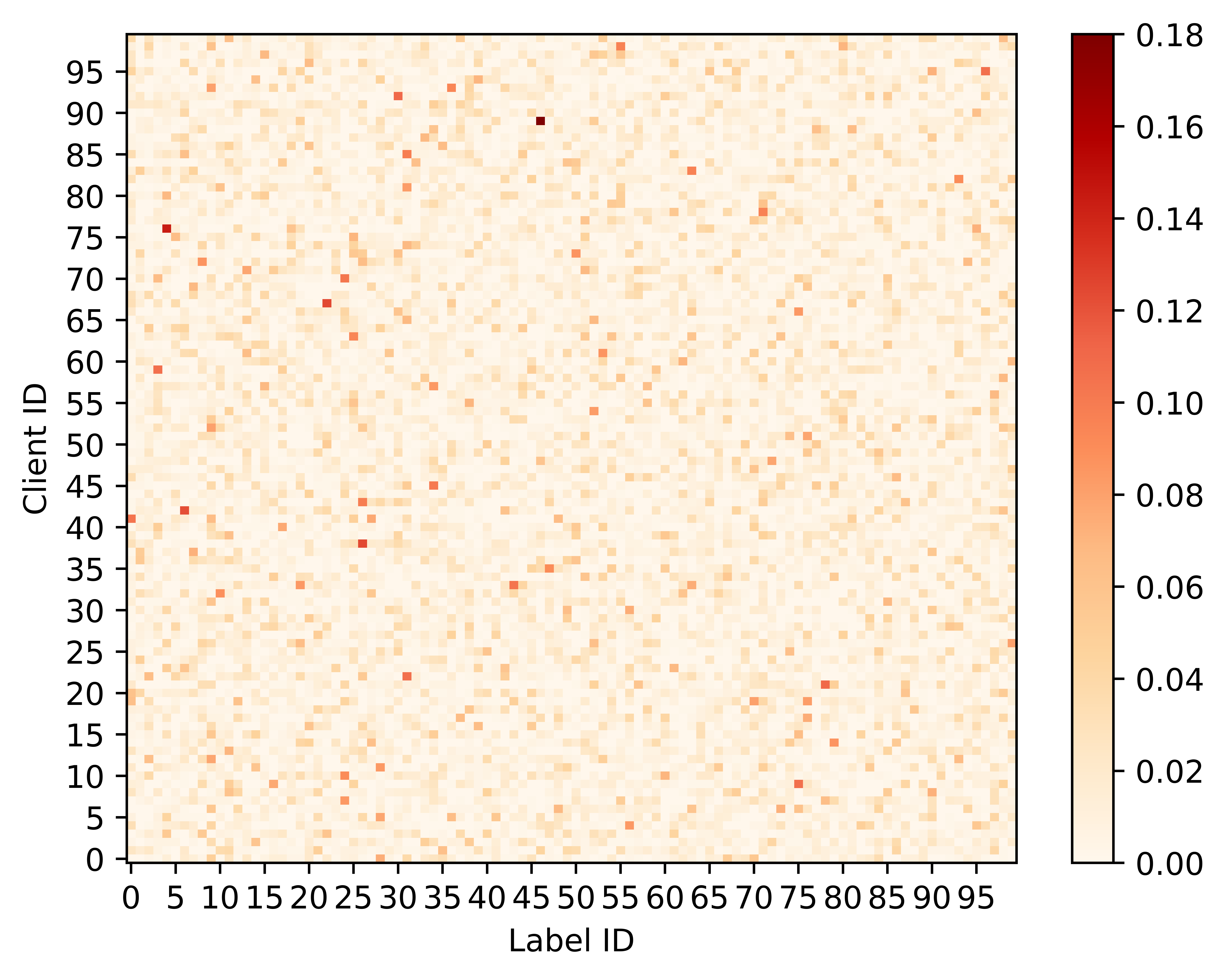}}
 \subcaptionbox{CIFAR100  $\alpha=0.1$}{\includegraphics[width=0.22\textwidth]{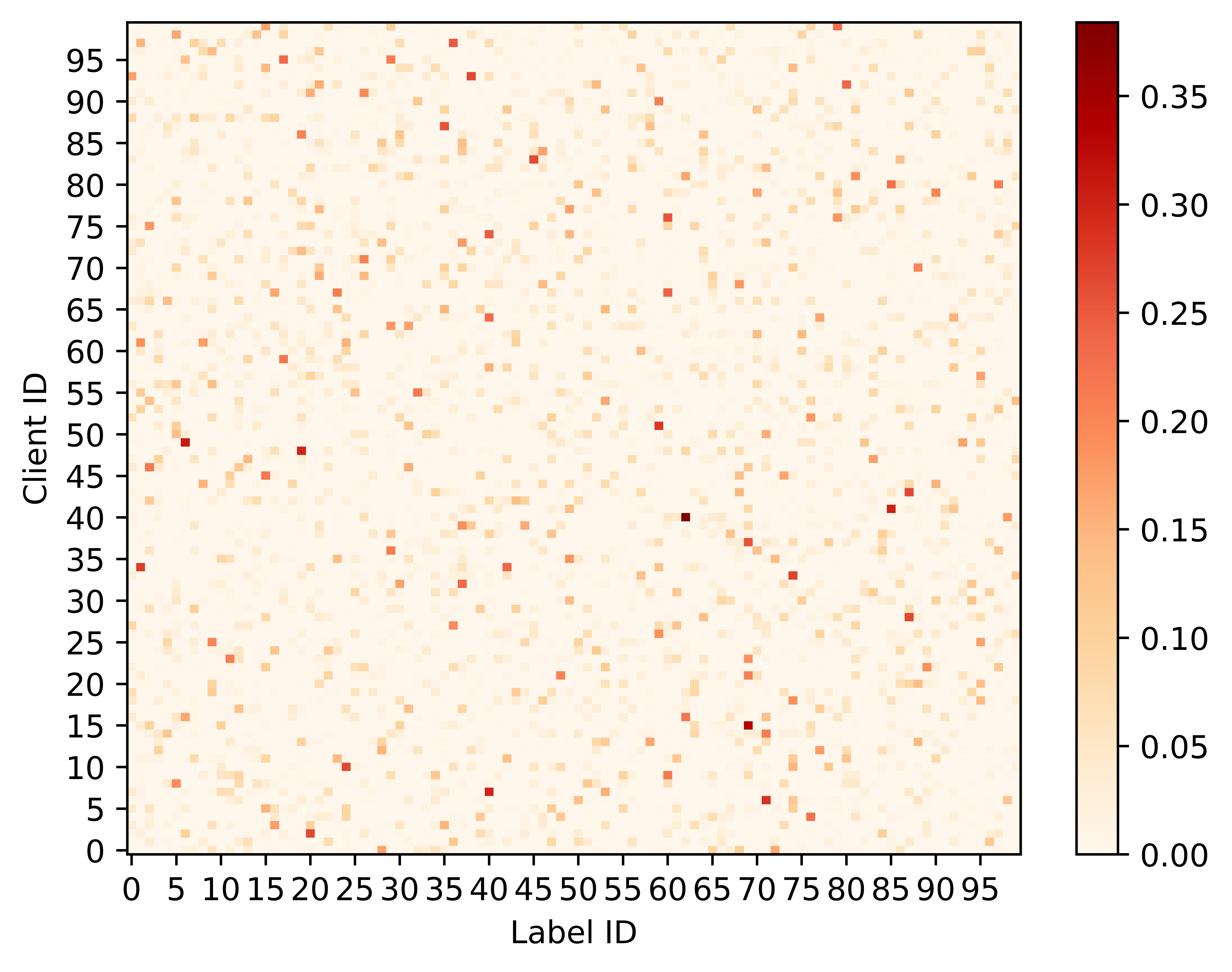}}
 \subcaptionbox{CIFAR100  $\beta=20$}{\includegraphics[width=0.22\textwidth]{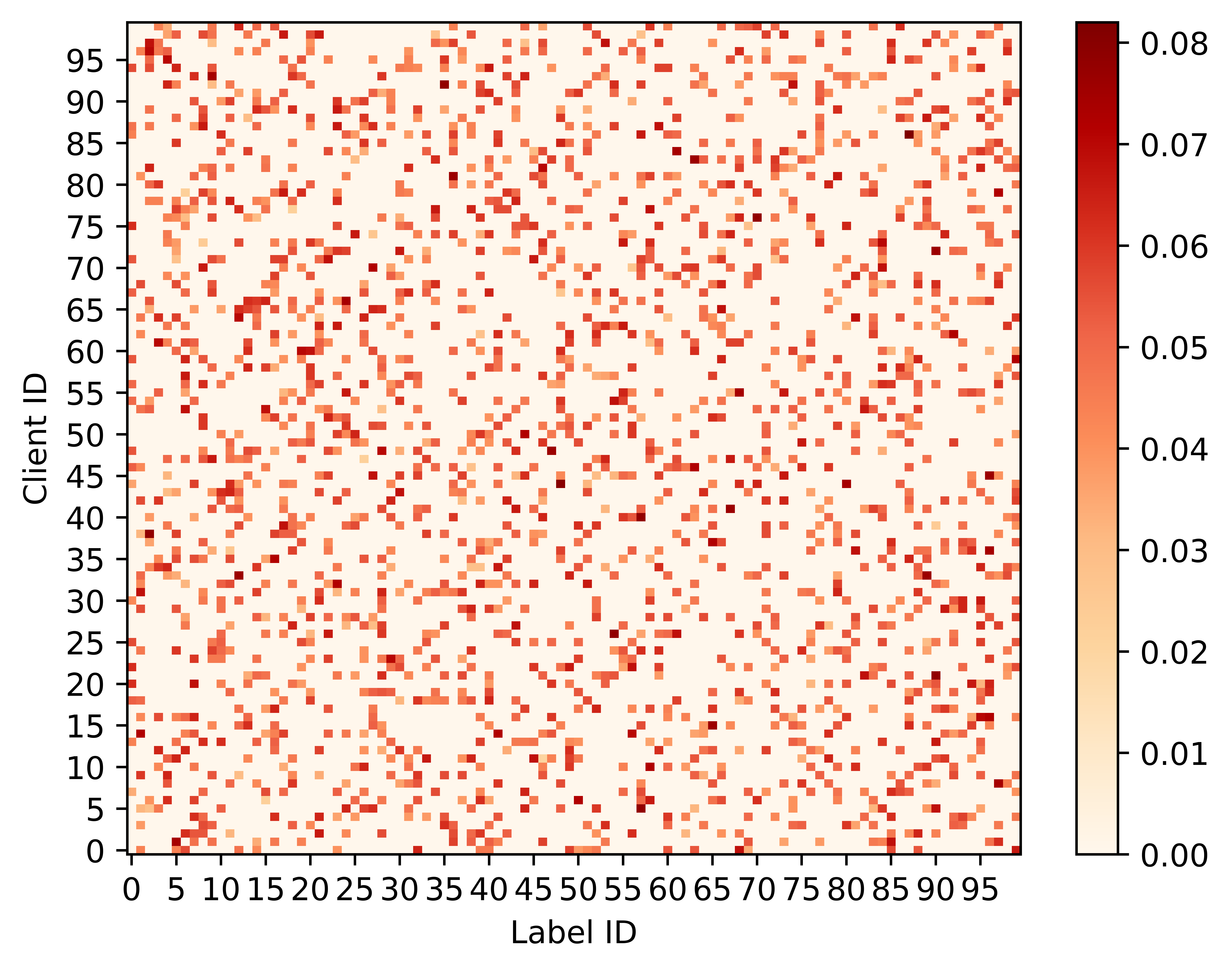}}
 \subcaptionbox{CIFAR100  $\beta=10$}{\includegraphics[width=0.22\textwidth]{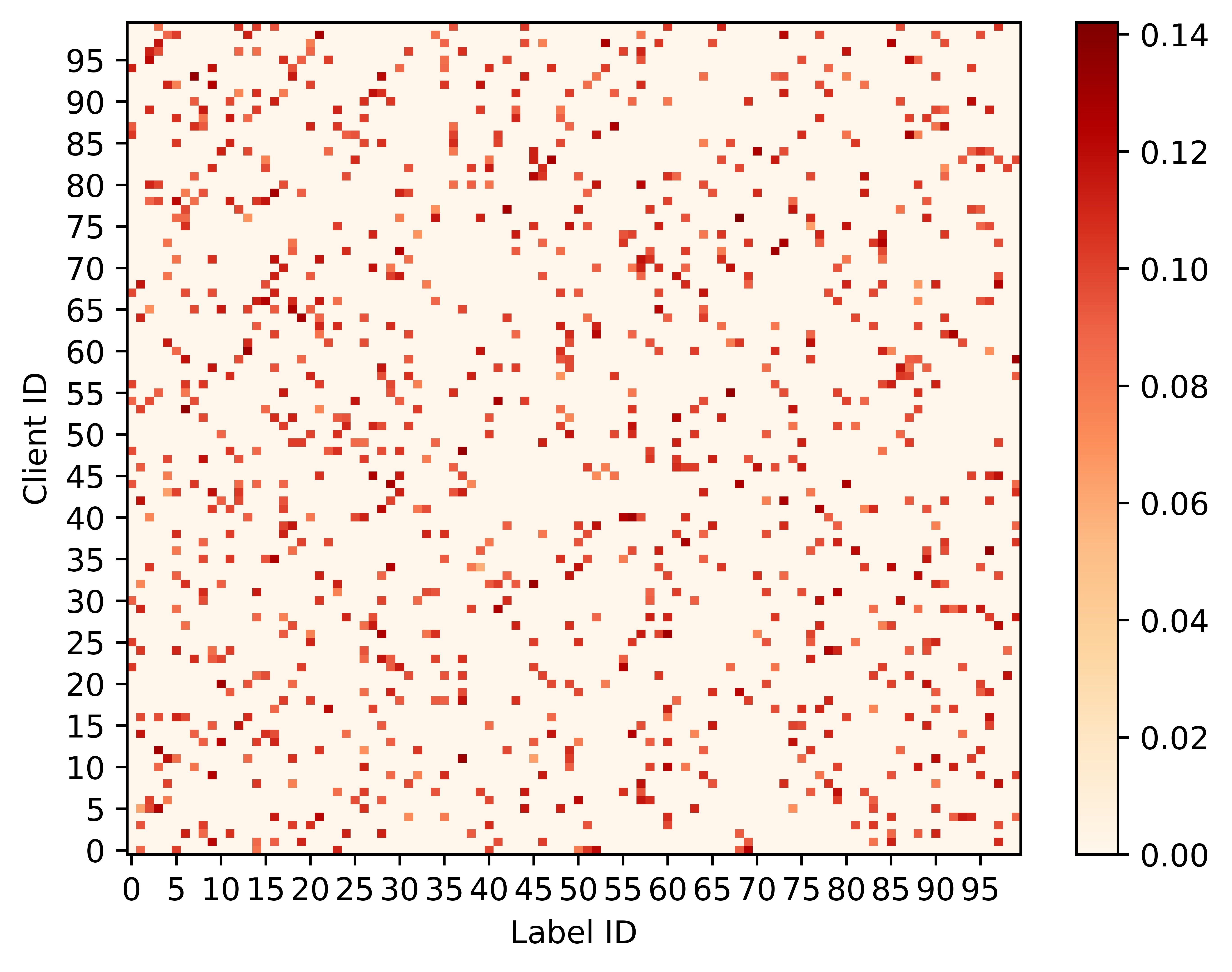}}

  \subcaptionbox{CIFAR10  $\alpha=0.6$}{\includegraphics[width=0.22\textwidth]{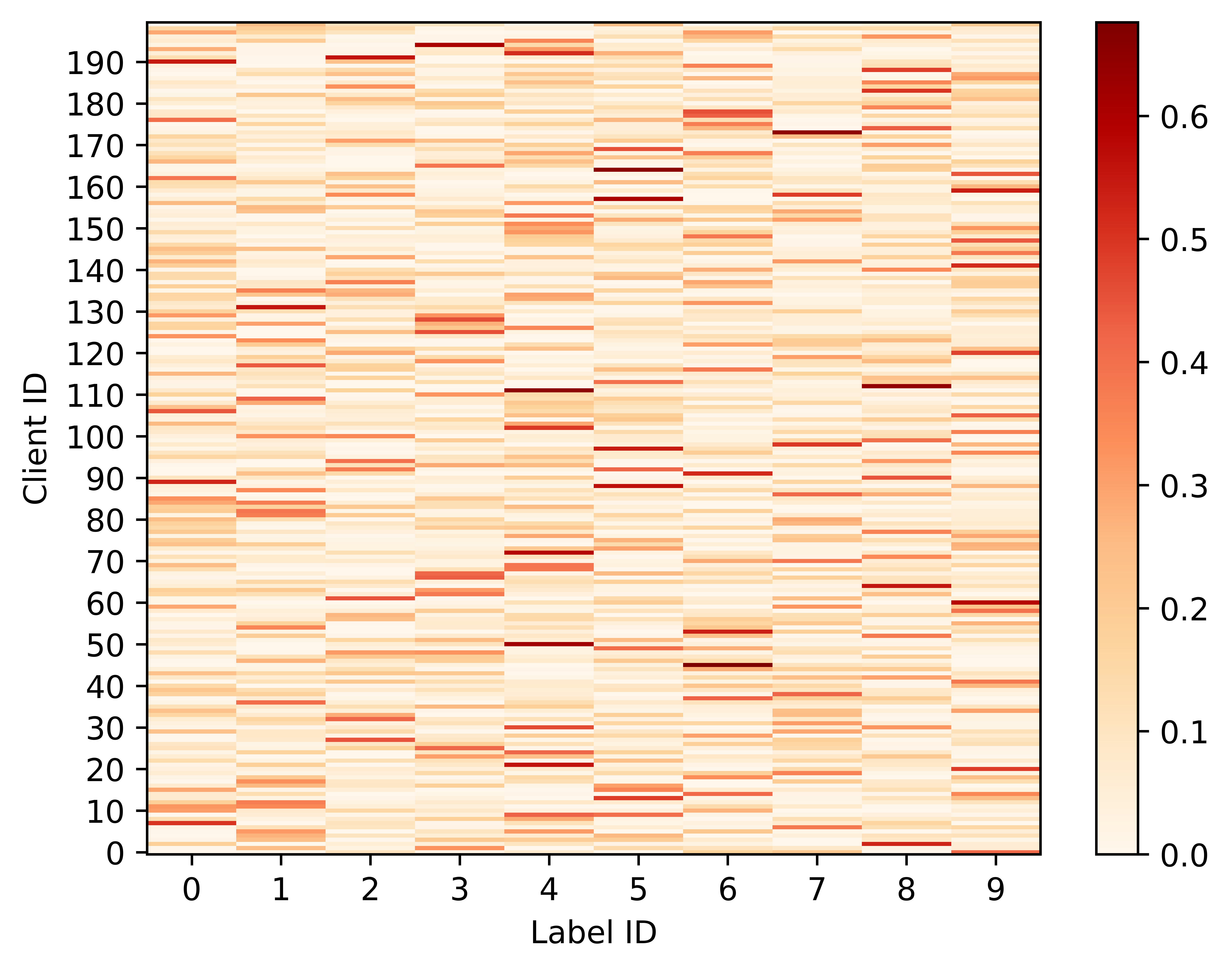}}
 \subcaptionbox{CIFAR10  $\alpha=0.1$}{\includegraphics[width=0.22\textwidth]{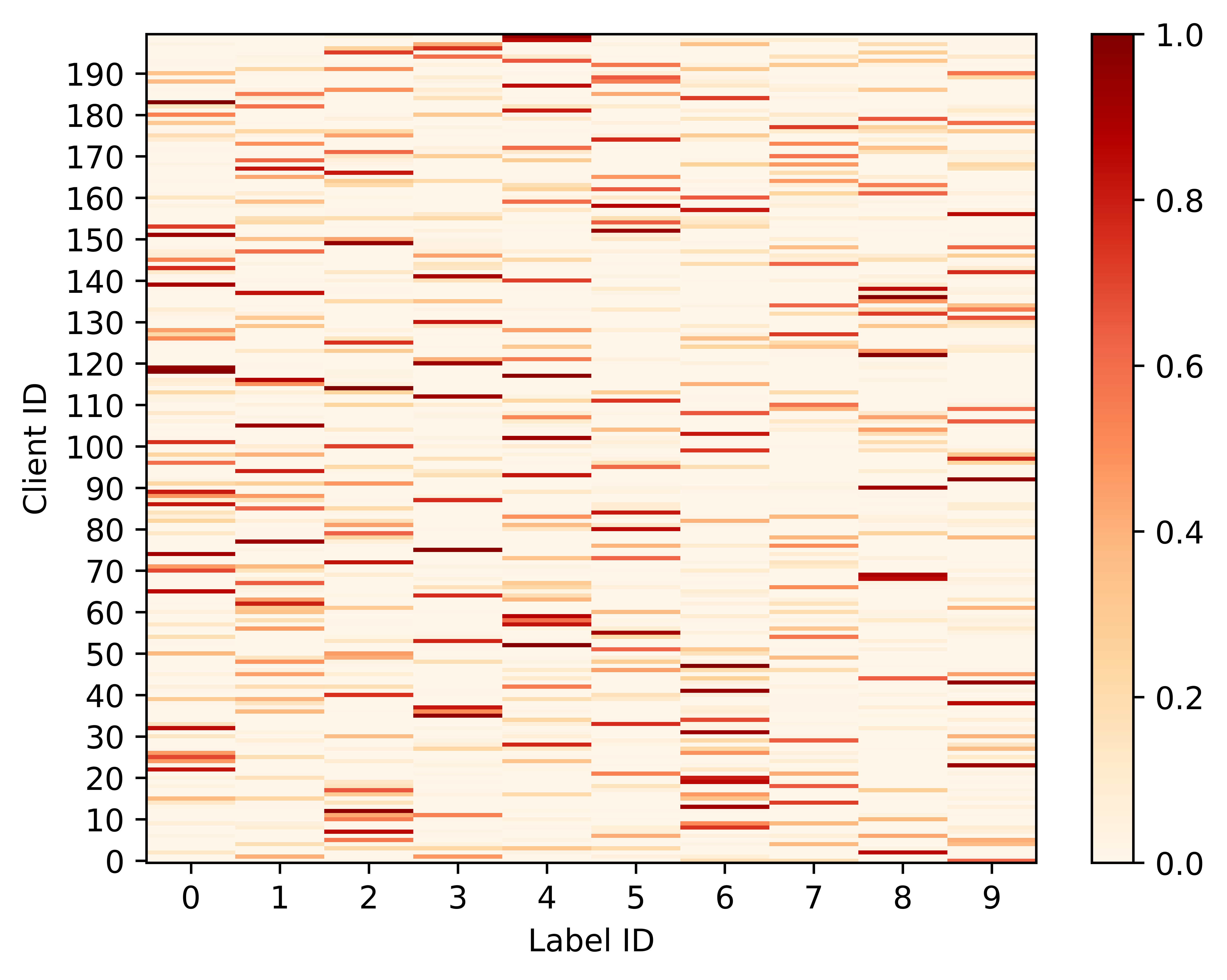}}
 \subcaptionbox{CIFAR10  $\beta=6$}{\includegraphics[width=0.22\textwidth]{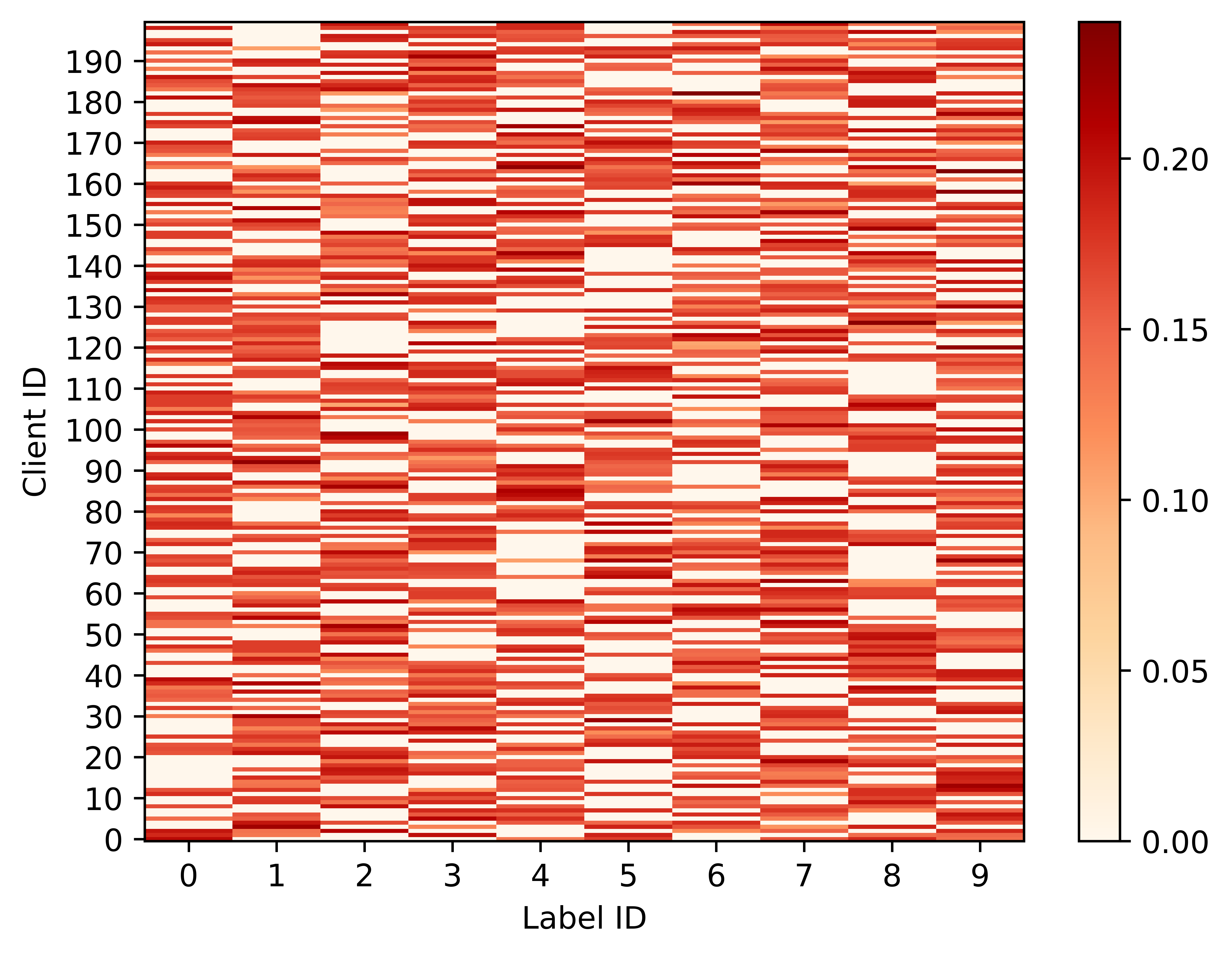}}
 \subcaptionbox{CIFAR10  $\beta=3$}{\includegraphics[width=0.22\textwidth]{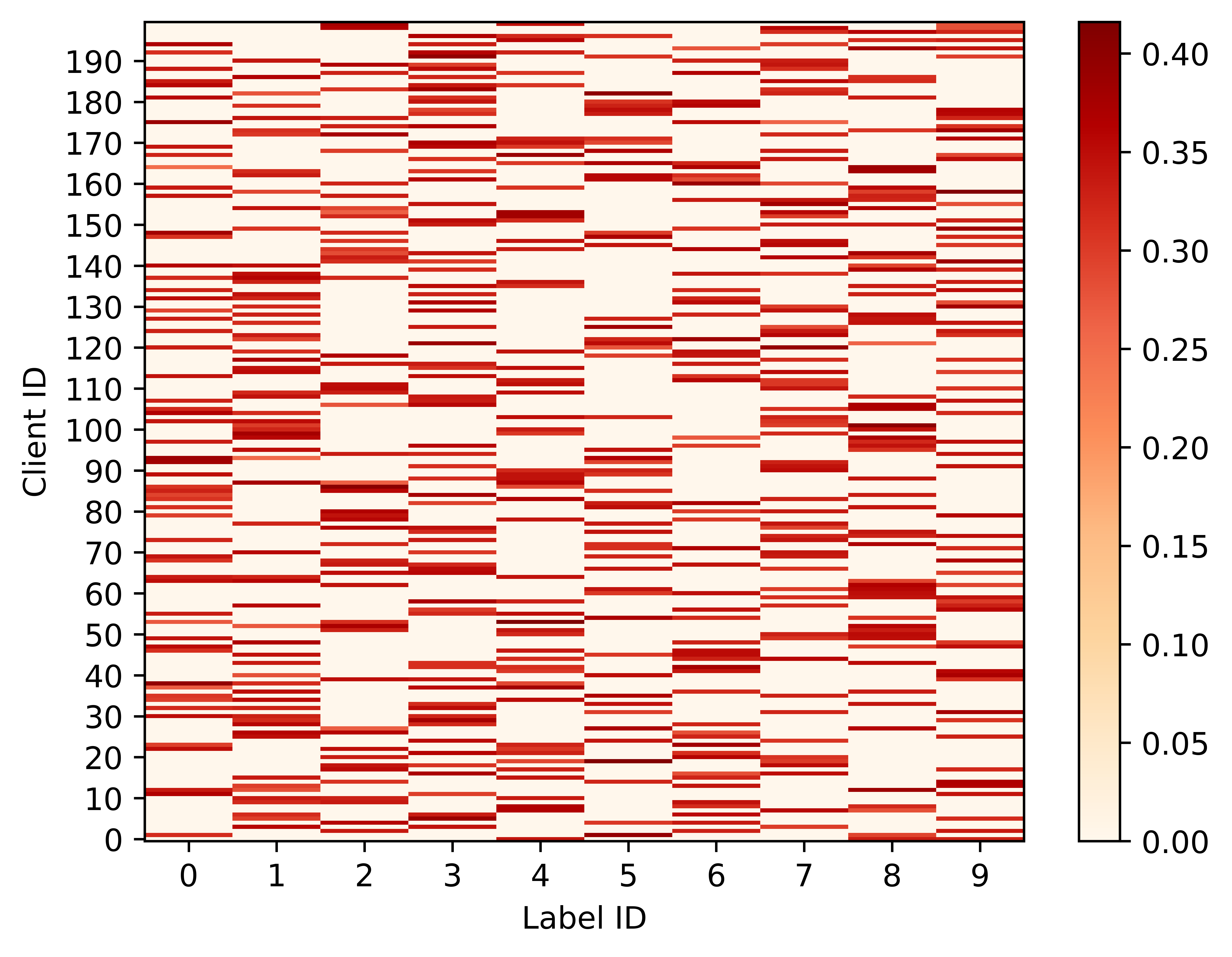}}

\subcaptionbox{CIFAR100  $\alpha=0.6$}{\includegraphics[width=0.22\textwidth]{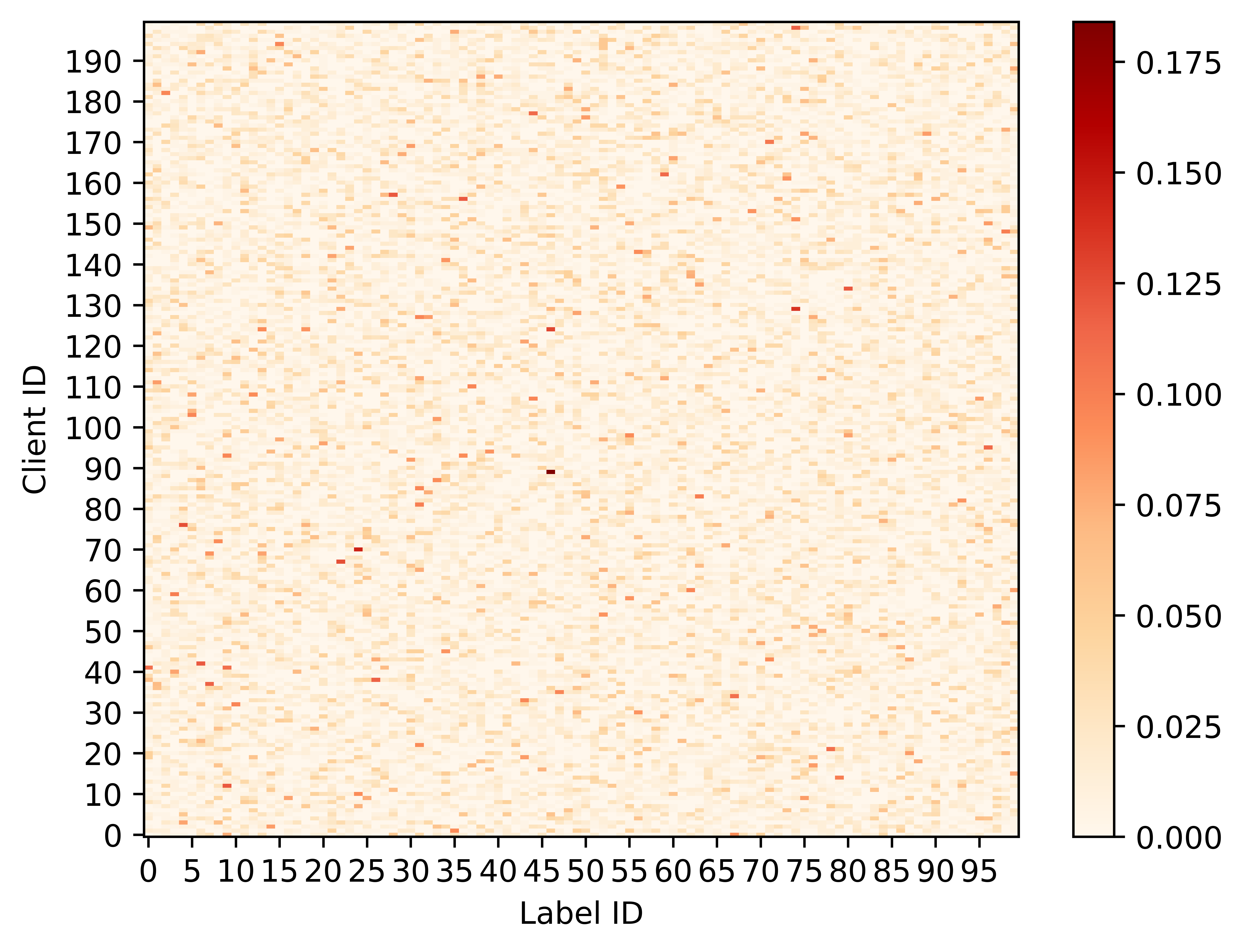}}
 \subcaptionbox{CIFAR100  $\alpha=0.1$}{\includegraphics[width=0.22\textwidth]{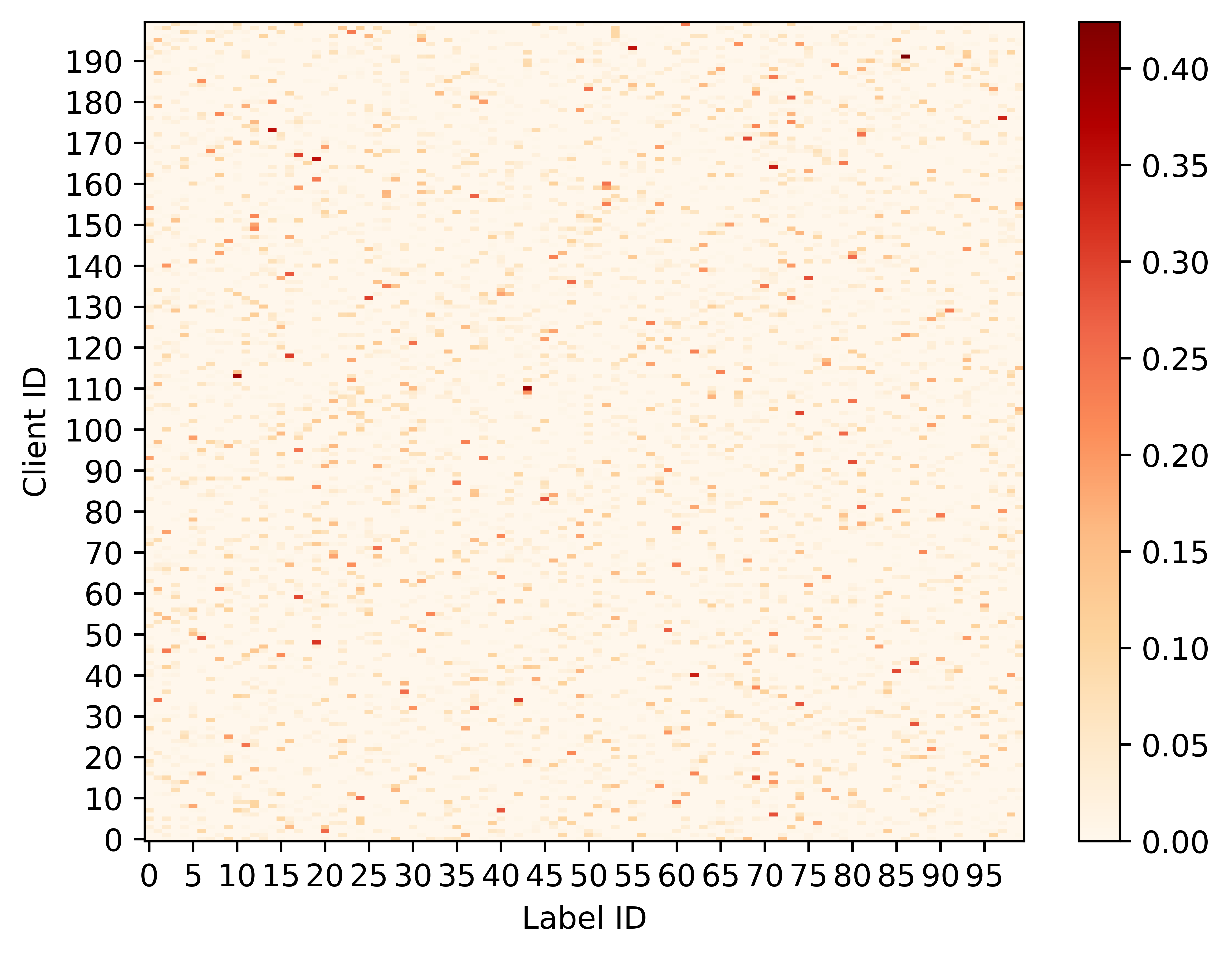}}
 \subcaptionbox{CIFAR100  $\beta=20$}{\includegraphics[width=0.22\textwidth]{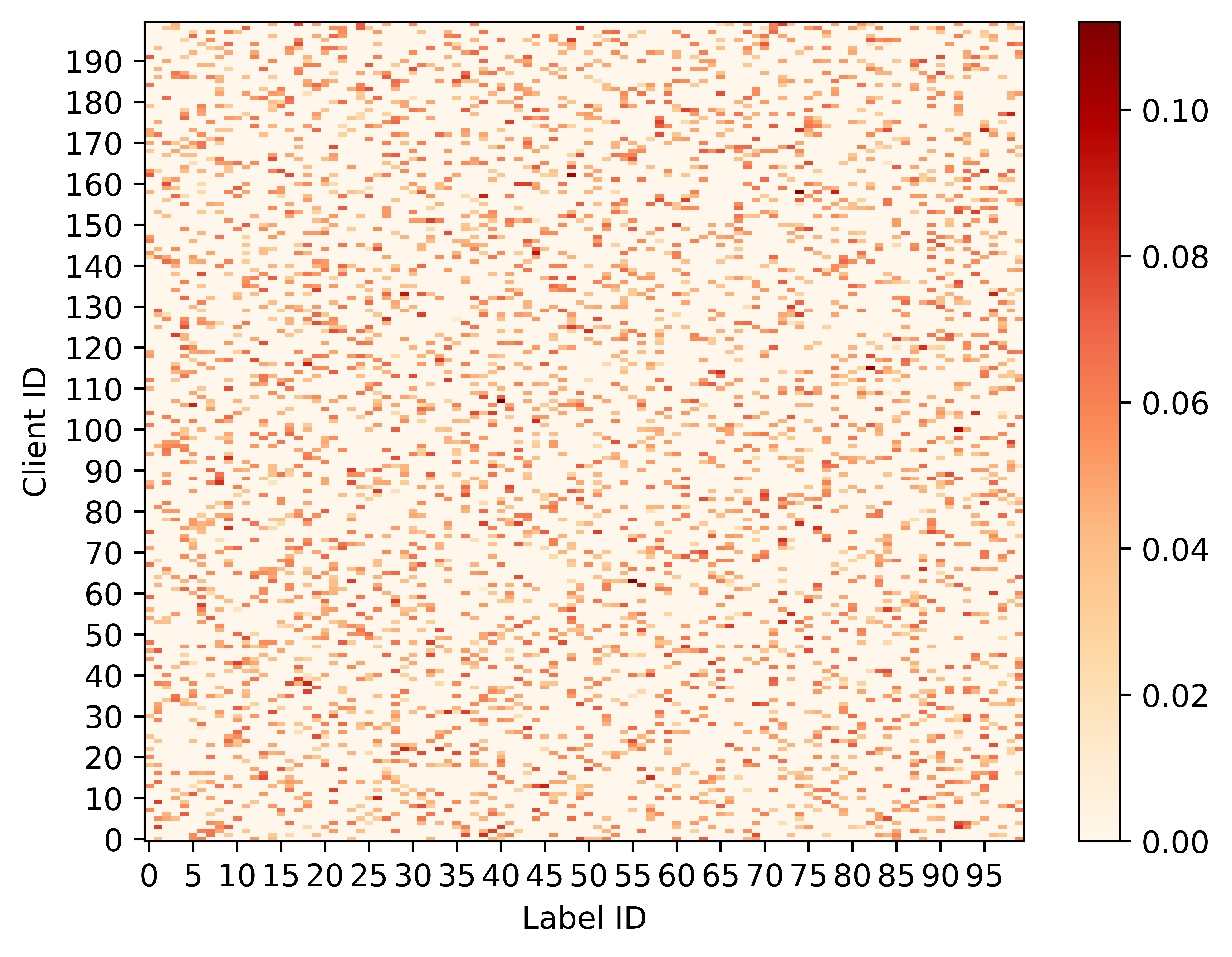}}
 \subcaptionbox{CIFAR100  $\beta=10$}{\includegraphics[width=0.22\textwidth]{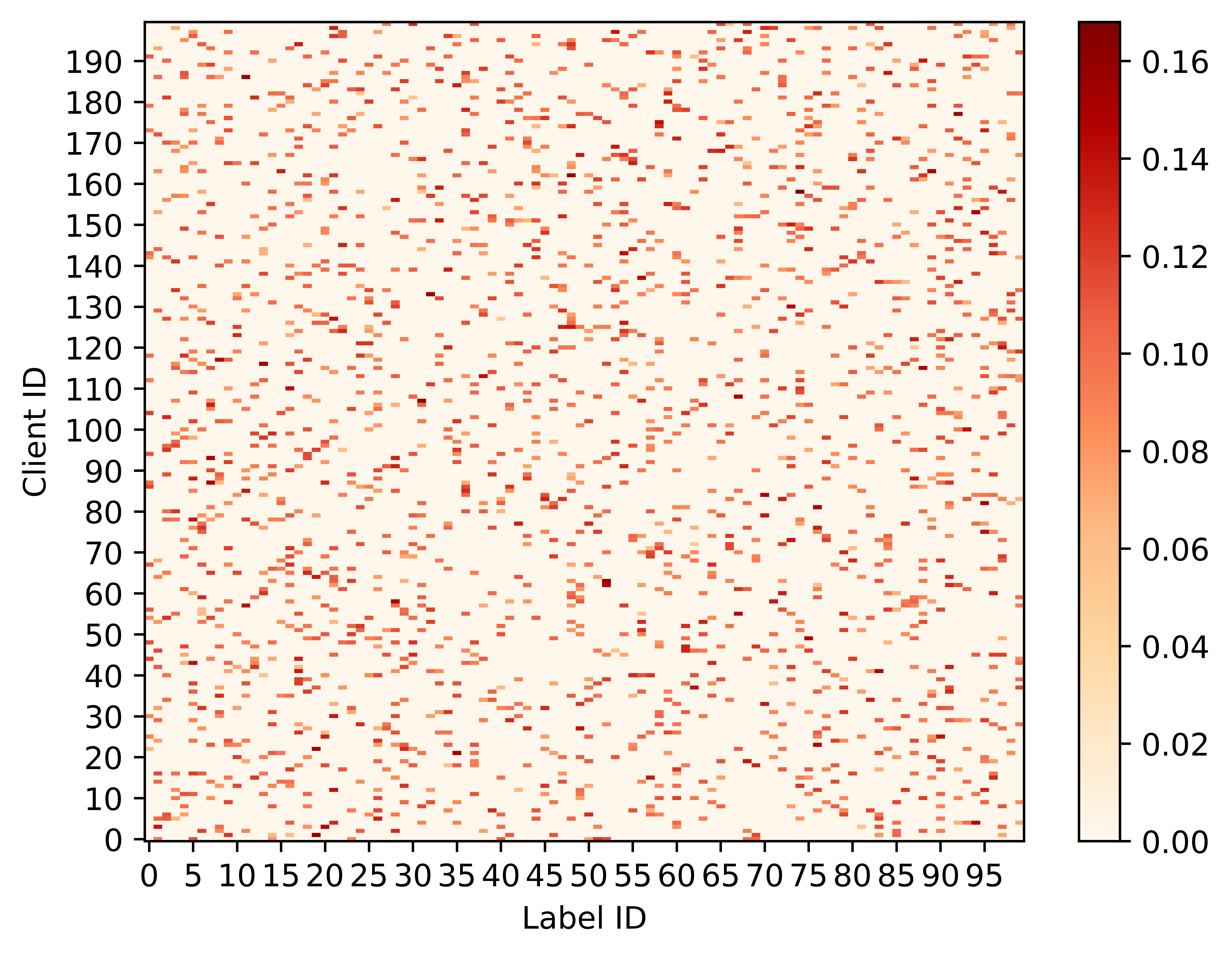}}


\caption{Heatmaps of the data distributions for ClAR10 and ClFAR100 for Dirichlet distributions with coefficients of 0.6 and 0.1, respectively, and for Pathological sampling probabilities with coefficients of 6/20 and 3/10. Both datasets consistently include 100 / 200 clients.}
\label{distribution_heatmap}
\end{figure*}

\noindent\textbf{Dirichlet Sampling}: The Dirichlet distribution can be thought of as the conjugate prior of a polynomial distribution, and is used to generate weights for a mixture model or to distribute samples in the context of a non-uniform category distribution. By adjusting the parameter $u$, it is possible to generate data ranging from extremely inhomogeneous (near-discrete concentration in a category) to uniformly distributed. The data exhibit a long-tailed distribution, see Fig. \ref{distribution_heatmap}.

\noindent\textbf{Pathological Sampling}: A typical feature of pathological sampling is extreme skewness or anomalies in the data distribution, which may lead to unstable training, convergence difficulties, or severe degradation of model performance. We used it to test and validate the performance of the model under adverse conditions. The data are presented in species isolation, see Fig.\ref{distribution_heatmap}.

The splitting strategy for all data is consistent with FedGAMMA\cite{10269141}, FedSMOO\cite{sun2023dynamic}, and FedLESAM\cite{FedLESAM}.

\newpage
\subsection{Evaluation Curves}

\begin{figure*}[htb]
\centering 
 \subcaptionbox{Accuracy $u=0.6$}{\includegraphics[width=0.24\textwidth]{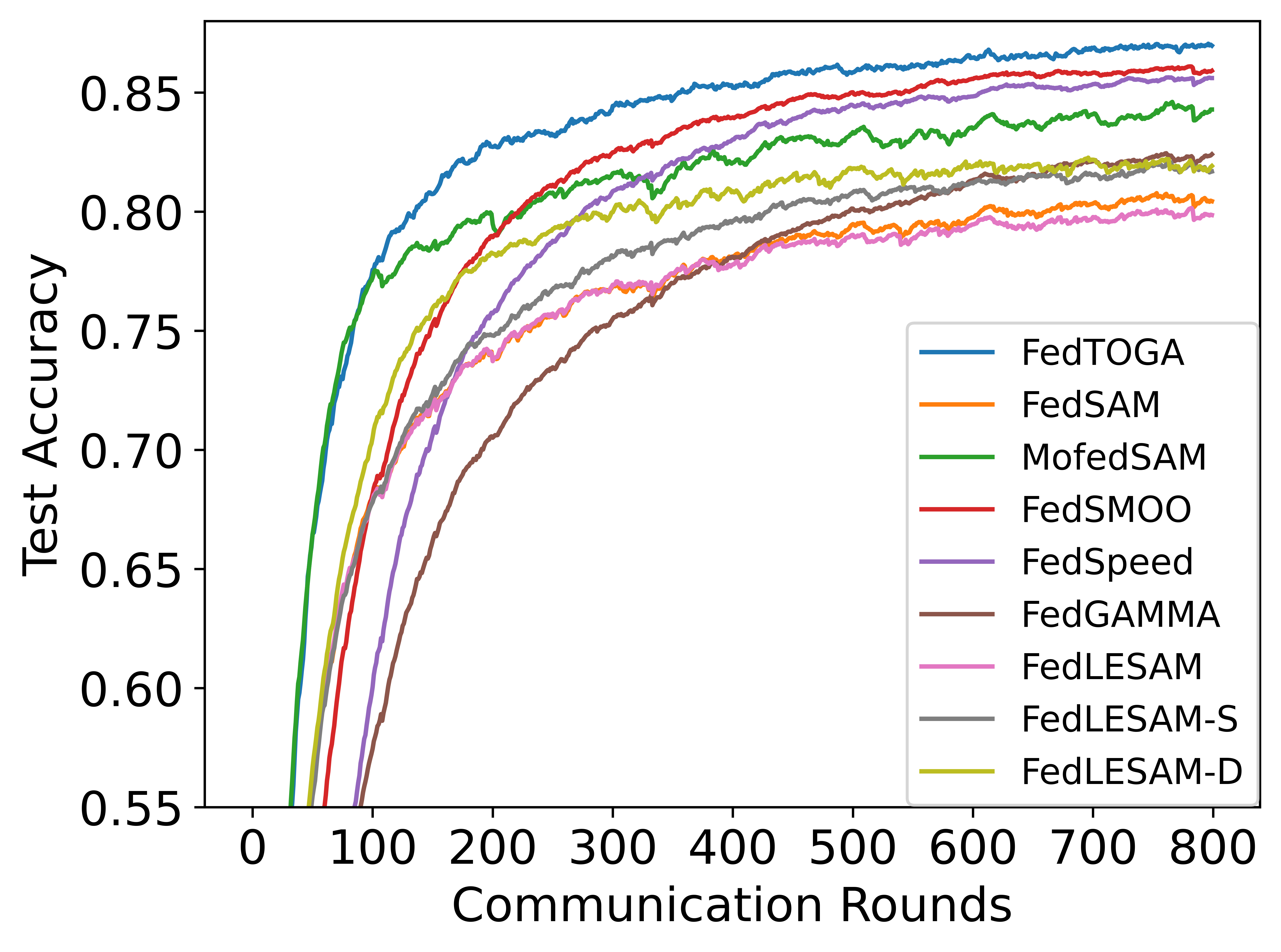}}
 \subcaptionbox{Accuracy $u=0.1$}{\includegraphics[width=0.24\textwidth]{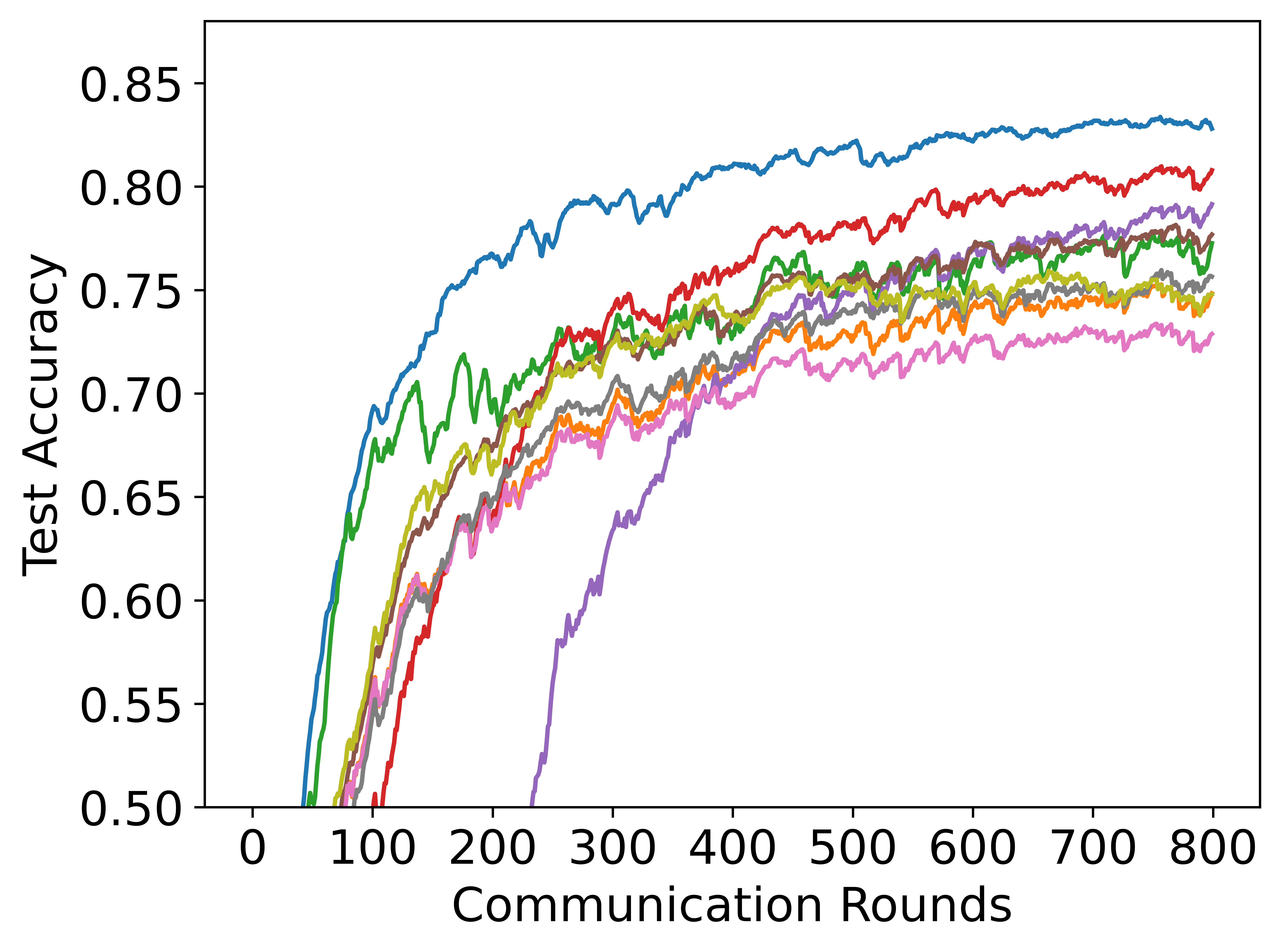}}
 \subcaptionbox{Accuracy $c=6$}{\includegraphics[width=0.24\textwidth]{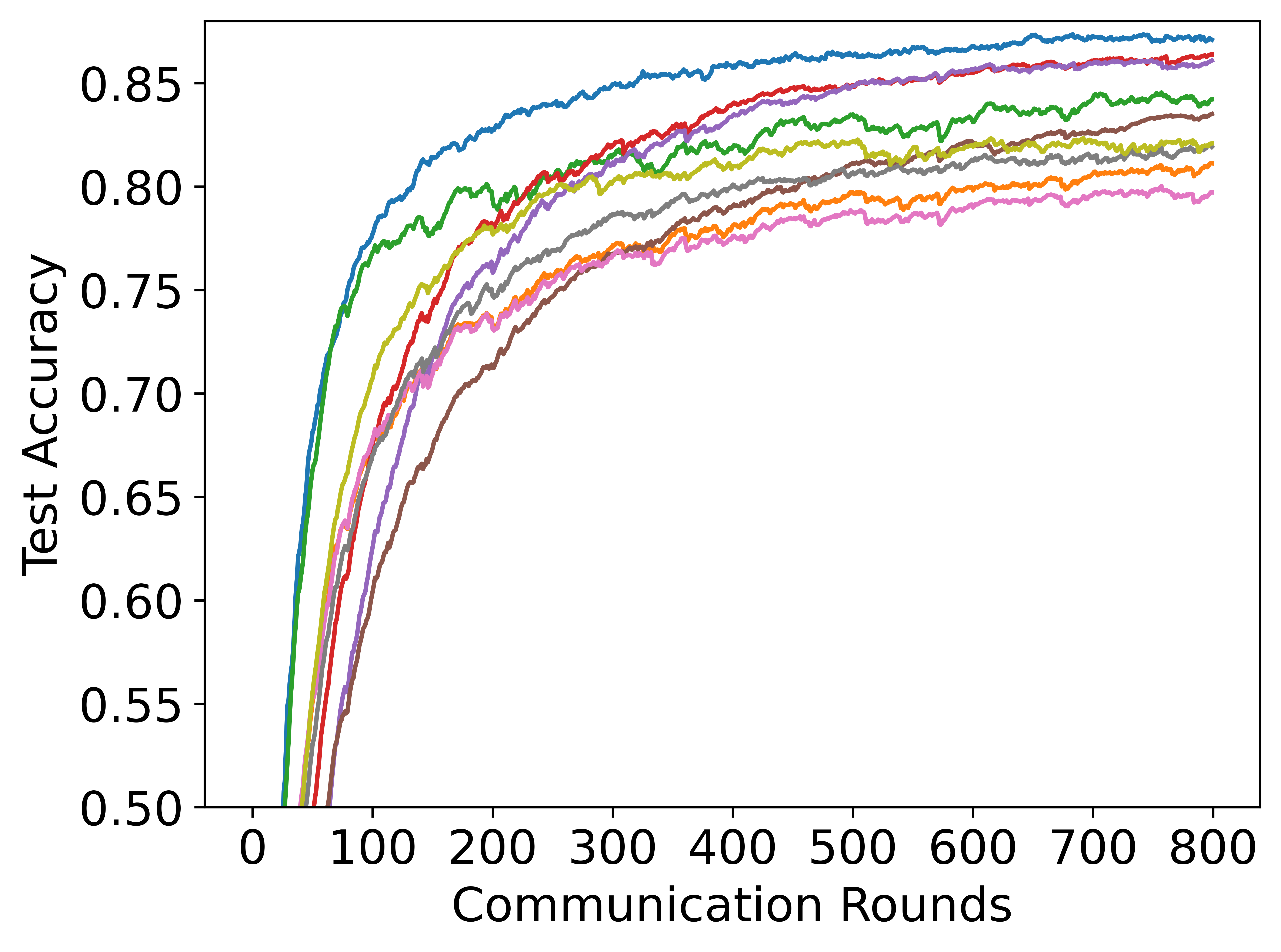}}
 \subcaptionbox{Accuracy $c=3$}{\includegraphics[width=0.24\textwidth]{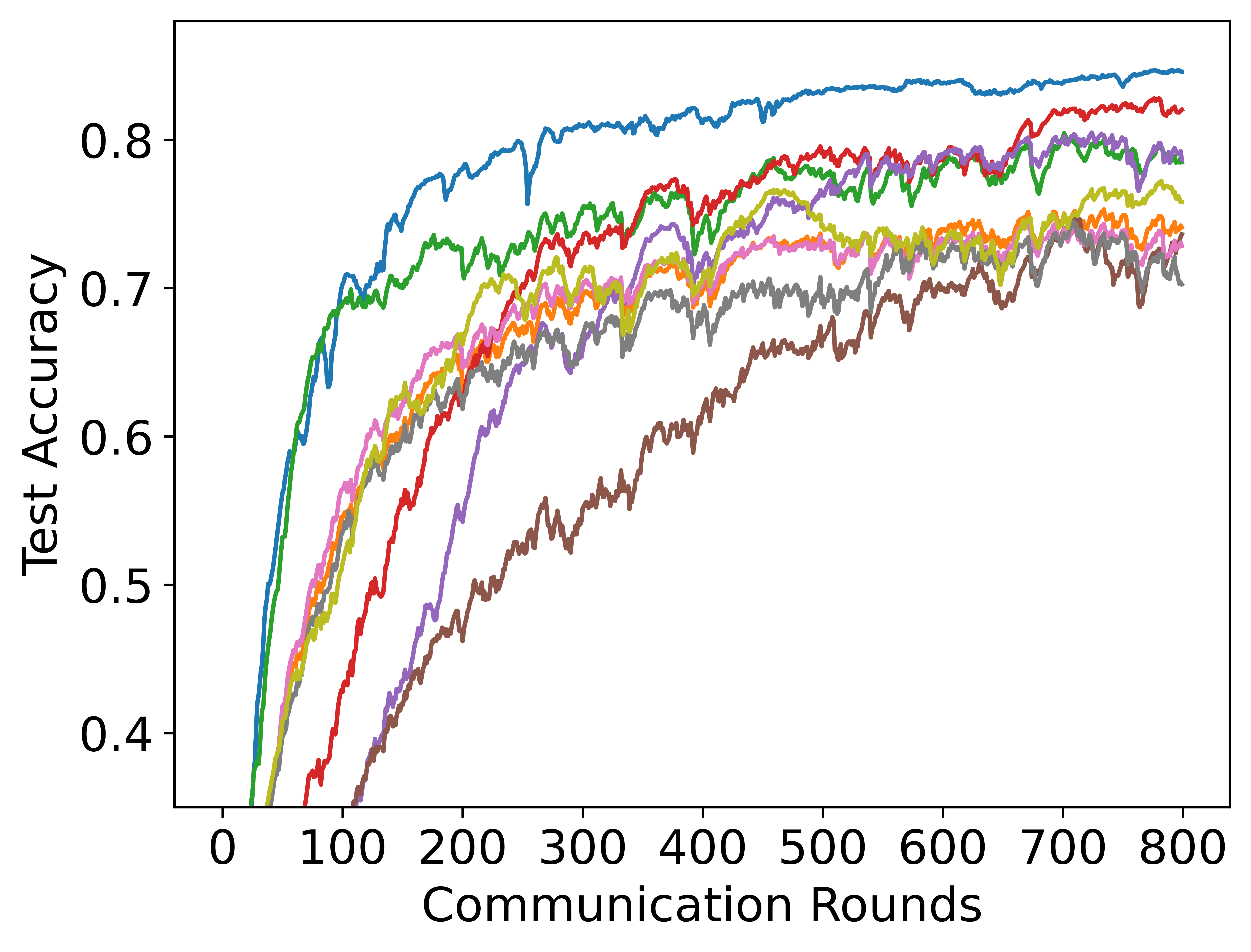}}

  \subcaptionbox{Loss  $u=0.6$}{\includegraphics[width=0.24\textwidth]{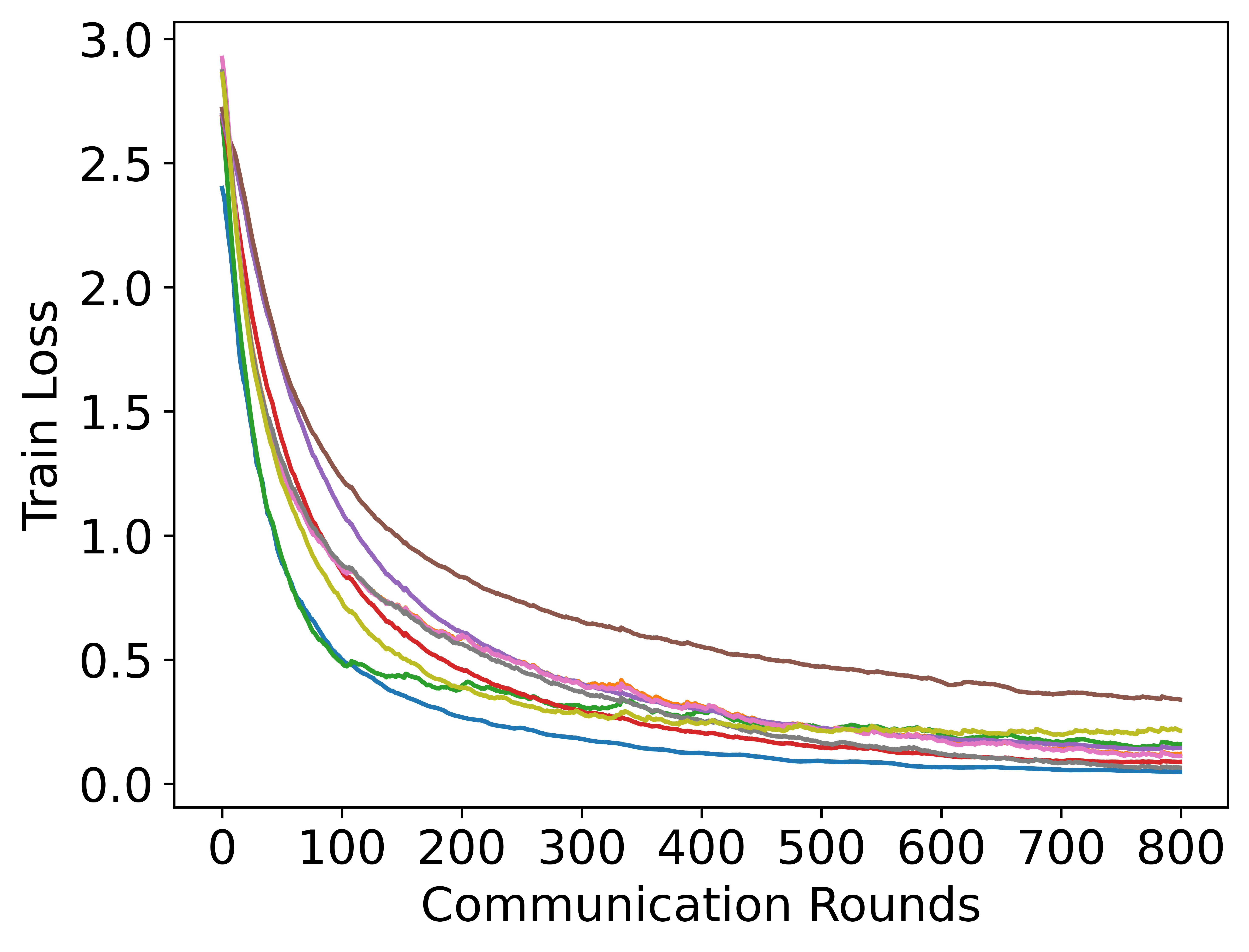}}
 \subcaptionbox{Loss  $u=0.1$}{\includegraphics[width=0.24\textwidth]{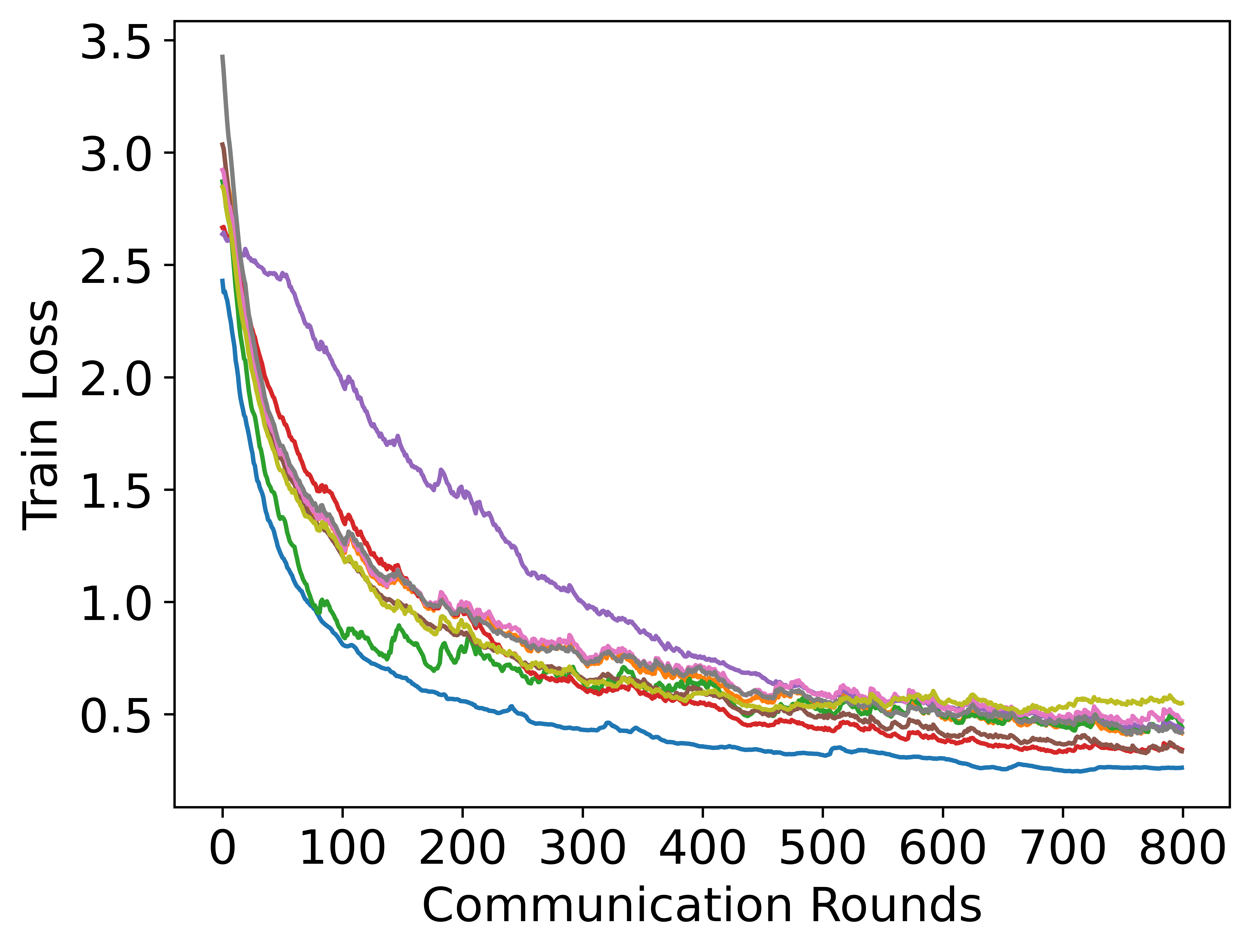}}
 \subcaptionbox{Loss  $c=6$}{\includegraphics[width=0.24\textwidth]{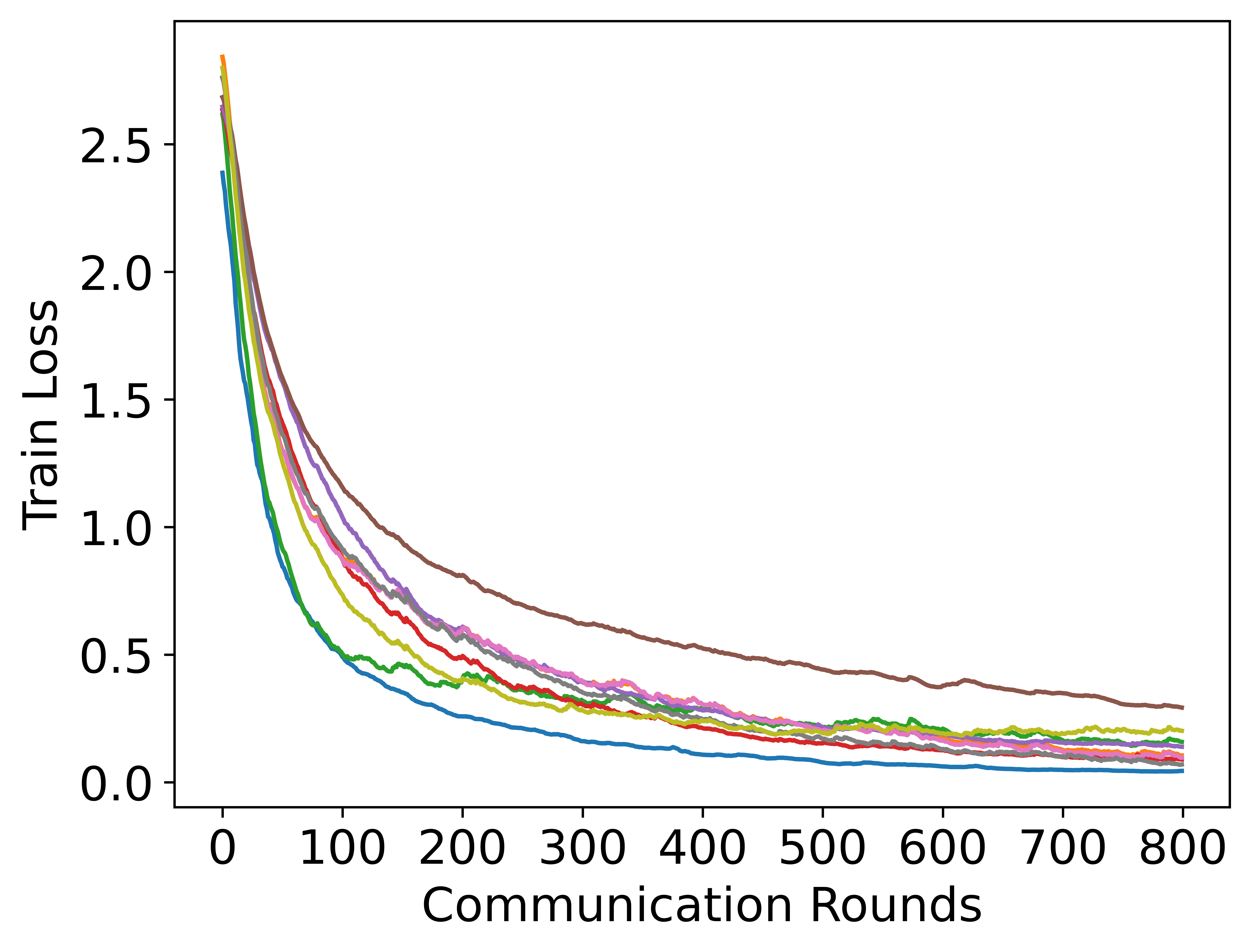}}
 \subcaptionbox{Loss  $c=3$}{\includegraphics[width=0.24\textwidth]{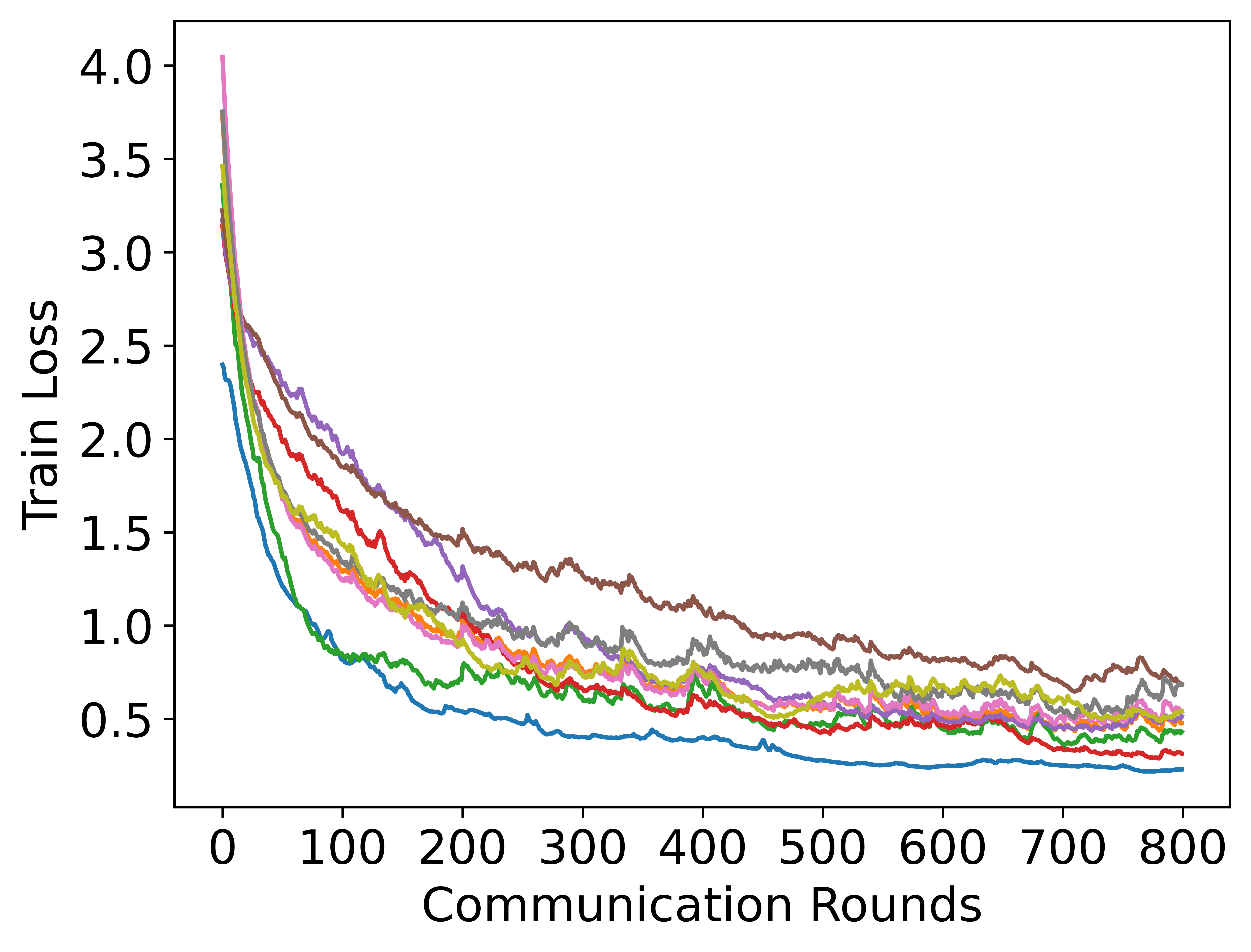}}

  \subcaptionbox{Accuracy  $u=0.6$}{\includegraphics[width=0.24\textwidth]{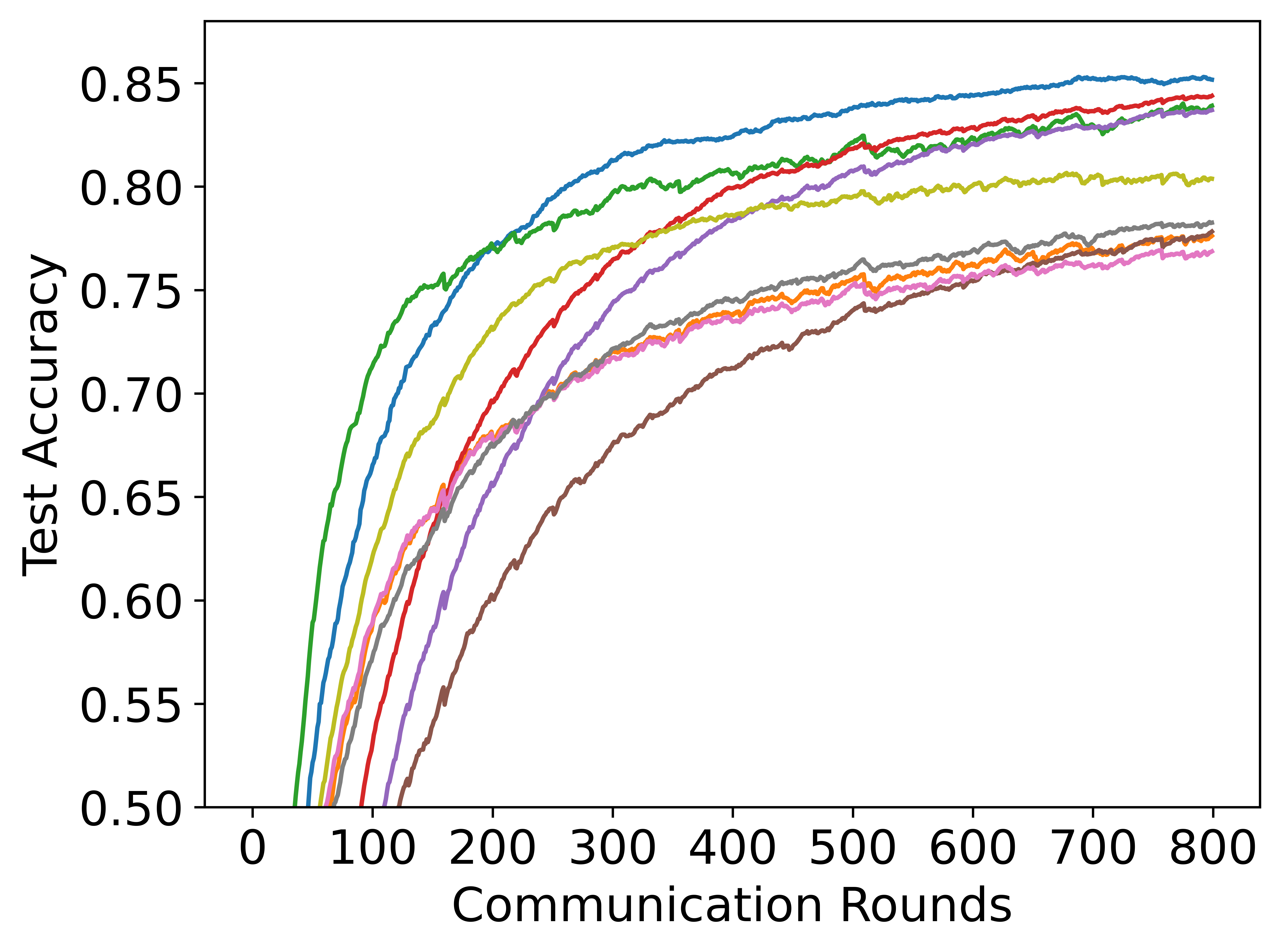}}
 \subcaptionbox{Accuracy  $u=0.1$}{\includegraphics[width=0.24\textwidth]{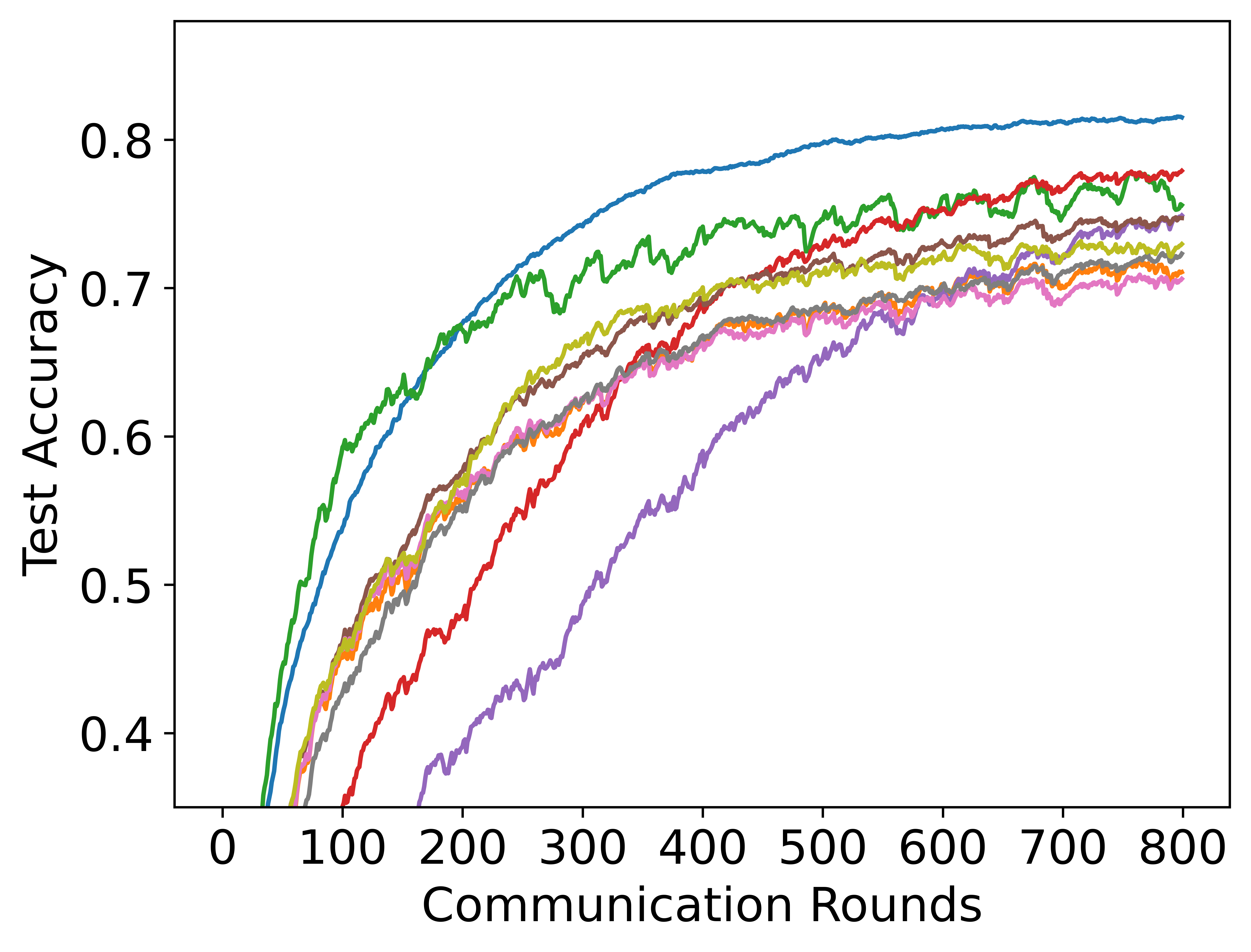}}
 \subcaptionbox{Accuracy  $c=6$}{\includegraphics[width=0.24\textwidth]{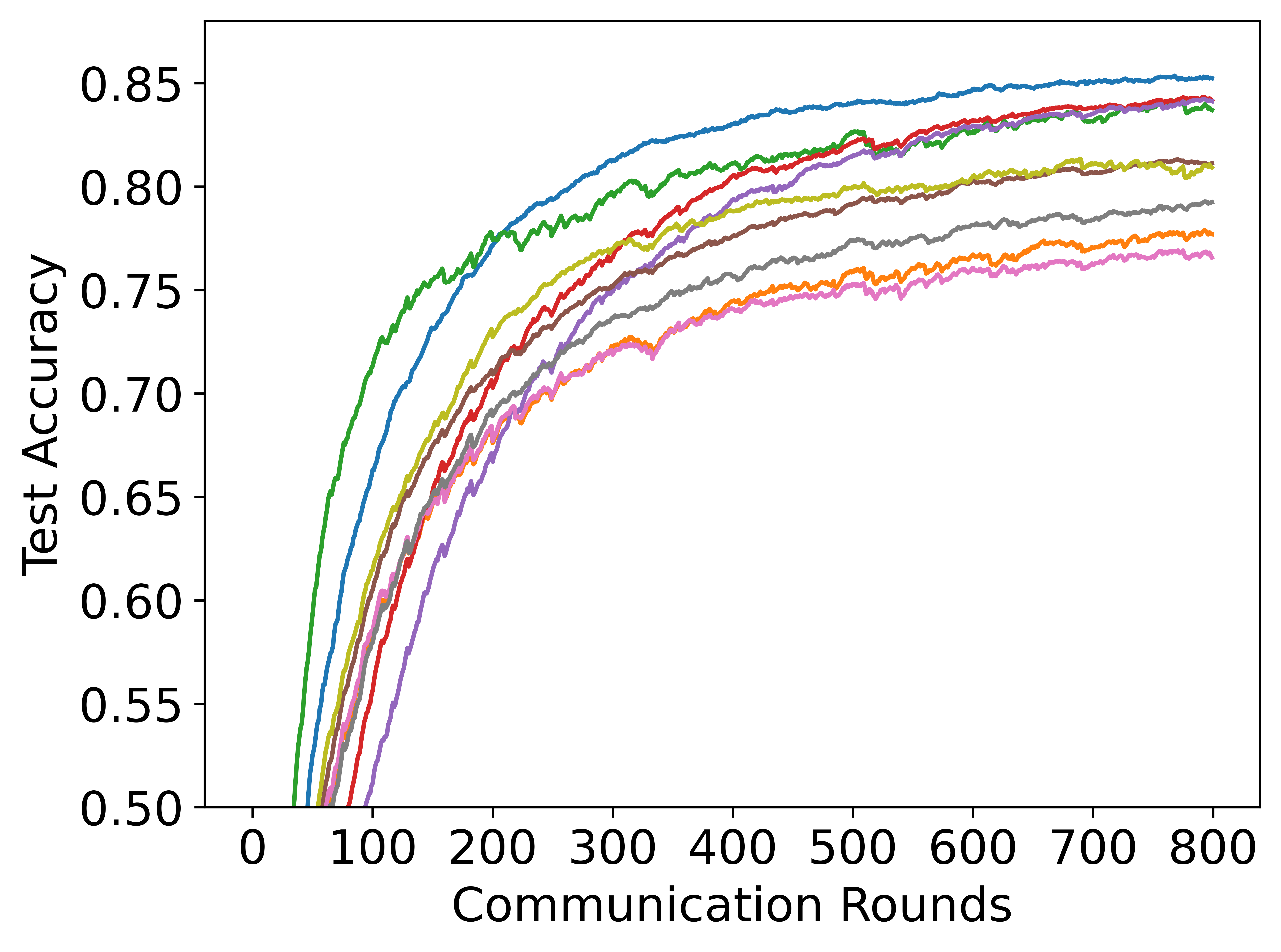}}
 \subcaptionbox{Accuracy  $c=3$}{\includegraphics[width=0.24\textwidth]{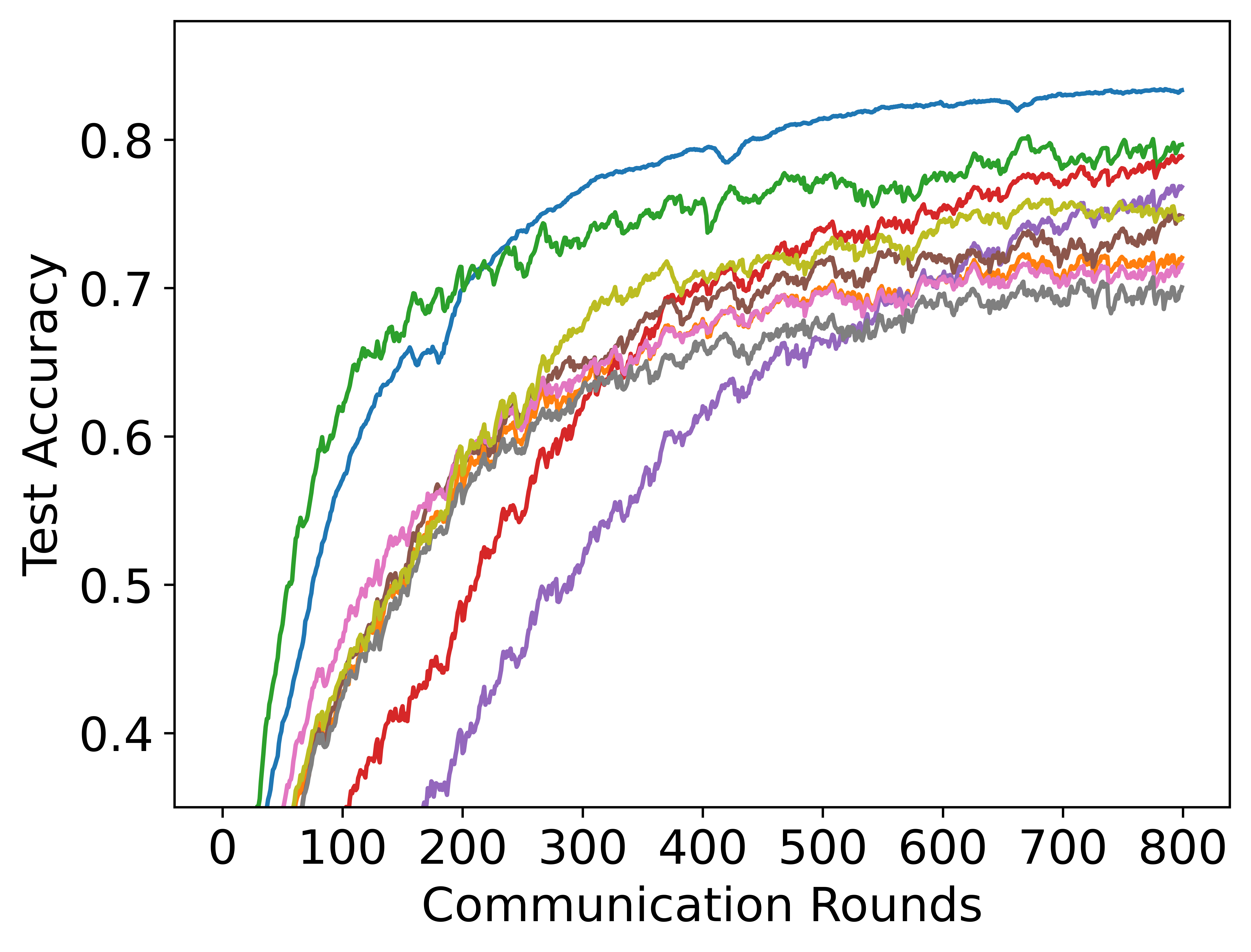}}

  \subcaptionbox{Loss  $u=0.6$}{\includegraphics[width=0.24\textwidth]{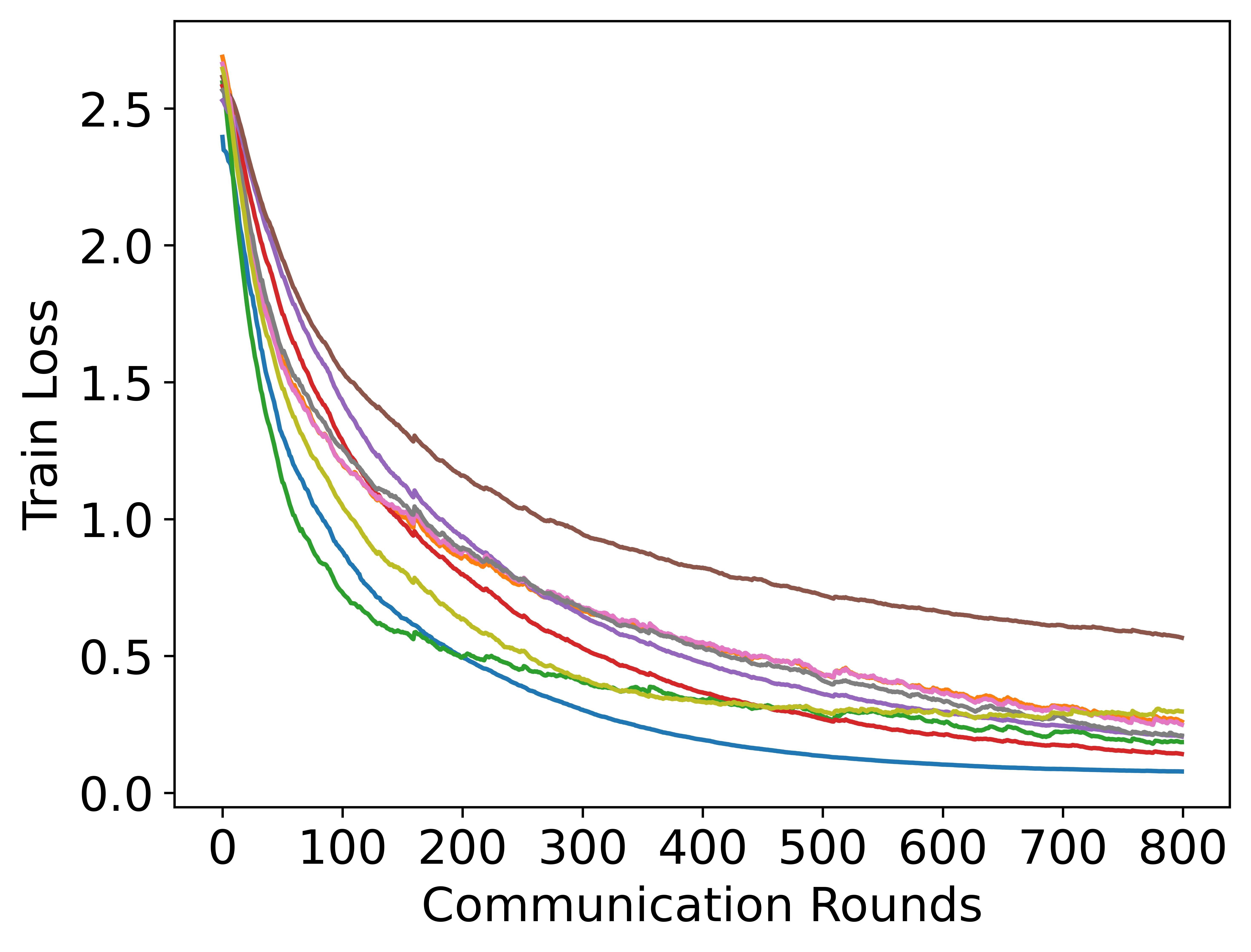}}
 \subcaptionbox{Loss  $u=0.1$}{\includegraphics[width=0.24\textwidth]{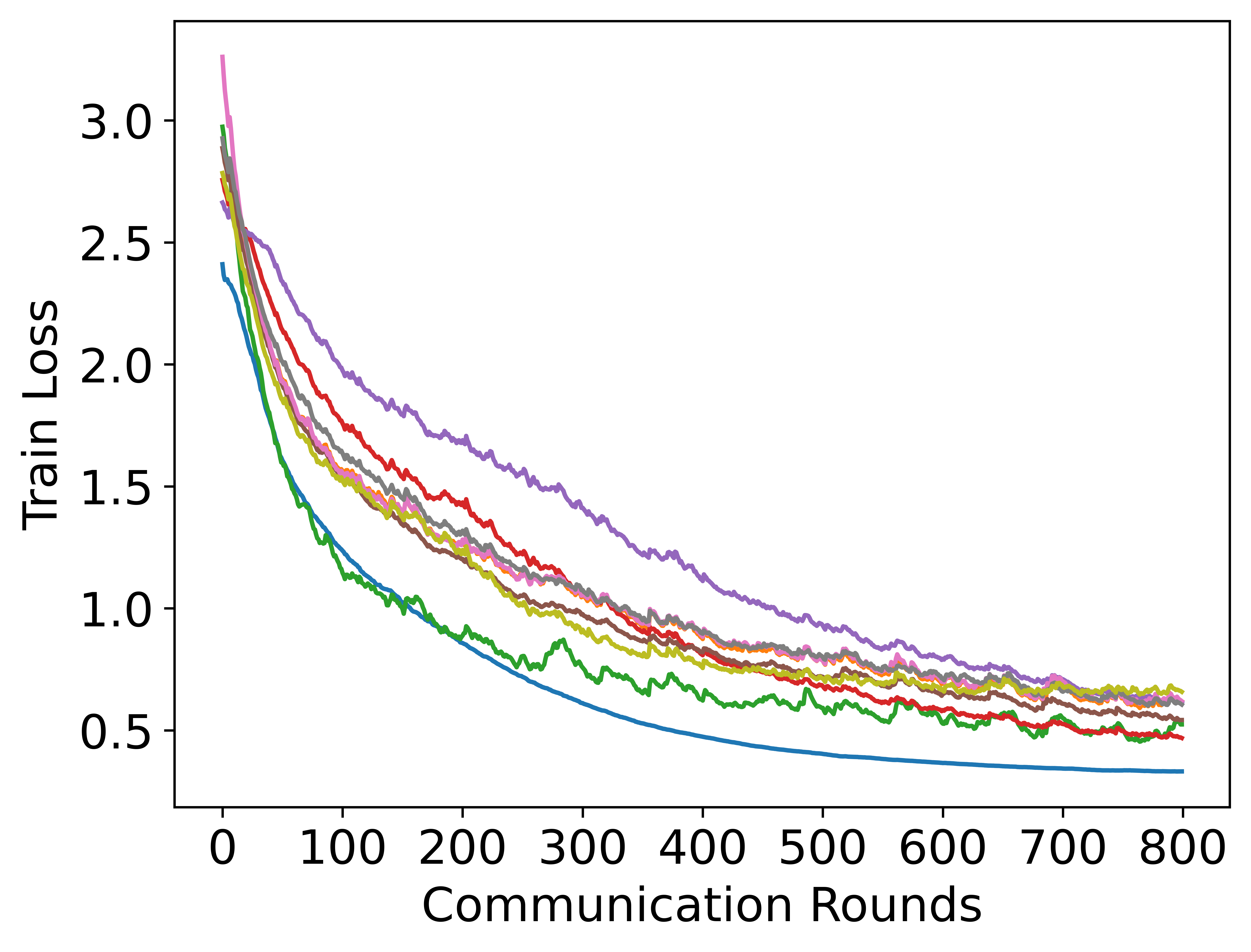}}
 \subcaptionbox{Loss  $c=6$}{\includegraphics[width=0.24\textwidth]{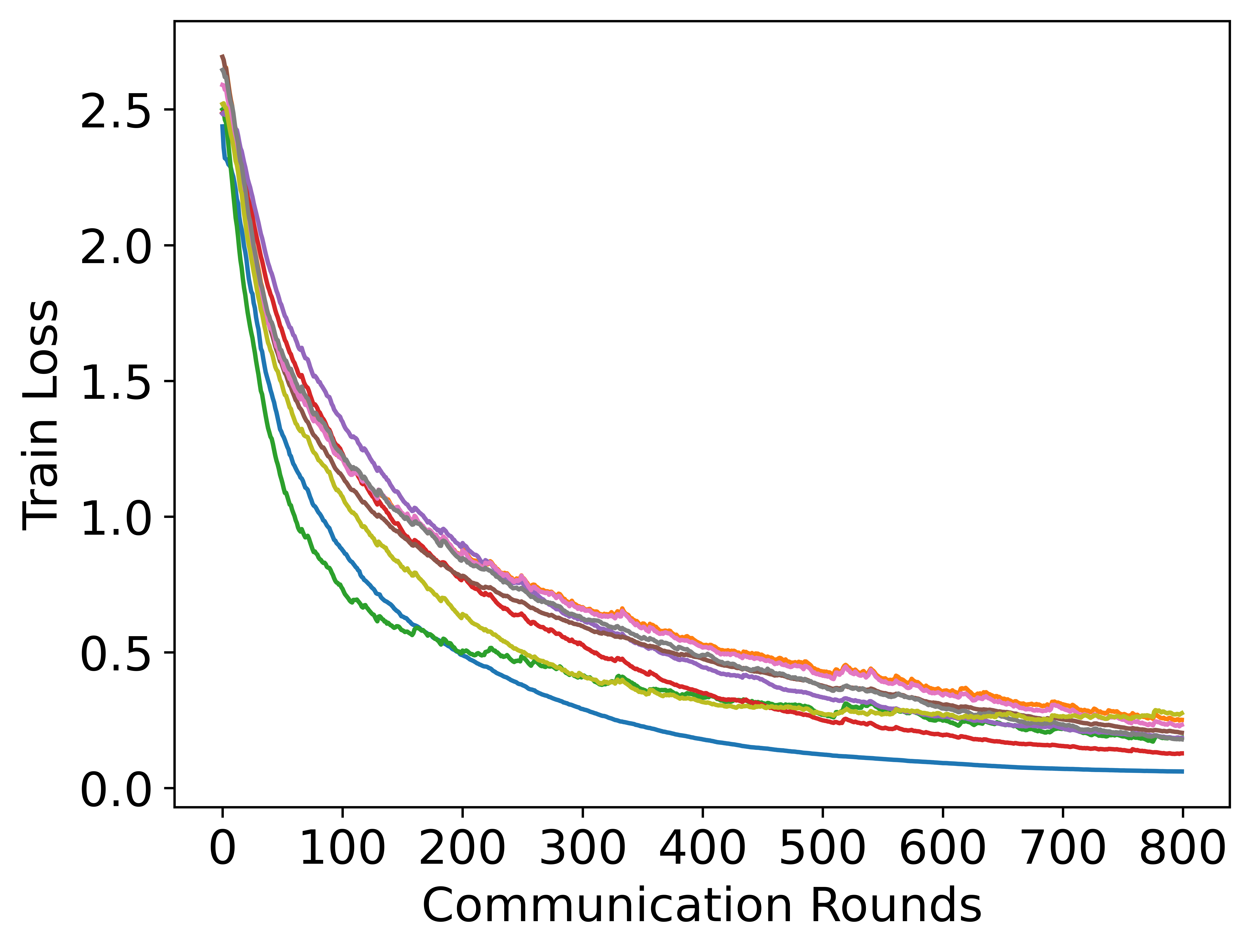}}
 \subcaptionbox{Loss  $c=3$}{\includegraphics[width=0.24\textwidth]{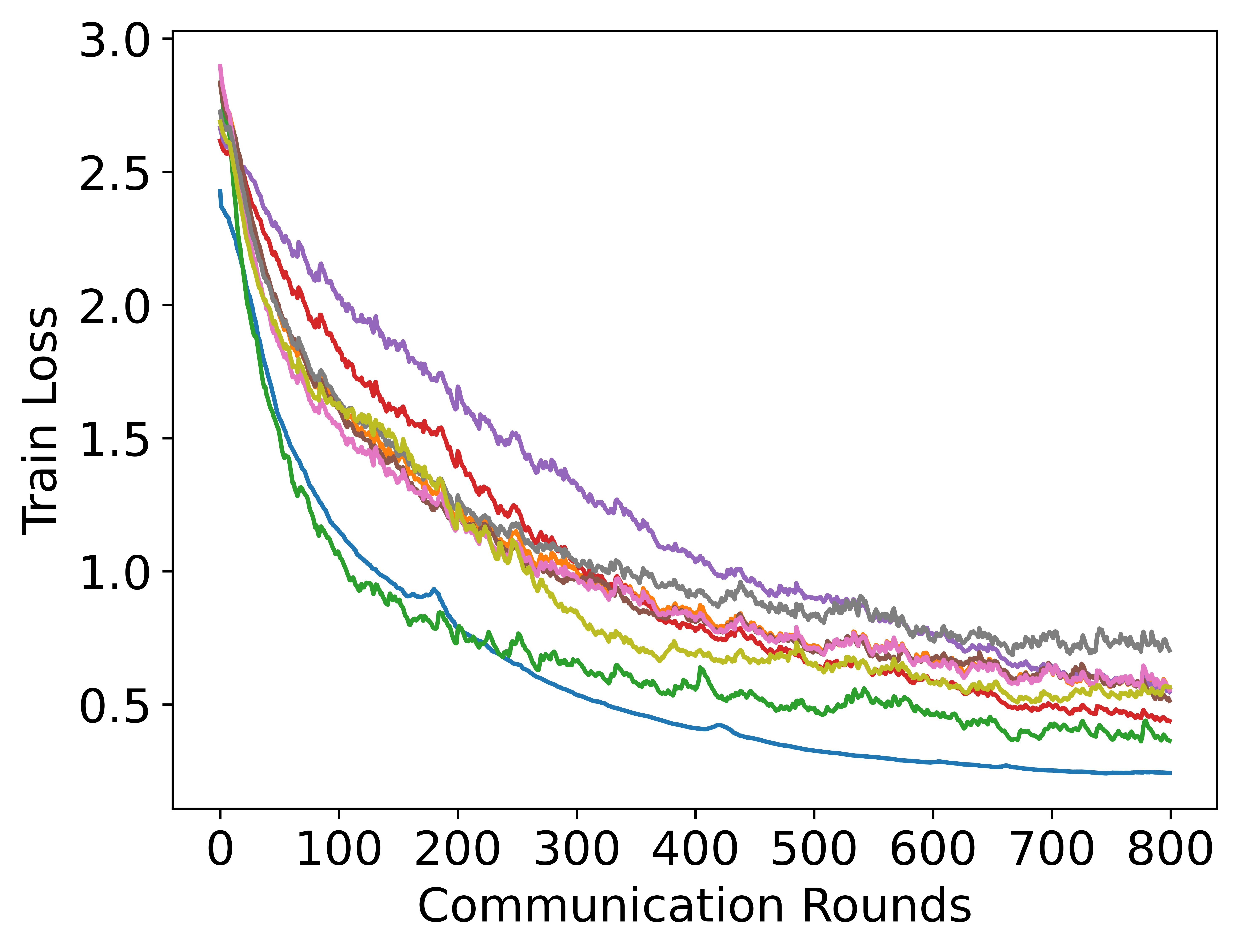}}

\caption{Accyracy/ Loss on the CIFAR-10 dataset under 10\% /5\% participation of total 100/200 clients}
\label{epx_curve}
\end{figure*}
As can be seen from the above figures, FedTOGA significantly outperforms other algorithms in scenarios with large heterogeneity (e.g., Dirichlet-0.1 and Pathological-3). Our algorithm still shows stability even when the number of clients decreases(e.g., 200 clients 5\% participation). These results are in line with our expectations. We aim to design an algorithm that enhances global consistency while efficiently finding a global flat minimum to improve generalization and reduce the computation and storage requirements of edge nodes.

In addition to the results shown in the main text, we also conducted experiments on CIFAR100. We followed the same parameter settings with FedSMOO, but we found that the baseline algorithm produced fluctuations in performance. Therefore, we report the best performance of previous studies (OLD) and the results of brand new experiments (NEW) in table \ref{tab_exp_resnet_extend}. In order to show the extraordinary performance of FedTOGA, we only report the historical best accuracy of all benchmarks in the main text.

\begin{sidewaystable}[]
\setlength{\tabcolsep}{0.1pt}
\centering
\caption{Test accuracy on CIFAR10/100 after 800 rounds under Dirichlet distribution and Pathological splits $u$ is the Dirichlet coefficient selected from $\{0.1,0.6\}$ and $c$ is the Pathological coefficient, which is the number of active categories in each client. The two datasets are divided into 100 clients and 10\% of them are active at each round in the upper part, while 200 and 5\% in the lower part.(ResNet18)}
\label{tab_exp_resnet_extend}
\scalebox{0.75}{
\begin{tabular}{l|c>{\columncolor{gray!60}}cc>{\columncolor{gray!60}}cc>{\columncolor{gray!60}}cc>{\columncolor{gray!60}}c|c>{\columncolor{gray!60}}cc>{\columncolor{gray!60}}cc>{\columncolor{gray!60}}cc>{\columncolor{gray!60}}c}
 \small
{Method}&\multicolumn{8}{c|}{CIFAR10}&\multicolumn{8}{c}{CIFAR100}\\ \cline{2-17}
 Partition&\multicolumn{4}{c}{Dirichlet}&\multicolumn{4}{c|}{Pathological}&\multicolumn{4}{c}{Dirichlet}&\multicolumn{4}{c}{Pathological}\\
Coefficient&\multicolumn{2}{c}{$u=0.6$}&\multicolumn{2}{c}{$u=0.1$}&\multicolumn{2}{c}{$c=6$}&\multicolumn{2}{c|}{$c=3$}&\multicolumn{2}{c}{$u=0.6$}&\multicolumn{2}{c}{$u=0.1$}&\multicolumn{2}{c}{$c=20$}&\multicolumn{2}{c}{$c=10$}\\\hline
Version&OLD&NEW&OLD&NEW&OLD&NEW&OLD&NEW&OLD&NEW&OLD&NEW&OLD&NEW&OLD&NEW\\\hline
FedAvg  &$79.52^{\pm 0.13}$&&$76.00^{\pm 0.18}$&&$79.91^{\pm 0.17}$&&$74.08^{\pm 0.22}$&&$46.35^{\pm 0.15}$&&$42.64^{\pm 0.22}$&&$44.15^{\pm 0.79}$&&$40.23^{\pm 0.31}$&\\
FedAdam &$77.08^{\pm 0.31}$&&$73.41^{\pm 0.33}$&&$77.05^{\pm 0.26}$&&$72.44^{\pm 0.29}$&&$48.35^{\pm 0.17}$&&$40.77^{\pm 0.31}$&&$41.26^{\pm 0.30}$&&$32.58^{\pm 0.22}$&\\
SCAFFOLD &$81.81^{\pm 0.17}$&&$78.57^{\pm 0.14}$&&$83.07^{\pm 0.10}$&&$77.02^{\pm 0.18}$&&$51.98^{\pm 0.23}$&&$44.41^{\pm 0.15}$&&$46.06^{\pm 0.22}$&&$41.08^{\pm 0.24}$&\\
FedCM &$82.97^{\pm 0.21}$&&$77.82^{\pm 0.16}$&&$83.44^{\pm 0.17}$&&$77.82^{\pm 0.19}$&&$51.56^{\pm 0.20}$&&$43.03^{\pm 0.26}$&&$44.94^{\pm 0.14}$&&$38.35^{\pm 0.27}$&\\
FedDyn &$83.22^{\pm 0.18}$&&$78.08^{\pm 0.19}$&&$83.18^{\pm 0.17}$&&$77.63^{\pm 0.14}$&&$50.82^{\pm 0.19}$&&$42.50^{\pm 0.28}$&&$44.19^{\pm 0.19}$&&$38.68^{\pm 0.14}$&\\
FedSAM&$80.10^{\pm 0.12}$&$81.46^{\pm 0.12}$&$76.86^{\pm 0.16}$&$77.03^{\pm 0.17}$&$80.80^{\pm 0.23}$&$81.13^{\pm 0.23}$&$75.51^{\pm 0.24}$&$78.30^{\pm 0.24}$&$47.51^{\pm 0.26}$&$48.09^{\pm 0.25}$&$43.43^{\pm 0.12}$&$44.56^{\pm 0.20}$&$45.46^{\pm 0.29}$&$47.68^{\pm 0.29}$&$40.44^{\pm 0.23}$&$42.35^{\pm 0.25}$\\
MoFedSAM &$84.13^{\pm 0.13}$&$85.29^{\pm 0.13}$&$78.71^{\pm 0.15}$&$80.25^{\pm 0.17}$&$84.82^{\pm 0.14}$&$84.74^{\pm 0.14}$&$79.57^{\pm 0.18}$&$83.09^{\pm 0.24}$&$54.38^{\pm 0.22}$&$54.50^{\pm 0.22}$&$44.85^{\pm 0.25}$&&$47.42^{\pm 0.26}$&$50.39^{\pm 0.30}$&$41.17^{\pm 0.22}$&$43.76^{\pm 0.21}$\\
FedGAMMA &$82.64^{\pm 0.14}$&$82.82^{\pm 0.16}$&$78.95^{\pm 0.15}$&$79.91^{\pm 0.15}$&$83.24^{\pm 0.19}$&$83.51^{\pm 0.18}$&$78.81^{\pm 0.14}$&$77.11^{\pm 0.14}$&$53.41^{\pm 0.20}$&&$46.39^{\pm 0.19}$&&$48.41^{\pm 0.14}$&&$43.24^{\pm 0.22}$&\\
FedSMOO &$84.55^{\pm 0.14}$&\textbf{86.08}$^{\pm 0.20}$&$\textbf{80.82}^{\pm 0.17}$&$\textbf{81.80}^{\pm 0.20}$&$85.39^{\pm 0.21}$&\textbf{86.38}$^{\pm 0.20}$&$81.58^{\pm 0.16}$&\textbf{82.79}$^{\pm 0.16}$&$53.92^{\pm 0.18}$&$52.12^{\pm 0.18}$&$46.48^{\pm 0.13}$&$47.94^{\pm 0.15}$&$48.87^{\pm 0.17}$&$49.01^{\pm 0.20}$&$44.10^{\pm 0.19}$&$43.40^{\pm 0.19}$\\
FedSpeed &-&{86.01}$^{\pm 0.16}$&-&{81.02}$^{\pm 0.16}$&-&$86.09^{\pm 0.19}$&-&$82.50^{\pm 0.16}$&-&$52.27^{\pm 0.18}$&-&&-&&-&\\
FedLESAM &$81.04^{\pm 0.19}$&$80.94^{\pm 0.16}$&$76.93^{\pm 0.16}$&$75.93^{\pm 0.15}$&$81.37^{\pm 0.17}$&$80.92^{\pm 0.19}$&$77.30^{\pm 0.22}$&$78.21^{\pm 0.21}$&$47.92^{\pm 0.19}$&&$44.48^{\pm 0.20}$&&$46.19^{\pm 0.21}$&&$41.20^{\pm 0.18}$&\\
FedLESAM-D &$84.27^{\pm 0.14}$&$83.60^{\pm 0.16}$&$80.08^{\pm 0.19}$&$78.87^{\pm 0.19}$&$85.62^{\pm 0.18}$&$83.66^{\pm 0.19}$&$83.00^{\pm 0.22}$&$82.21^{\pm 0.21}$&$53.27^{\pm 0.17}$&&$46.42^{\pm 0.23}$&&$48.26^{\pm 0.18}$&&$43.26^{\pm 0.18}$&\\
FedLESAM-S&\textbf{84.94$^{\pm 0.12}$}&$83.66^{\pm 0.13}$&$79.52^{\pm 0.17}$&$78.77^{\pm 0.16}$&$\textbf{85.88}^{\pm 0.19}$&$82.99^{\pm 0.19}$&$\textbf{82.18}^{\pm 0.15}$&{81.01}$^{\pm 0.17}$&$54.61^{\pm 0.20}$&$48.47^{\pm 0.20}$&$48.07^{\pm 0.19}$&&$50.26^{\pm 0.18}$&&$44.42^{\pm 0.17}$&\\
FedTOGA(ours)&&\textcolor{red}{\textbf{86.99}$^{\pm 0.13}$}&&\textcolor{red}{\textbf{83.16}$^{\pm 0.17}$}&&\textcolor{red}{\textbf{87.21}$^{\pm 0.18}$}&&\textcolor{red}{\textbf{84.55}$^{\pm 0.15}$}&&\textcolor{red}{\textbf{55.20}$^{\pm 0.17}$}&&\textcolor{red}{\textbf{48.72}$^{\pm 0.17}$}&&&&\\\hline
FedAvg  &$75.90^{\pm 0.21}$&&$72.93^{\pm 0.19}$&&$77.47^{\pm 0.34}$&&$71.68^{\pm 0.34}$&&$44.70^{\pm 0.22}$&&$40.41^{\pm 0.33}$&&$38.32^{\pm 0.25}$&&$36.79^{\pm 0.32}$&\\
FedAdam&$75.55^{\pm 0.38}$&&$69.70^{\pm 0.32}$&&$75.74^{\pm 0.22}$&&$70.49^{\pm 0.26}$&&$44.33^{\pm 0.26}$&&$38.04^{\pm 0.25}$&&$35.14^{\pm 0.16}$&&$30.28^{\pm 0.28}$&\\
SCAFFOLD&$79.00^{\pm 0.26}$&&$76.15^{\pm 0.15}$&&$80.69^{\pm 0.21}$&&$74.05^{\pm 0.31}$&&$50.70^{\pm 0.18}$&&$41.83^{\pm 0.29}$&&$39.63^{\pm 0.31}$&&$37.98^{\pm 0.36}$&\\
FedCM &$80.52^{\pm 0.29}$&&$77.28^{\pm 0.22}$&&$81.76^{\pm 0.24}$&&$76.72^{\pm 0.25}$&&$50.93^{\pm 0.31}$&&$42.33^{\pm 0.19}$&&$42.01^{\pm 0.17}$&&$38.35^{\pm 0.24}$&\\
FedDyn&$80.69^{\pm 0.23}$&&$76.82^{\pm 0.17}$&&$82.21^{\pm 0.18}$&&$74.93^{\pm 0.22}$&&$47.32^{\pm 0.18}$&&$41.74^{\pm 0.21}$&&$41.55^{\pm 0.18}$&&$38.09^{\pm 0.27}$&\\
FedSAM&$76.32^{\pm 0.16}$&$78.32^{\pm 0.16}$&$73.44^{\pm 0.14}$&$74.00^{\pm 0.14}$&$78.16^{\pm 0.27}$&$78.75^{\pm 0.27}$&$72.41^{\pm 0.29}$&$75.12^{\pm 0.29}$&$45.98^{\pm 0.27}$&$44.93^{\pm 0.26}$&$40.22^{\pm 0.27}$&$41.13^{\pm 0.27}$&$38.71^{\pm 0.23}$&$43.20^{\pm 0.23}$&$36.90^{\pm 0.29}$&$38.67^{\pm 0.29}$\\
MoFedSAM &$82.58^{\pm 0.21}$&$84.76^{\pm 0.20}$&$78.43^{\pm 0.24}$&$80.10^{\pm 0.14}$&$84.46^{\pm 0.20}$&$85.00^{\pm 0.27}$&$79.93^{\pm 0.19}$&$82.13^{\pm 0.29}$&$53.51^{\pm 0.25}$&$49.97^{\pm 0.27}$&$42.22^{\pm 0.23}$&$42.23^{\pm 0.23}$&$42.77^{\pm 0.27}$&$45.39^{\pm 0.23}$&$39.81^{\pm 0.21}$&$38.83^{\pm 0.25}$\\
FedGAMMA &$80.72^{\pm 0.19}$&$78.31^{\pm 0.19}$&$76.41^{\pm 0.17}$&$76.70^{\pm 0.14}$&$81.81^{\pm 0.17}$&$81.59^{\pm 0.27}$&$76.58^{\pm 0.21}$&$77.44^{\pm 0.29}$&$50.61^{\pm 0.19}$&&$43.77^{\pm 0.19}$&&$43.35^{\pm 0.24}$&&$38.46^{\pm 0.22}$&\\
FedSMOO &$82.94^{\pm 0.19}$&$84.96^{\pm 0.19}$&$79.76^{\pm 0.19}$&$77.90^{\pm 0.14}$&$84.82^{\pm 0.18}$&$84.32^{\pm 0.27}$&$81.01^{\pm 0.19}$&$78.91^{\pm 0.29}$&$53.45^{\pm 0.19}$&&$45.83^{\pm 0.18}$&&$44.70^{\pm 0.21}$&&$43.41^{\pm 0.22}$&\\
FedSpeed &-&$84.12^{\pm 0.18}$&-&$76.74^{\pm 0.14}$&-&$84.78^{\pm 0.27}$&-&$79.09^{\pm 0.29}$&-&&-&&-&&-&\\
FedLESAM &$77.74^{\pm 0.18}$&$77.80^{\pm 0.18}$&$73.73^{\pm 0.22}$&$73.03^{\pm 0.14}$&$78.44^{\pm 0.20}$&$77.91^{\pm 0.27}$&$74.53^{\pm 0.19}$&$74.47^{\pm 0.29}$&$45.00^{\pm 0.16}$&&$41.87^{\pm 0.23}$&&$42.14^{\pm 0.18}$&&$39.32^{\pm 0.24}$&\\
FedLESAM-D&$82.53^{\pm 0.19}$&$81.69^{\pm 0.18}$&$79.56^{\pm 0.27}$&$75.17^{\pm 0.14}$&$85.04^{\pm 0.21}$&$82.07^{\pm 0.27}$&$81.10^{\pm 0.19}$&$77.93^{\pm 0.29}$&$51.14^{\pm 0.20}$&&$45.09^{\pm 0.24}$&&$43.97^{\pm 0.26}$&&$42.63^{\pm 0.29}$&\\
FedLESAM-S&$83.22^{\pm 0.22}$&$78.89^{\pm 0.18}$&$78.69^{\pm 0.17}$&$73.80^{\pm 0.14}$&$85.02^{\pm 0.24}$&$82.07^{\pm 0.27}$&$80.57^{\pm 0.17}$&$74.62^{\pm 0.29}$&$52.26^{\pm 0.18}$&&$44.82^{\pm 0.20}$&&$45.68^{\pm 0.19}$&&$43.89^{\pm 0.23}$&\\
FedTOGA(ours)&&\textcolor{red}{\textbf{85.21}$^{\pm 0.17}$}&&\textcolor{red}{\textbf{81.60}$^{\pm 0.16}$}&&\textcolor{red}{\textbf{85.24}$^{\pm 0.19}$}&&\textcolor{red}{\textbf{83.25}$^{\pm 0.20}$}&&&&&&&&\\\hline
\multicolumn{17}{l}{\textbf{Note}: Since replication in the same experimental setting produced differences in performance, we report in the main text the best performance of all experiments in previous studies.}
\end{tabular}}
\end{sidewaystable}
\newpage

\subsection{Training Speed}

\begin{table*}[htb]
\centering
\caption{Number of communication rounds required to reach a given accuracy. We recorded the first round of communication to reach a given accuracy to show the improvement in the number of training rounds compared to the other algorithms in the Dirichlet-0.1/0.6 and Pathological-6.0/3.0 settings. We mainly compared the SAM-based FL algorithms benchmark.}
\label{tab_time_extend}
\begin{tabular}{l|cccc|cccc}
Partition&\multicolumn{4}{c|}{Dirichlet}&\multicolumn{4}{c}{Pathological}\\ \cline{2-9}
 Coefficient&\multicolumn{2}{c}{$u=0.6$}&\multicolumn{2}{c|}{$u=0.1$}&\multicolumn{2}{c}{$c=6$}&\multicolumn{2}{c}{$c=3$}\\
Acc/Rounds&{80\%}&{82\%}&{76\%}&{78\%}&{80\%}&{82\%}&{76\%}&{78\%}\\\hline
FedSAM  &491&800+&587&800+&443&790&465&691\\
MoFedSAM \cite{pmlr-v162-qu22a}  &167&270&303&425&135&253&167&265\\
FedGAMMA\cite{10269141} &458&630&369&591&407&550&701&800+\\
FedSMOO \cite{sun2023dynamic}  &190&253&302&402&205&263&262&322\\
FedSpeed \cite{sun2023fedspeed}  &262&318&445&530&233&292&349&438\\
FedLESAM  &588&800+&800+&800+&620&800+&497&778\\
FedLESAM-D &248&418&369&663&224&376&393&452\\
FedLESAM-S\cite{FedLESAM} &390&643&529&800+&348&602&497&800+\\
FedTOGA(ours) &135&170&184&226&134&166&158&200\\\hline
\multicolumn{9}{l}{\textbf{Note}: The SGD method is not considered.}
\end{tabular}
\end{table*}
According to the above table \ref{tab_time_extend}, we can see that FedTOGA performs far better than the rest of the algorithms. It has the fastest convergence rate while maintaining high accuracy. The SAM optimizer usually slows down the whole training process due to the need to compute additional perturbations to the ascent process, which will be improved by enhancing consistency. MoFedSAM enforces consistency by employing global momentum on each local client and weighting it by a factor a (usually 0.1), which means that local knowledge will be forcibly overwritten by global gradient, which, while speeding up convergence in the early stage, may not be able to take any further from the local knowledge in the later stage. However, it may not be able to draw further effective learning progress from the local in the later stage. FedTOGA corrects local perturbations and dynamic regularizers by guiding global updates, greatly enhancing the consistency of generalization and optimization. Therefore, our method can effectively accelerate the modeling speed and improve the modeling accuracy, especially in the case of large-scale heterogeneous. From Table \ref{tab_time2}, it can be observed that FedTOGA has a similar local computation time as the SAM-based FL algorithm.
\begin{small}
\begin{table*}[htb]
\centering
\caption{ wall clock time(training, loading, evaluation) on CIFAR10 ResNet18 $u=0.6$, 0.1 and 100 clients.}
\label{tab_time2}
\begin{tabular}{c|ccccccc}\hline
  &FedSAM & MoFedSAM & FedGAMMA&FedSpeed& FedSMOO&FedLESAM-D&FedTOGA\\\hline
time&25.71s&28.73s&29.88s&28.98s&29.67s&25.70s&29.12s\\\hline
\end{tabular}
\end{table*}
\end{small}

\subsection{Ablation studies}
\begin{table*}[htb]
\centering
\caption{ Ablation studies of different modules.}
\begin{tabular}{cccc|cc}\hline
SAM  & Dynamic Regularization & Dual Correction & SAM Correction & CIFAR10 Acc& CIFAR100 Acc\\\hline
\checkmark&-&-&-&81.39\%&48.08\%\\\hline
\checkmark&\checkmark&-&-&84.14\%&53.79\%\\\hline
\checkmark&\checkmark&\checkmark&-&85.54\%&56.85\%\\\hline
\checkmark&\checkmark&\checkmark&\checkmark&86.01\%&57.25\%\\\hline
\end{tabular}
\end{table*}
We tested the performance of different modules called “SAM”, “Dynamic Regularization”, “Dual Variable Correction” and “SAM Perturbation Correction” modules on the Dirichlet 0.6 partitioned CIFAR-10/100 dataset, LeNet network. The benchmark is FedSAM. After the sequential introduction of the different modules, the CIFAR10 accuracy increased by 2.75\%, 4.15\%, and 4.62\% compared to the FedSAM; the CIFAR100 accuracy increased by 5.71\%, 8.77\%, and 9.17\% compared to the FedSAM.

\end{document}

%% file: preamble.tex
\usepackage[dvipsnames]{xcolor}
